\documentclass[5p,times,twocolumn,authoryear]{elsarticle}

\usepackage[OT1]{fontenc} %
\usepackage{amsmath,amssymb,amsfonts,bm}
\usepackage{booktabs}
\usepackage{graphicx}
\usepackage{enumitem}
\usepackage{adjustbox}
\usepackage{array}
\usepackage{microtype}
\usepackage{xcolor}
\definecolor{linkblue}{HTML}{335A8A}
\usepackage[colorlinks=true,linkcolor=linkblue,citecolor=linkblue,urlcolor=linkblue]{hyperref}
\usepackage{url}
\usepackage[font=small,labelfont=bf,labelsep=period,skip=4pt]{caption}
\newcolumntype{L}[1]{>{\raggedright\arraybackslash}p{#1}}
\setlist{nosep,leftmargin=*}
\makeatletter\g@addto@macro{\UrlBreaks}{\do\-}\makeatother
\biboptions{authoryear,round}
\journal{arXiv}

\newcommand{\ms}{\texttt{MedScope-4B}}          %
\newcommand{\msp}{\texttt{MedScope-4B-P}}       %
\newcommand{\msn}{\texttt{MedScope-4B-N}}       %
\newcommand{\base}{\texttt{Qwen3.5-4B}}         %
\newcommand{\gem}{\texttt{gemini-4-argon}}
\newcommand{\opus}{\texttt{claude-opus-5-5}}
\newcommand{\gpt}{\texttt{gpt-6-astra}}
\newcommand{\ci}[2]{[#1,\,#2]}                  %

\begin{document}

\begin{frontmatter}

\title{Answers of vision-language models persist after lesions are erased from medical images: a matched counterfactual evaluation}

\author[ind]{Bhaskar Gurram\corref{cor}}
\ead{gurrambhaskar.ai@gmail.com}
\author[ind]{Yoshitha Challagulla}
\author{Jyothirmaye Challagulla}
\author[shn,uoft]{Samir C. Grover}
\cortext[cor]{Corresponding author.}
\address[ind]{Independent Researcher}
\address[shn]{Scarborough Health Network Research Institute, Toronto, Ontario, Canada}
\address[uoft]{Department of Medicine, University of Toronto, Toronto, Ontario, Canada}

\begin{abstract}
Medical vision-language models (VLMs) are tested on unaltered images, where a correct answer cannot show use of the image. We erased each lesion by inpainting, made a matched control edit elsewhere and asked one yes/no question per view (507 cases, five public sources). The evidence sensitivity gap (ESG), the share of erased minus control-edited images answered ``absent'' or referred, is zero for constant answers and pure edit detectors. Nine commercial VLMs, correct on 82.9\% to 88.8\% of lesion images, reported the finding on 41.1\% to 61.6\% of 219 erased images (ESG 0.269 to 0.470), 9.0\% to 30.9\% of 288 lesion-free images and no blank image. \ms, trained on such views, reported 9.6\% (ESG 0.758, 95\% CI 0.701 to 0.815) while correct on 84.9\%. On a fresh external set it reported 8.3\% against 31.5\% to 42.3\% for three reference models, with fewer lesion images correct (47.6\% against 74.1\% for \gem). On 103 unedited same-patient colonoscopy pairs it showed no advantage (natural ESG 0.243, as for its base model). A post hoc variant also trained on unedited images (\msn) had natural ESG 0.612 at a hospital not used in training, above two reference models but not shown to exceed the third, and reported 22.0\% of erased lesions on the fresh external set. On 20 of 36 erasures that specialists accepted, the nine VLMs still reported the finding on 43.3\% (95\% CI 31.7 to 56.1) and \ms{} on 1. Counterfactual training reduced persistence on edited images; unedited pairs beyond colonoscopy are needed.
\end{abstract}


\end{frontmatter}

\section{Introduction}\label{sec:intro}

Vision-language models (VLMs) are being proposed as assistants for the interpretation of medical images: a clinician shows an image and asks whether a named finding is present \citep{sellergren2025medgemma,xu2025lingshu,chen2024huatuogpt}. A finding reported on an image that does not show it can lead to further imaging, biopsy or treatment that the patient did not need, and a finding missed on an image that shows it delays diagnosis. A model can avoid either error by making the other, so the two have to be measured together. Neither can be interpreted unless the answer depends on what the image shows.

Current evaluations measure accuracy on unaltered images with a reference answer \citep{lau2018vqarad,liu2021slake,he2020pathvqa}. Accuracy on unaltered images has a logical limit: when the image is never varied, a correct answer cannot show that the image was used, because a model that answers from the finding named in the question, or from what usually accompanies the anatomy shown, gives the same answer. Probes that vary the question on an unaltered image show that models affirm findings that are absent \citep{li2023pope,yan2024probmed}, but they leave the same limit in place.

Studies that changed the text or the whole image show that this limit matters in practice. On published case challenges, the accuracy of one model fell steeply when a misleading vignette was added \citep{buckley2026text}. On questions chosen because they need the image, most models still answered well above chance without it \citep{gu2025illusion}, and when a chest radiograph was exchanged for one with the opposite label, several systems kept many of their correct answers \citep{lotfinia2026image}. Image classifiers have likewise relied on the hospital of origin or on skin markings rather than on signs of disease \citep{zech2018variable,degrave2021shortcuts,winkler2019markings,oakden2020hidden,geirhos2020shortcut}. Removing or exchanging the whole image, however, changes everything in it at once. It cannot show whether an answer about one finding depends on the visible evidence for that finding.

We set four requirements for an intervention on the finding itself. First, the lesion must be removed while the rest of the image, and the question, stay the same. Second, the act of editing needs a control: an edited image differs from an unedited one in more than the lesion, and inpainting can reveal what was removed through the traces it leaves \citep{rong2022road}, so a model could respond to the edit and not to the absence of the lesion. The control is the same edit applied where there is no lesion. Third, the removal must be checked, because an automatic erasure can leave part of the lesion or its secondary signs, and whether the finding is gone is a judgement for a specialist reader. Fourth, the result must be tested on images with no edit, because models do not meet edited images in use; unedited images of the same patient with and without the finding provide such a test. To our knowledge, no evaluation of medical VLMs has combined an erasure confined to the lesion, a matched control edit and a test on unedited same-patient pairs (Section~\ref{sec:related}).

The aim of this study was to test whether the answer of a VLM about a named finding depends on the visible evidence for that finding, and whether training on images that differ in that evidence changes this, on edited and on unedited images. For each case we built an erased image, a control-edited image and an unedited lesion-free image (together, MedScope-Bench). We evaluated nine VLMs accessed through their providers' interfaces (the comparison models) on these images, trained a small open-weight model, \ms, on images built in the same way, and tested it with editing tools not used in training, on external datasets and on unedited same-patient colonoscopy pairs. Physicians who were not told which images were edited read a sample of the erased images. After the result on unedited pairs was known, a second model trained in addition on unedited images, \msn, was designed post hoc and tested on pairs from a hospital not used in training. The images came from public datasets of colonoscopy, chest X-ray, breast ultrasound, magnetic resonance imaging (MRI) and computed tomography (CT).

\section{Related work}\label{sec:related}

\textit{Changing the question.} Most tests of whether a VLM reports a finding that is absent change the question and leave the image as it is. POPE asks balanced yes/no questions about objects that are or are not in a photograph \citep{li2023pope}. ProbMed pairs each true question about a medical image with an adversarial question that names an invented finding or attribute, and scores the pair only when both answers are right \citep{yan2024probmed}. CARES, Med-HallMark and MedHEval grade answers or generated text about unaltered medical images over wider sets of properties \citep{xia2024cares,chen2024medhallmark,chang2024medheval}. \citet{jin2024hidden} had physicians read the rationales of one model and found flawed rationales behind many correct answers, most often in image comprehension. In all of these studies the image is held fixed, so a correct ``no'' cannot be attributed to the image.

\textit{Changing the image.} A second group holds the question fixed and changes the whole image. MedVH replaces a chest radiograph with an unrelated image \citep{gu2024medvh}, and \citet{gu2025illusion} remove the image from multiple-choice cases or substitute one that fits another answer option. The closest work is that of \citet{lotfinia2026image}. They exchanged a chest radiograph for another patient's with the same or the opposite label, occluded the radiologist-marked region or an equal region elsewhere, and had radiologists read the same altered images. A language model given only the question scored $55.3\%$ on their pooled questions, above two multimodal systems. The authors state that their swaps are not minimal pairs, and that images which remove the finding and preserve the image statistics would be more faithful. Such paired edits exist for general images \citep{agarwal2020towards}: HallusionBench pairs an original with a hand-edited image for which the correct answer differs \citep{guan2024hallusionbench}. In medical imaging, RadEdit edits chest radiographs with a diffusion model to stress-test classifiers and segmentation models \citep{perez2024radedit}, and HEAL-MedVQA swaps the queried anatomical region of a chest radiograph \citep{healmedvqa2024}. Recent preprints replace the whole image \citep{zafar2026coral}, exchange tumour intensities between MRI sequences \citep{sun2026prive}, or build counterfactuals from image features or local perturbations \citep{huang2026medcap,lin2026medcausalx}. None of the studies that edit the image content (RadEdit and those listed after it in Table~\ref{tab:related}) pairs a lesion-shaped erasure with an edit of the same shape and tool elsewhere in the same image, or reports an assessment of the edited images by clinicians.

\textit{Training remedies.} Similar pairs have been used for training. MMedPO adds noise to lesion regions to build the rejected side of preference pairs \citep{zhu2024mmedpo}. Several methods train general VLMs on minimal-pair or contrasting images with preference optimisation \citep{svco2024,wu2025empo,jiang2025mfpo,rafailov2023dpo}, and CORAL penalises answers that stay the same when the image is replaced \citep{zafar2026coral}. These methods are evaluated by accuracy, or by the rate of unsupported statements, on benchmarks of unaltered images. Such measures do not separate answering ``absent'' more often from answering ``absent'' when the finding is absent. A matched control edit does, because a model that answers ``absent'' more often also does so when the lesion is intact. The present study applies this control to a trained model and to the models it is compared with, and adds a test on unedited image pairs.

\begin{table*}[!t]
\caption{Design features of the closest prior work and of this study, taken from the published descriptions. Yes: the feature is present. Partly: present in the restricted form given in the note. No: absent. --: not applicable. Not stated: not stated in the sources consulted. Works are ordered as in the text.}
\label{tab:related}
\centering
\footnotesize
\setlength{\tabcolsep}{3pt}
\renewcommand{\arraystretch}{1.05}
\begin{tabular}{@{}L{0.27\textwidth}*{7}{>{\centering\arraybackslash}p{0.088\textwidth}}@{}}
\toprule
\textbf{Study} & \textbf{Medical images} & \textbf{Image varied, question fixed} & \textbf{Edit confined to the finding} & \textbf{Matched control edit} & \textbf{Altered images assessed by clinicians} & \textbf{Unedited same-patient pairs} & \textbf{Trained remedy} \\
\midrule
ProbMed \citep{yan2024probmed} & Yes & No & -- & -- & -- & No & No \\
MedVH \citep{gu2024medvh} & Yes & Yes$^{a}$ & No & No & No & No & No \\
\citet{gu2025illusion} & Yes & Yes$^{a}$ & No & No & Yes$^{b}$ & No & No \\
\citet{lotfinia2026image} & Yes & Yes & Partly$^{c}$ & Yes$^{d}$ & Yes & No$^{e}$ & No \\
HallusionBench \citep{guan2024hallusionbench} & No & Yes & Partly$^{f}$ & No & -- & No & No \\
RadEdit \citep{perez2024radedit} & Yes & Yes$^{g}$ & Yes & No & No & No & No \\
HEAL-MedVQA \citep{healmedvqa2024} & Yes & Yes & Partly$^{h}$ & No & No & Not stated & Yes \\
PriVE-Bench \citep{sun2026prive} & Partly$^{i}$ & Yes & Partly$^{i}$ & No & No & Not stated & Not stated \\
MMedPO \citep{zhu2024mmedpo} & Yes & No$^{j}$ & Partly$^{j}$ & No & No & Not stated & Yes \\
CORAL \citep{zafar2026coral} & Yes & Yes$^{a}$ & No & No & No & Not stated & Yes \\
\midrule
This study & Yes & Yes & Yes & Yes & Partly$^{k}$ & Yes$^{l}$ & Yes \\
\bottomrule
\multicolumn{8}{@{}p{0.97\textwidth}@{}}{$^{a}$\,The whole image is removed or replaced. $^{b}$\,Clinicians verified that each substituted image supported the new answer. $^{c}$\,The radiologist-marked region is occluded, not removed. $^{d}$\,An equal region occluded elsewhere, and a swap with a radiograph of the same label. $^{e}$\,Swaps use unedited radiographs of other patients. $^{f}$\,Hand edits of general images (charts, maps, illusions). $^{g}$\,Classification and segmentation models; no questions are asked. $^{h}$\,The queried anatomical region is swapped in from another radiograph. $^{i}$\,One of five domains is medical: tumour-region intensities exchanged between MRI sequences. $^{j}$\,Noise in lesion regions, used to build training pairs; evaluation uses unaltered images. $^{k}$\,Two trainee physicians (all imaging types), a board-certified radiologist and a board-certified gastroenterologist (Section~\ref{subsec:res_readers}). $^{l}$\,Colonoscopy frames, and chest X-ray follow-up pairs as a secondary analysis; the pneumothorax films are unedited but not paired by patient.} \\
\end{tabular}
\end{table*}

\section{Materials and methods}
\label{sec:methods}

\subsection{Study design and predictions}
\label{subsec:design}

\begin{figure*}[!t]
\centering
\includegraphics[width=\textwidth]{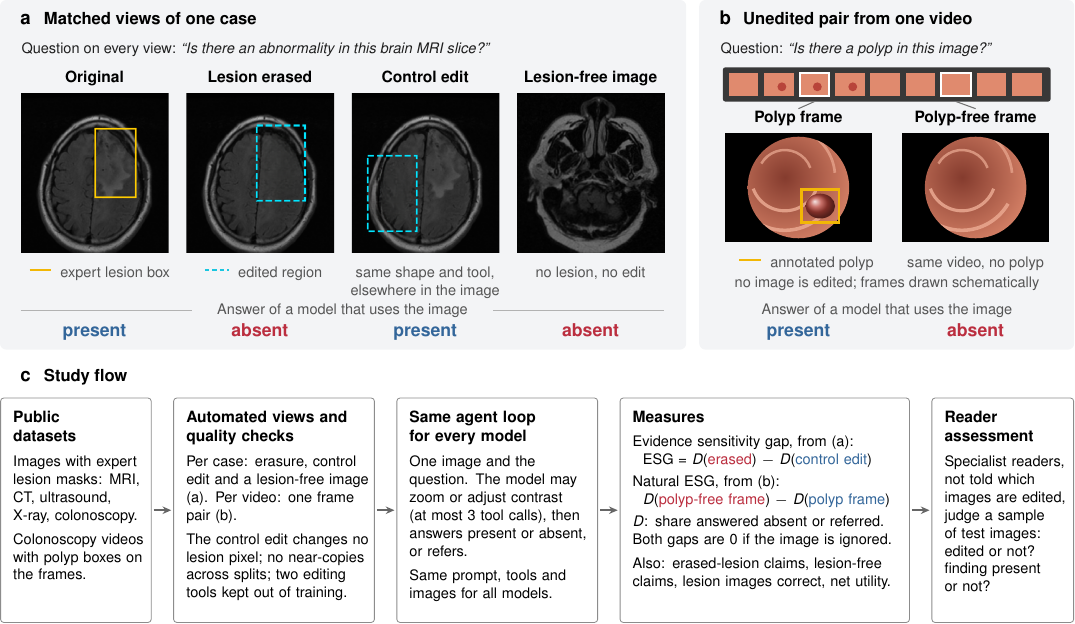}
\caption{Study design. (a) Matched views of one case (brain MRI, TCGA-LGG): the original image, the erased image, the control edit (a region of the same shape filled with the same tool elsewhere; lesion untouched) and an unedited lesion-free image of the same imaging type. The expert lesion annotation and the edited regions are outlined. Every view is shown with the same yes/no question, and the answer expected from a model whose answer depends on the lesion is given below each view. (b) The second design: an unedited frame with an annotated polyp and an unedited polyp-free frame of the same colonoscopy video (frames drawn schematically). (c) Study flow, from the public datasets to the measures and the reader assessment. ESG: evidence sensitivity gap.}
\label{fig:design}
\end{figure*}

The study varies the image and holds the question fixed (Fig.~\ref{fig:design}). Each test case is an image with an expert lesion mask and a yes/no question whose answer depends on that lesion, for example ``Is there a polyp in this image?''. In the \emph{erased} image the lesion region is filled in by inpainting. In the \emph{control edit} the same tool fills a region of the same shape elsewhere in the image and leaves the lesion untouched. Unedited \emph{lesion-free} images of the same imaging type receive the same kind of question. We call the test images and their scoring MedScope-Bench.

A model whose answer depends on the lesion should report the finding on the original image, answer ``absent'' (or refer) on the erased image, and still report it after the control edit. A model that answers from the question or the surrounding anatomy answers all three alike, and one that answers ``absent'' to any trace of editing does so on both edited images. The evidence sensitivity gap (ESG; Section~\ref{subsec:measures}) is zero for both and for any constant answer. Because reporting nothing also avoids reporting erased lesions, the share of lesion images answered correctly is given with every ESG.

The design measures whether the answer changes when the lesion region is filled in, not whether the finding has gone: an erasure may leave part of the lesion or its secondary signs. Physicians therefore read a sample of the images without being told which were edited (Section~\ref{subsec:readers}), and a second design, with no edit, pairs an unedited image that shows a lesion with one of the same patient that does not (Section~\ref{subsec:pairs}). Table~\ref{tab:overview} lists the questions in the order of the Results.

\begin{table*}[!t]
\caption{Overview of the study, in the order of the Results. Each row gives a question, the design element that answers it, the test data and the section that reports it. Every analysis, with its status (pre-specified, descriptive or post hoc), its decision rule and its outcome, is listed in Supplementary Tables~\ref{tab:s_overview} and~\ref{tab:s_plan}. Comparison models: the nine models called through their providers' interfaces; reference models: \gem, \opus{} and \gpt{} (Section~\ref{subsec:models}). ESG: evidence sensitivity gap.}
\label{tab:overview}
\centering
\footnotesize
\setlength{\tabcolsep}{3pt}
\renewcommand{\arraystretch}{1.06}
\begin{tabular}{@{}L{0.25\textwidth}L{0.33\textwidth}L{0.29\textwidth}L{0.09\textwidth}@{}}
\toprule
\textbf{Question} & \textbf{Design element} & \textbf{Test data} & \textbf{Section} \\
\midrule
Do the answers of the comparison models persist after the lesion region is erased? & Erased image against matched control edit; unedited lesion-free and blank images; question type; prompting conditions & Main test: 507 cases (219 with a lesion), 1,115 views & \ref{subsec:res_main} \\
Do physicians still report the finding on erased images? & Readers without labels: a specialist round (radiologist, gastroenterologist), one earlier trainee round; model answers on reader-accepted erasures & 212 and 128 main-test images & \ref{subsec:res_readers} \\
Does training on counterfactual views change the answers, and do other designs do better? & \ms; first-stage models with and without counterfactual views over three seeds and on a second base model; seventeen alternative designs and 270 decision rules & Main test; validation set; tools not used in training; fresh external set & \ref{subsec:res_trained}, \ref{subsec:res_designs} \\
Is the result of \ms{} explained by detection of the edit? & Matched control edit; tools not used in training; edit-site control and partial erasures; low-visibility erasures & Main test (125 to 219 lesion cases); validation lesions & \ref{subsec:res_alternatives} \\
Does it hold on data not used in training? & \ms{} run once on each set; \msp{} (post hoc) for pneumothorax & External datasets (169 cases); fresh internal (76); fresh external (316); 200 new SIIM-ACR patients & \ref{subsec:res_external} \\
Does it hold on unedited same-patient pairs? & Polyp frame against polyp-free frame of the same video; \ms, and \msn{} (post hoc) & REAL-Colon: 103 pairs; LDPolypVideo: 116 pairs & \ref{subsec:res_pairs} \\
What do unedited training images change on the other tests? & \msn{} (post hoc) on every earlier test set and on unedited chest films; three seeds & Main test; external datasets; fresh sets; 600 CheXthought and 200 SIIM-ACR films & \ref{subsec:res_nmodel} \\
\bottomrule
\end{tabular}
\end{table*}

\subsection{Data, splits and leakage controls}
\label{subsec:data}

We chose sources that release expert lesion masks and lesion-free images of the same imaging type; training, validation and the main test come from five public sources (Table~\ref{tab:datasets}). Lesions under 20 pixels or over 20\% of the image and 623 diffuse chest X-ray findings were excluded, so the test covers focal lesions that can be erased. The lesion-free images of a source are not always from the same anatomical sites as its lesion images (in colonoscopy they are HyperKvasir frames of normal landmarks), which affects lesion-free claims and net utility but not ESG, a comparison of two edits of one image. On 26 of the 288 lesion-free test images at least seven of the nine comparison models reported a finding, so some lesion-free claims may be correct readings.

Every image was assigned to its split before any view was built. Splits were made by patient where the source releases patient identifiers (ChestX-Det, TCGA-LGG) and otherwise by group of near-identical images; because two dissimilar images of one patient can then fall in different splits, an analysis removes the test cases most similar to any training image (Section~\ref{subsec:res_trained}). A hash comparison of 6,849 unedited images found no shared patient or group identifier and one image shared across sets, a validation image that is also a fresh external film (\ref{sec:s_data}). Thresholds, selection rules and fitted quantities used training or validation data only.

The main test holds 507 cases from 416 patients or image groups (219 with a lesion mask, 288 lesion-free) and 1,115 views: 219 each of original, erased and control-edited images, 288 lesion-free images and 170 degraded copies, on which the expected answer is a request for a better image. Development versions of the trained model had been evaluated on the main test and on four external datasets not used in training (169 cases), and the training of \ms{} was changed in response to their weaknesses; their number was not recorded (\ref{sec:s_training}). We therefore added two sets on which no model had been evaluated before \ms{} was fixed: a fresh internal set (76 cases from unused ChestX-Det patients) and a fresh external set (316 cases from four further datasets). Test-set sizes were set by the available cases, not by a power calculation.

\begin{table*}[!t]
\caption{Evaluation sets, with their role, size, source datasets and whether any image of the source was used to train a model of this study (``other patients'': training and test images come from different patients or, without patient identifiers, from different groups of near-identical images). The validation set, the views remade with tools not used in training, the chest X-ray follow-up pairs and the standard question-answering sets are in Supplementary Table~\ref{tab:s_datasets}, with cases per source.}
\label{tab:datasets}
\centering
\footnotesize
\setlength{\tabcolsep}{4pt}
\renewcommand{\arraystretch}{1.08}
\begin{tabular}{@{}L{0.11\textwidth}L{0.17\textwidth}L{0.17\textwidth}L{0.28\textwidth}L{0.18\textwidth}@{}}
\toprule
\textbf{Evaluation set} & \textbf{Role} & \textbf{Size} & \textbf{Source datasets} & \textbf{Source used in training?} \\
\midrule
Main test & Primary comparison & 507 cases (219 with a lesion), 1,115 views & SLAKE \citep{liu2021slake}; Kvasir-SEG with HyperKvasir \citep{jha2020kvasir,borgli2020hyperkvasir}; ChestX-Det \citep{liu2020chestxdet}; BUSI \citep{al2020dataset}; TCGA-LGG \citep{buda2019association} & Yes (other patients) \\
External datasets & Datasets not used in training & 169 cases (119 with a lesion) & CVC-ClinicDB \citep{bernal2015cvc}; QaTa-COV19 \citep{degerli2022qata}; BUS-BRA \citep{gomez2024busbra}; BraTS-Africa 2023 \citep{adewole2023bratsafrica} & No \\
Fresh internal & No model evaluated on it before \ms{} was fixed & 76 cases (36 with a lesion) & ChestX-Det \citep{liu2020chestxdet} & Yes (other patients) \\
Fresh external & No model evaluated on it before \ms{} was fixed & 316 cases & BraTS 2023 meningioma \citep{labella2023meningioma}; BUS-UCLM \citep{vallez2025busuclm}; SIIM-ACR pneumothorax (256-pixel copies) \citep{siim2019pneumothorax}; ETIS-Larib \citep{silva2014etis} & \ms: no. \msp{} and \msn: SIIM-ACR (other patients) \\
Colonoscopy pairs & Unedited same-patient pairs & REAL-Colon: 103 pairs from 46 videos, four centres; LDPolypVideo: 116 pairs from 160 clips, one hospital & REAL-Colon \citep{biffi2024realcolon}; LDPolypVideo \citep{ma2021ldpolyp} & REAL-Colon: \msn{} only, answers cross-validated by video. LDPolypVideo: no \\
Chest films & Pneumothorax on unedited films & 200 new SIIM-ACR patients (100 with pneumothorax); 600 CheXthought test films (300 with pneumothorax) & SIIM-ACR \citep{siim2019pneumothorax}; CheXthought \citep{sharma2026chexthought} on CheXpert Plus images \citep{chambon2024chexpertplus} & SIIM-ACR: \msp{} and \msn. CheXthought: \msn{} (other patients) \\
\bottomrule
\end{tabular}
\end{table*}

\subsection{Erased, control-edited and lesion-free views}
\label{subsec:views}

The erased image is the original with the lesion mask, dilated by 5 pixels, filled by the inpainting model LaMa \citep{suvorov2022lama}. The control edit leaves the lesion untouched and fills a region of the same shape, with the same tool, in tissue elsewhere, so no lesion pixel changes; it matches the erasure in every property except the removal of the lesion region. Lesion-free images are unedited and carry the same expected answer as an erased image, so a model that answers ``absent'' only when it detects an edit fails on them.

Training used two editing tools, LaMa and tissue transplant (tissue from the same image blended over the lesion). TELEA \citep{telea2004image} and Stable Diffusion inpainting \citep{rombach2022high} were not used in training, and the main-test erasures and control edits were remade with each on the same masks. An \emph{edit-site control} on validation images fills a ring around the intact lesion, starting 2 pixels outside it; the same analysis erased half of the mask, or pasted the lesion pixels back into the erased image.

Edited views passed automated checks and were not selected by hand (\ref{sec:s_views}). For the blank-image control, 183 main-test lesion cases were shown as a uniform grey image with the same question. Erasures that a trained edit detector scored no higher than the median control edit are called \emph{low-visibility}. Questions are \emph{finding-specific} when they name a finding (114 lesion cases) and \emph{generic} otherwise (105).

\subsection{Outcome measures}
\label{subsec:measures}

The scoring code read each final answer from its action tag (a reply without one was read as a plain answer; \ref{sec:s_prompts}) and classed it as \emph{present}, \emph{absent}, \emph{refer} (the image is usable but the model does not decide) or a request for a better image. A lesion is \emph{denied} when the answer is absent or refer. The primary measure is the evidence sensitivity gap,
\begin{equation}
\label{eq:esg}
\mathrm{ESG} = \Pr(\text{denied} \mid \text{erased}) - \Pr(\text{denied} \mid \text{control edit}).
\end{equation}
ESG is zero for any constant answer and for any model that responds to the two edits alike. It is a Youden index for the paired contrast: the share of erased images denied plus the share of control-edited images not denied, minus one.

\emph{Erased-lesion claims} are the share of erased images answered present, and \emph{lesion-free claims} the share of unedited lesion-free images answered present. \emph{Lesion images correct} is the share of original and control-edited images answered correctly.

\emph{Net utility} is the mean reward per view under a cost matrix that we wrote and did not elicit from clinicians (Supplementary Table~\ref{tab:s_cost}): a correct answer earns $+1$ ($+0.25$ on a degraded copy), a wrong answer $-2$ ($-4$ on a lesion image whose finding we marked in advance as critical), a referral $+0.5$ on erased and lesion-free images and $-0.3$ on lesion images, and a request for a better image $+0.5$ on a degraded copy and $-0.3$ or $-0.5$ elsewhere. A correct box on a lesion image adds $+0.5$, each tool call beyond the first costs $0.05$, an unreadable output scores $-1$, and erased and lesion-free images are scored alike, so detecting an edit earns nothing. \ms{} was trained with this reward (Section~\ref{subsec:models}) and the comparison models were not shown it, so net utility favours the trained model by construction; ESG and the claim rates do not depend on the matrix.

For unedited pairs, the \emph{natural ESG} is the share of lesion-free images denied minus the share of lesion images of the same patients denied, so a request for a better image on a lesion image counts as not denied. For chest films we report the \emph{Youden index} \citep{youden1950index}, the share of pneumothorax films answered present plus the share of films without pneumothorax denied, minus one, which counts that request as a miss. The reference standard is the expert mask or label of the source dataset; for an erased image it is absence by construction, which the readers assess. The primary outcomes, fixed before \ms{} was trained, are ESG, erased-lesion claims and net utility on the main test.

\subsection{Reader study}
\label{subsec:readers}

The specialist round is the primary reader assessment (protocol in \ref{sec:s_readers}). \emph{Images:} 212 main-test images from different cases, 36 of each kind: unedited lesion and lesion-free images, and erasures and control edits made with LaMa and with Stable Diffusion (32 Stable Diffusion control edits). \emph{Readers:} board-certified radiologists with at least five years of practice after certification, and a board-certified gastroenterologist for colonoscopy images; none had read in the earlier round. \emph{Independence:} each reader worked alone, without labels, kinds of edit, model outputs or other readers' answers; readers who complete the reading are offered authorship, and the gastroenterologist is an author. \emph{Presentation:} one image at a time with its clinical question, in a shuffled order; readers were told only that the study concerns automatically edited images. \emph{Questions:} whether the image had been digitally edited, and the clinical question (yes, no or cannot tell). \emph{Criterion:} for each editing tool, the erasures are accepted when the share of erased images that the reader majority calls present has an exact 95\% upper limit of at most 10\%, which with 36 erased images per tool is possible only if none is called present; with two radiologists, a majority means both. The gastroenterologist's answers count for colonoscopy only. The round closes when two radiologists have read all images or on a fixed date. \emph{Re-scoring:} the stored model answers were scored on the LaMa erasures that no specialist for the imaging type called present (Stable Diffusion views are not in the main test); the persistence is taken to hold if the nine comparison models together have a 95\% lower bound above 20\% on these images, and with fewer than 10 such images the result is descriptive. This rule was written before the re-scoring was run, after the first radiologist had finished.

An earlier round used two trainee physicians (internal medicine and radiology; one is an author), 128 main-test images (32 each of lesion images, LaMa erasures, LaMa control edits and lesion-free images) and the same questions and criterion; the model answers were then re-scored on the erasures that neither trainee called present.

\subsection{Models}
\label{subsec:models}

Every model answered under one protocol (\ref{sec:s_prompts}, with access dates and settings): the image, with its long side capped at 768 pixels, the question and the same system prompt, which addresses the model as MedScope. It could make up to three tool calls (enlarge a region, adjust contrast) and ended with an answer (with an optional box), a referral or a request for a better image. A failed request was repeated up to six times and then scored as unreadable. On the main test, unreadable answers were $0.0\%$ for \ms, \msn{} and six of the nine comparison models and at most $0.4\%$ for the other three; among open-weight models they reached $3.9\%$ (MedGemma 1.5 4B).

{\sloppy Nine comparison models were called through their providers' interfaces: \texttt{gpt-6-sol}, \gpt{} and \texttt{gpt-6.1-sol} (OpenAI); \texttt{claude-fable-5-1}, \opus{} and \texttt{claude-sonnet-5-5} (Anthropic); and \gem, \texttt{gemini-3.1-pro-preview} and \texttt{gemini-3.8-flash} (Google). Each was run once at its provider's default settings, so run-to-run variation was not measured. Analyses not run on all nine used one model per provider, \gem, \opus{} and \gpt, the three reference models. They were chosen after the nine had been run on the main test, and the criterion was not recorded; the set contains the comparison model with the highest main-test ESG (\gem), and the other two were not their providers' highest (Table~\ref{tab:main}). Six open-weight models, among them \base{} without fine-tuning, were run locally with greedy decoding.\par}

Table~\ref{tab:models} gives the three trained models named in the text; the suffixes stand for pneumothorax films added (-P) and natural, that is unedited, images added (-N). \ms{} is \base, an open-weight vision-language model, with a low-rank adapter (rank 16) \citep{hu2022lora} on the language model and a frozen vision encoder. It was trained on examples built automatically from the dataset labels, in three stages: supervised fine-tuning on 16,000 original, edited, lesion-free and degraded views; fine-tuning on its own highest-scoring sampled answers under the net-utility reward; and fine-tuning on 5,000 views mixed with questions from three public question-answering datasets \citep{lau2018vqarad,liu2021slake,he2020pathvqa}. A \emph{first-stage model} is a model after the first stage only. A rule fixed in advance chose among the stages on validation data, and the chosen model was run once on each test set. Ablations trained first-stage models without counterfactual views, with the share of ``absent'' answers matched by lesion-free images, and on MedGemma 1.5 4B \citep{sellergren2025medgemma}.

\begin{table}[!t]
\caption{The three trained models named in the text. Each is described by the question it tests and what its training adds; \ms{} is the primary model. Variants tested on the way to \msn{} are in Supplementary Table~\ref{tab:s_models}.}
\label{tab:models}
\centering
\footnotesize
\setlength{\tabcolsep}{3pt}
\renewcommand{\arraystretch}{1.1}
\begin{tabular}{@{}L{0.23\columnwidth}L{0.44\columnwidth}L{0.27\columnwidth}@{}}
\toprule
\textbf{Model} & \textbf{Question; what its training adds} & \textbf{Status} \\
\midrule
\ms & What does training on counterfactual views change? \base{} with an adapter, trained in three stages on erased, control-edited, lesion-free and degraded views & Primary; training, selection rule and tests fixed before its training data were built \\
\addlinespace
\msp & Do pneumothorax films restore pneumothorax detection? Films of 150 patients with and 150 without pneumothorax (SIIM-ACR) added to every stage & Post hoc; validation rule fixed before its test results \\
\addlinespace
\msn & Do unedited training images bring detection on unedited images? \msp{} with 1,727 polyp and 1,881 polyp-free REAL-Colon frames, and 1,500 CheXthought films with and 1,500 without pneumothorax, added to the last stage & Post hoc; training mix and tests fixed before it was trained \\
\bottomrule
\end{tabular}
\end{table}

\begin{table*}[!t]
\caption{Main test: 507 cases and 1{,}115 views, every model under the same protocol. Lesion images correct: share of the 438 original and control-edited images answered correctly. Erased-lesion claims: share of the 219 erased images on which the finding was reported. Lesion-free claims: share of the 288 unedited lesion-free images on which a finding was reported. ESG: evidence sensitivity gap. Net utility: mean reward per view under the cost matrix of Section~\ref{subsec:measures}. Brackets are 95\% bootstrap intervals over cases (2{,}000 resamples); they were computed for \ms{} only. Models are ordered by ESG within each group. A constant strategy gives the same answer to every view.}
\label{tab:main}
\centering
\small
\begin{tabular}{lccccc}
\toprule
Model & \parbox[b]{2.2cm}{\centering Lesion images correct (\%)} & \parbox[b]{2.6cm}{\centering Erased-lesion claims (\%)} & \parbox[b]{2.6cm}{\centering Lesion-free claims (\%)} & ESG & Net utility \\
\midrule
\multicolumn{6}{l}{\textit{Comparison models (nine)}} \\
\gem & $87.4$ & $41.1$ & $9.0$ & $0.470$ & $0.565$ \\
\texttt{claude-sonnet-5-5} & $87.0$ & $48.4$ & $16.7$ & $0.402$ & $0.381$ \\
\texttt{gemini-3.1-pro-preview} & $84.9$ & $51.6$ & $26.7$ & $0.384$ & $0.306$ \\
\texttt{gpt-6-sol} & $82.9$ & $51.6$ & $25.7$ & $0.342$ & $0.261$ \\
\texttt{claude-fable-5-1} & $85.6$ & $52.1$ & $23.3$ & $0.342$ & $0.255$ \\
\opus & $86.3$ & $53.9$ & $23.6$ & $0.329$ & $0.358$ \\
\texttt{gpt-6.1-sol} & $88.8$ & $58.9$ & $30.9$ & $0.324$ & $0.276$ \\
\gpt & $88.6$ & $58.9$ & $27.8$ & $0.306$ & $0.309$ \\
\texttt{gemini-3.8-flash} & $86.8$ & $61.6$ & $27.8$ & $0.269$ & $0.263$ \\
\midrule
\multicolumn{6}{l}{\textit{Trained on counterfactual views (Section~\ref{subsec:res_trained})}} \\
\ms & $84.9$ & $9.6$ & $7.6$ & $0.758$ & $0.732$ \\
 & & $\ci{6.1}{13.7}$ & $\ci{4.8}{10.9}$ & $\ci{0.701}{0.815}$ & $\ci{0.649}{0.813}$ \\
\midrule
\multicolumn{6}{l}{\textit{Open-weight models (six), not fine-tuned}} \\
Lingshu-7B & $52.7$ & $32.0$ & $12.5$ & $0.416$ & $-0.249$ \\
\base{} (base of \ms) & $75.8$ & $42.0$ & $19.4$ & $0.338$ & $0.134$ \\
Qwen3-VL-8B & $55.3$ & $12.8$ & $12.5$ & $0.297$ & $0.186$ \\
MedGemma 1.5 4B & $67.4$ & $54.3$ & $24.0$ & $0.283$ & $-0.255$ \\
HuatuoGPT-Vision-7B & $49.8$ & $30.1$ & $22.9$ & $0.265$ & $-0.426$ \\
Qwen2.5-VL-7B & $14.4$ & $5.5$ & $24.7$ & $0.146$ & $-0.742$ \\
\midrule
\multicolumn{6}{l}{\textit{Constant strategies}} \\
Always reports the finding & $100.0$ & $100.0$ & $100.0$ & $0.000$ & $-0.478$ \\
Always answers ``absent'' & $0.0$ & $0.0$ & $0.0$ & $0.000$ & $-0.848$ \\
Always refers & $0.0$ & $0.0$ & $0.0$ & $0.000$ & $0.110$ \\
\bottomrule
\end{tabular}
\end{table*}

\msp{} repeats the three stages with the same data, settings and seed, with films of 150 patients with and 150 without pneumothorax from SIIM-ACR, none of them in a test set, added to every stage. \msn{} was designed after the result of \ms{} on unedited pairs was known and after each of its two additions had been tested alone (Supplementary Table~\ref{tab:s_models}); its training mix was chosen on validation data, and its tests were fixed before it was trained (Supplementary Tables~\ref{tab:s_plan} and~\ref{tab:s_mix}). Because REAL-Colon serves in its training and testing, its REAL-Colon answers are \emph{cross-validated}: three copies were each trained without one third of the videos, and each pair is answered by the copy trained without its video.

The training of \ms{} and of \msn{} was repeated with two further seeds; the first seed is the reported model. For \msn{} the pre-specified rule made a three-seed majority vote (three model runs per image) the reported operating point; in a deviation from the plan we report the first seed instead, because the confirmatory test at the new hospital was specified for it, and give the vote beside it. Seventeen alternative designs and 270 decision rules were each tested against a criterion fixed in advance (Supplementary Table~\ref{tab:s_plan}).

\subsection{Unedited image pairs}
\label{subsec:pairs}

REAL-Colon \citep{biffi2024realcolon} holds 60 colonoscopy videos from four centres, with a box on every frame that shows a resected polyp. For each of a video's three longest runs of boxed frames, a polyp frame from the middle of the run was paired with a polyp-free frame (no box marked) of the same video, 10 to 30~s from the run and at least 10~s from any boxed frame (\ref{sec:s_views}). Both frames received the question ``Is there a polyp in this image?'', which gave 103 pairs from 46 videos; 96 pairs also received a \emph{context-matched} polyp-free frame, 3 to 10~s from the nearest boxed frame. LDPolypVideo \citep{ma2021ldpolyp} holds 160 annotated clips from one hospital on which no model of this study had been trained, selected or tested; because its clips state no frame rate, the polyp-free frame had to lie at least 5 frames from any polyp frame, a much shorter distance than at REAL-Colon, and 116 clips gave a pair.

Chest films were not paired by patient: pneumothorax detection was tested on the 200 new SIIM-ACR patients \citep{siim2019pneumothorax} and on the 600 CheXthought test films \citep{sharma2026chexthought}, whose labels combine the CheXpert Plus label \citep{chambon2024chexpertplus} with our rule-based reading of radiologists' statements. Same-patient follow-up pairs from NIH ChestX-ray14 \citep{wang2017chestx} are a secondary analysis (\ref{sec:s_views}).

\subsection{Analysis plan and statistics}
\label{subsec:stats}

The analysis plan was written during the study, each rule after the results that preceded it. It is our own dated record in the public project repository (\url{https://github.com/bhaskargurram-ai/medscope}), not a third-party registry; Supplementary Tables~\ref{tab:s_overview} and~\ref{tab:s_plan} give every analysis with its date, status, rule and outcome. An analysis is \emph{pre-specified} when its test set, comparison and decision rule were fixed and dated before the data it applies to were seen, \emph{descriptive} when it was fixed in advance without a decision rule, and \emph{post hoc} otherwise. Confirmatory analyses, in which no model of this study had been run on the test data when the rule was dated, were \ms{} on the fresh internal and fresh external sets and on the REAL-Colon pairs, \msn{} on the LDPolypVideo pairs, the first test on the new SIIM-ACR patients and the two reader rounds. Other analyses with a dated rule are secondary, and four of them re-analyse main-test answers whose pooled results were known. Post hoc analyses, among them the models \msp{} and \msn{} and the family of 27 primary comparisons, are marked where reported.

\emph{Primary comparisons:} \ms{} against each of the nine comparison models on the three primary outcomes, 27 comparisons, each tested at a Bonferroni-corrected level of $0.05/27$. \emph{Other comparisons:} two-sided at the 0.05 level, uncorrected, read with their intervals. \emph{Intervals:} 95\% confidence intervals (CI) by percentile bootstrap from 2,000 resamples. \emph{Resampled unit:} the case on the edited test sets (all views of a case together), the video or clip for colonoscopy pairs and the patient for chest films (patients or image groups in a sensitivity analysis; \ref{sec:s_stats}). \emph{Comparisons between models:} paired, with both models scored on the same resampled units; the $p$ value is twice the smaller share of resampled differences on either side of zero. With 2,000 resamples the smallest non-zero $p$ value is 0.001, so the primary conclusions rest on the paired intervals. \emph{Reader proportions:} exact binomial intervals. \emph{What an interval reflects:} the sampling of test cases for one trained model and one run of each comparison model. Variation between training seeds is reported separately, as the mean and standard deviation (SD) over three seeds, and the two are never combined. \emph{Software:} Python with NumPy for the analyses; PyTorch 2.11, Transformers 5.22.1 and PEFT 0.21.2 for \ms; PyTorch 2.14.1, Transformers 5.24.1 and PEFT 0.22.0 for the final runs of \msn.

\section{Results}\label{sec:results}

\begin{figure*}[!t]
\centering
\includegraphics[width=\textwidth]{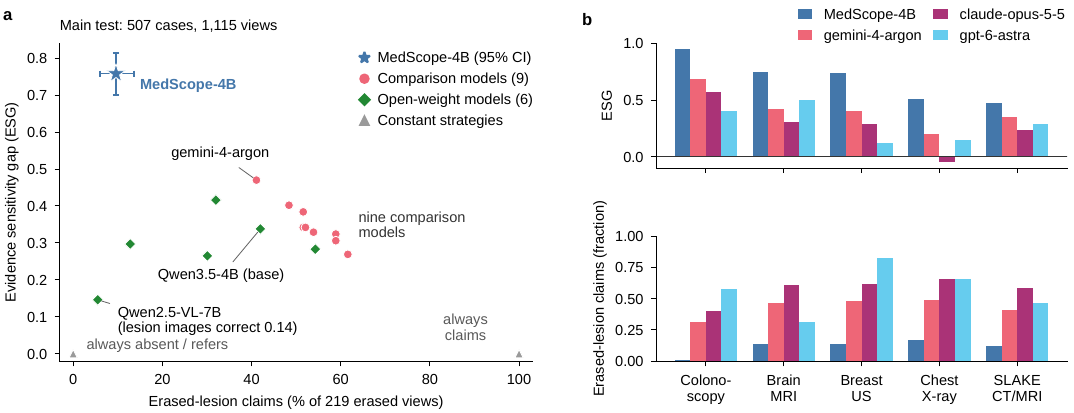}
\caption{Main test (507 cases, 1{,}115 views). (a) Evidence sensitivity gap (ESG) against erased-lesion claims (share of the 219 erased images on which the finding was reported) for the nine comparison models, the six open-weight models, \ms{} and the three constant strategies. Bars on \ms{} are 95\% bootstrap intervals over cases (2{,}000 resamples). (b) ESG (top) and erased-lesion claims (bottom, as a fraction of erased images) by imaging type for \ms{} and the three reference models. The strata hold 57 to 174 cases; intervals were not computed for them. Values are in Table~\ref{tab:main} and Supplementary Table~\ref{tab:s_modality}. CT, computed tomography; MRI, magnetic resonance imaging; US, ultrasound.}
\label{fig:main}
\end{figure*}

Brackets give 95\% bootstrap intervals over cases; differences are paired by case.

\subsection{Answers of the comparison models often persisted after erasure}\label{subsec:res_main}

On the main test (507 cases, 1{,}115 views), the nine comparison models reported the finding on $41.1\%$ (\gem) to $61.6\%$ (\texttt{gemini-3.8-flash}) of the 219 images from which the lesion had been erased, with ESG between $0.269$ and $0.470$ (Table~\ref{tab:main}, Fig.~\ref{fig:main}a).
They were correct on $82.9\%$ to $88.8\%$ of lesion images, so models that differed by less than 6 percentage points in lesion images correct differed by about 20 percentage points in erased-lesion claims.
The share of erased lesions reported is not a clinical error rate: an erased image is an edited image, and whether readers accept it as lesion-free is the subject of Section~\ref{subsec:res_readers}.
On the 288 unedited lesion-free images the comparison models reported a finding on $9.0\%$ to $30.9\%$.

The question alone did not produce these reports.
When the image of 183 lesion cases was replaced by a uniform grey image, none of the nine models reported the finding, and each asked for a better image.
On the erased images of the same cases they reported it on $42.1\%$ to $56.8\%$ (all $p < 0.001$ against the blank image; Supplementary Table~\ref{tab:s_blank}).
What in the erased image led to the reports (surrounding tissue, the instrument, residual lesion or the edit) is examined in Sections~\ref{subsec:res_readers} and~\ref{subsec:res_alternatives}.

\begin{table*}[!t]
\caption{Reader assessment of main-test images. Readers saw one image and its question at a time, in random order, without being told which images were edited, and answered whether the finding was present and whether the image looked edited (Section~\ref{subsec:readers}). Entries are numbers of images on which the reader called the finding present; parentheses give the percentage and its exact (Clopper--Pearson) 95\% interval. Each specialist's entries are for the images of the specialty: colonoscopy for the gastroenterologist, and brain MRI, breast ultrasound, chest X-ray and SLAKE for the radiologist. The trainee round used LaMa erasures only. AUC: area under the receiver operating characteristic curve for separating edited from unedited images by the reader's answers.}
\label{tab:readers}
\centering
\small
\setlength{\tabcolsep}{5pt}
\begin{tabular}{L{2.6cm}L{4.0cm}L{4.0cm}L{1.8cm}L{1.8cm}L{1.8cm}}
\toprule
 & \multicolumn{2}{l}{Erased images called present} & Lesion-free images called present & Unedited lesion images called present & Edited against unedited, AUC \\
\cmidrule(lr){2-3}
Reader & LaMa & Stable Diffusion & & & \\
\midrule
\multicolumn{6}{l}{\textit{Specialist round (212 images read by each reader; 61 colonoscopy and 151 radiology images)}} \\
Radiologist & 13 of 29 ($44.8\%$; $26.4$ to $64.3$) & 16 of 26 ($61.5\%$; $40.6$ to $79.8$) & 5 of 29 & 25 of 27 & $0.64$ \\
Gastroenterologist & 3 of 7 ($42.9\%$; $9.9$ to $81.6$) & 1 of 10 ($10.0\%$; $0.3$ to $44.5$) & 0 of 7 & 8 of 9 & $0.68$ \\
\quad both specialists & 16 of 36 ($44.4\%$) & 17 of 36 ($47.2\%$) & 5 of 36 & 33 of 36 & \\
\midrule
\multicolumn{6}{l}{\textit{Earlier round (128 images read, all imaging types)}} \\
Trainee, internal medicine & 19 of 32 & not read & 16 of 32 & 29 of 32 & $0.54$ \\
Trainee, radiology & 16 of 32 & not read & 7 of 32 & 26 of 32 & $0.55$ \\
\quad both trainees & 13 of 32 ($40.6\%$; $23.7$ to $59.4$) & & & & \\
\bottomrule
\end{tabular}
\end{table*}

On the 114 cases whose question names the finding, the comparison models reported it on $37.7\%$ to $53.5\%$ of erased images (ESG $0.298$ to $0.439$).
On the 105 cases with a generic question (``Is there an abnormality?''), for which ``yes'' on an erased image is correct if another abnormality remains, they reported it on $44.8\%$ to $70.5\%$ (ESG $0.238$ to $0.505$).
In colonoscopy they reported $45\%$ of the 83 erased polyps: on average $26.9\%$ of the 24 asked about with a polyp-specific question and $52.4\%$ of the 59 asked about with a generic one, about twice as often (Supplementary Table~\ref{tab:s_qtype}).

The persistence was general across models, sources and prompts.
Each of the nine comparison models reported at least $41.1\%$ of erased lesions, and each of the three reference models at least $31\%$ in every imaging type (Fig.~\ref{fig:main}b; Supplementary Table~\ref{tab:s_modality}).
On 150 main-test cases the reference models reported $36.8\%$ to $66.2\%$ of erased lesions under every prompting condition: the default, a cautionary prompt, solved examples, the highest reasoning setting, the highest image resolution, and a plain prompt without tools, the last two added post hoc (ESG $0.250$ to $0.500$; Supplementary Table~\ref{tab:s_prompts}).

\subsection{Reader assessment of the erased images}\label{subsec:res_readers}

Two board-certified specialists read all 212 images of the specialist round without labels: a radiologist (8 years of practice) for brain MRI, breast ultrasound, chest X-ray and SLAKE images, and a gastroenterologist (20 years) for colonoscopy (Table~\ref{tab:readers}).
Further radiologists are reading, and the results below are interim.
Of the 36 erasures made with LaMa, the tool of the main test, the specialist for the imaging type accepted 20 as free of the finding (15 with an explicit ``no'') and called the finding present on 16.
On the 20 accepted erasures the nine comparison models still reported the finding on $43.3\%$ (95\% CI $31.7$ to $56.1$; 5 to 12 of 20 per model), and on $37.8\%$ ($23.7$ to $53.3$) of the 15 with an explicit ``no'' (Supplementary Table~\ref{tab:s_specialist}).
\ms{} and \msn{} each reported it on 1 of 20, $38.3$ percentage points less often than the comparison models on the same images (95\% CI $22.8$ to $52.8$), and found 59 and 61 of the 66 lesions that the specialists confirmed.
The lower bound of $31.7\%$ is above the $20\%$ set for this analysis, the bound used for the trainee round, so the persistence holds on erasures that specialists accepted.

The specialists did not accept every erasure.
The radiologist called the finding present on 13 of 29 LaMa erasures ($44.8\%$; exact 95\% CI $26.4$ to $64.3$) and 16 of 26 Stable Diffusion erasures ($61.5\%$; $40.6$ to $79.8$) of radiology images, and on 5 of 29 lesion-free images; the gastroenterologist on 3 of 7 LaMa and 1 of 10 Stable Diffusion erasures of colonoscopy images, and on 0 of 7 lesion-free images.
Neither separated edited from unedited images well (area under the receiver operating characteristic curve, AUC, $0.64$ and $0.68$; for the radiologist, erasures against control edits $0.53$).
The pre-specified criterion (upper limit at most $10\%$ for each tool) was not met.
On the 16 erasures a specialist called present, the comparison models reported the finding on $64.6\%$ ($44.4$ to $83.4$) and \ms{} on 3.

In an earlier round, two trainee physicians each read 128 main-test images and did not separate edited from unedited images (AUC $0.54$ and $0.55$).
They called a finding present on 35 of 64 reads of erased images ($54.7\%$) and on 23 of 64 reads of lesion-free images ($35.9\%$).
Both did so on 13 of 32 erased images ($40.6\%$; exact 95\% CI $23.7$ to $59.4$), so the pre-specified criterion was not met.
The rates on erased and on lesion-free images were close in colonoscopy (3 of 12 against 2 of 12 reads) and in SLAKE CT and MRI (5 of 12 against 6 of 12, where half of the lesion-free reads were also called present); in the other three sources the erased rate was higher (Supplementary Table~\ref{tab:s_trainee}).

Where both trainees called the finding present (13 erased images), the nine comparison models together reported it on $82\%$ of images; where neither did (10 images), on $26\%$ (95\% CI $10\%$ to $42\%$), and the pre-specified lower bound of $20\%$ was not met (Supplementary Table~\ref{tab:s_rescoring}).
Part of the erased-lesion claims therefore fell on images that readers did not accept as lesion-free, and part on images that the specialists did accept.
Two of the four readers who have finished are co-authors; all read without labels.

\subsection{A model trained on counterfactual views}\label{subsec:res_trained}

\begin{figure*}[!t]
\centering
\includegraphics[width=\textwidth]{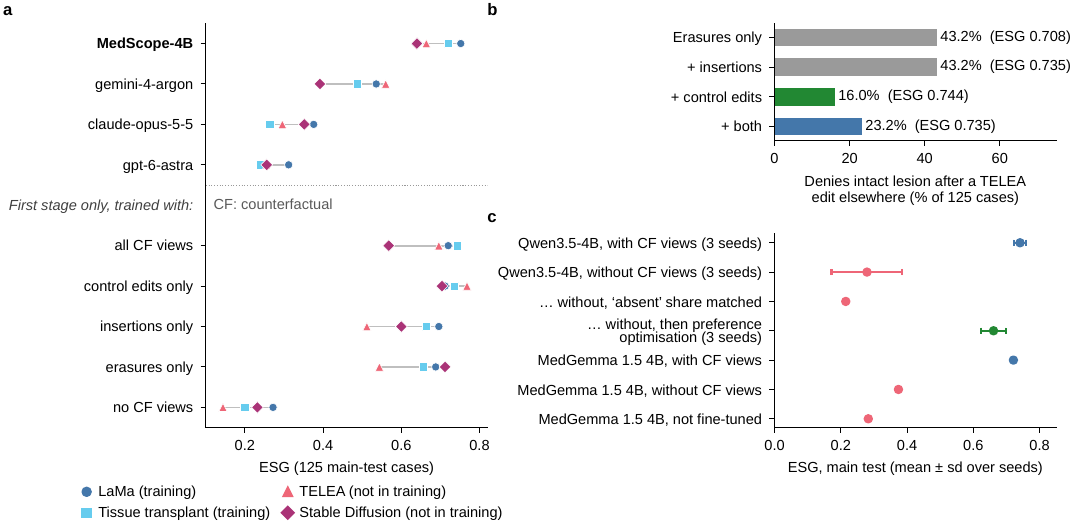}
\caption{Editing tools and training views. (a) Evidence sensitivity gap (ESG) on 125 main-test cases whose erased and control-edited images were remade on the same masks with four editing tools, two used in training (LaMa, tissue transplant) and two not used (TELEA, Stable Diffusion): \ms, the three reference models and, below the dotted line, five first-stage models (after the first of three training stages). (b) Share of 125 intact lesions denied (answered ``absent'' or referred) after a TELEA control edit by first-stage models trained with erasures only, with lesion insertions added, with control edits added, and with both (the views used for \ms; the two middle arms were added post hoc); main-test ESG in parentheses. Grey bars: no control edits in training; green: control edits added; blue: the views used for \ms. (c) Main-test ESG (507 cases) of first-stage models trained with and without counterfactual (CF) views on \base{} (mean and standard deviation, SD, over three seeds where marked) and on MedGemma 1.5 4B. ``Absent share matched'': no CF views, with the share of ``absent'' answers matched by lesion-free images. ``Then preference optimisation'': no CF views, followed by preference optimisation on erased images (Section~\ref{subsec:res_designs}). Blue: models with CF views; red: without them; green: without them, followed by preference optimisation. Labels that begin with an ellipsis continue ``\base, without CF views''. In (a) and (b), rows named after one kind of view denote erasures plus that kind. Intervals were not computed for (a) and (b). Values are in Supplementary Tables~\ref{tab:s_tools} and~\ref{tab:s_ablations}.}
\label{fig:tools}
\end{figure*}

On the main test \ms{} reported the finding on $9.6\%$ $\ci{6.1}{13.7}$ of the 219 erased images and had an ESG of $0.758$ $\ci{0.701}{0.815}$ (Table~\ref{tab:main}, Fig.~\ref{fig:main}a).
Its ESG exceeded that of each comparison model by $0.288$ $\ci{0.208}{0.367}$ (\gem) to $0.489$, and its erased-lesion claims were lower by $31.5$ to $52.1$ percentage points.
On the main test this result did not come from answering ``absent'' more often: \ms{} was correct on $84.9\%$ of lesion images, against $82.9\%$ to $88.8\%$ for the comparison models (differences $-3.9$ to $+2.1$ percentage points, all $p \ge 0.12$).
The lower limits reach $-9$ percentage points, so a small deficit is not excluded.
On the 288 unedited lesion-free images it reported a finding on $7.6\%$ $\ci{4.8}{10.9}$, a lower share than eight comparison models and not shown to differ from \gem{} ($9.0\%$; $p = 0.59$).
Net utility, a measure that favours the trained model (Section~\ref{subsec:measures}), was higher by $0.168$ to $0.477$.
All 27 primary comparisons had $p < 0.001$, below the corrected level of $0.05/27$, and every interval excluded zero; the smallest differences were those from \gem: ESG $0.288$ $\ci{0.208}{0.367}$, erased-lesion claims $-31.5$ percentage points $\ci{-38.5}{-24.3}$ and net utility $0.168$ $\ci{0.083}{0.254}$ (Supplementary Table~\ref{tab:s_primary}).

The advantage in ESG and erased-lesion claims held in point estimate against each reference model in each of the five sources (ESG from $0.95$ in colonoscopy to $0.47$ in SLAKE CT and MRI; Fig.~\ref{fig:main}b), although in SLAKE CT and MRI its net utility ($0.22$) was below that of \gem{} ($0.38$) and \opus{} ($0.32$).
On the 114 cases whose question names the finding, the ESG of \ms{} was $0.728$ $\ci{0.649}{0.807}$, against $0.298$ to $0.439$ for the nine comparison models (smallest difference $0.289$ $\ci{0.175}{0.412}$), and the pre-specified criterion was met.

\begin{table*}[!t]
\caption{Analyses of the alternative explanation that \ms{} responds to the edit and not to the absence of the lesion. Each row gives what a model that responds only to edit traces would be expected to do and what was observed. Rows 1, 2 and 5 use \ms{} on the main test; rows 3 and 4 use the first-stage model on validation cases whose lesion it found on the original image. Denied: answered ``absent'' or referred. Brackets are 95\% bootstrap intervals over cases. Reference models are given in the order \gem, \opus, \gpt. ESG: evidence sensitivity gap.}
\label{tab:alternatives}
\centering
\small
\setlength{\tabcolsep}{4pt}
\begin{tabular}{L{4.8cm}cL{3.3cm}L{4.2cm}L{3.6cm}}
\toprule
Analysis & $n$ & Expected of a model that responds only to the edit & Observed & Comparison \\
\midrule
1. Control-edited images: lesion denied & 219 & $100\%$ (ESG $0$) & $14.6\%^{a}$; ESG $0.758$ $\ci{0.701}{0.815}$ & nine comparison models: ESG $0.269$ to $0.470$ \\
\addlinespace
2. Tools not used in training: ESG with TELEA; with Stable Diffusion & 125 & no advantage & $0.664$; $0.640$ & reference models, TELEA: $0.560$, $0.296$, $0.256$; Stable Diffusion: $0.392$, $0.352$, $0.256$ \\
\addlinespace
3. Lesion pixels pasted back into the erased image (LaMa): lesion denied & 143 & as after erasure ($86.7\%$) & $7.0\%$ (half of the mask erased: $19.6\%$) & distant control edit: $4.2\%$ \\
\addlinespace
4. Edit-site control (TELEA): intact lesion denied after a ring edit at the lesion & 143 & more often than after a distant edit & $20.3\%$ against $12.6\%$; difference $7.7$ percentage points $\ci{0.0}{15.4}$ & base model: $11.6\%$ against $7.0\%$ ($n = 127$) \\
\addlinespace
5. Low-visibility erasures$^{b}$: ESG & 111 & no advantage & $0.640$ $\ci{0.550}{0.721}$ & reference models lower by $0.270$ $\ci{0.153}{0.387}$, $0.450$ $\ci{0.324}{0.568}$, $0.378$ $\ci{0.243}{0.514}$ \\
\bottomrule
\end{tabular}
\\[3pt]\parbox{17.4cm}{\footnotesize $^{a}$\,Obtained from ESG and erased-lesion claims ($90.4\%$ of erased images denied). $^{b}$\,Erasures that a trained edit detector scored no higher than the median control edit.}
\end{table*}

The advantage came from the counterfactual views.
Over three training seeds, models after the first of the three training stages (first-stage models) had a main-test ESG of $0.741$ (SD $0.019$) with counterfactual views and $0.279$ (SD $0.107$) without them (Fig.~\ref{fig:tools}c).
Matching the share of ``absent'' answers without such views did not raise it ($0.215$), and the effect was reproduced on a second base model (MedGemma 1.5 4B: $0.721$ against $0.374$; difference $0.347$ $\ci{0.276}{0.425}$; Supplementary Table~\ref{tab:s_ablations}).
Two further seeds of the complete training gave ESG of $0.744$ and $0.731$, each above each reference model (lesion-free claims $17.0\%$ and $9.0\%$; Supplementary Table~\ref{tab:s_seeds}).

Removing from each source the quarter of test cases most similar to a training image (378 kept) left the ESG of \ms{} at $0.755$ and its advantage over each reference model at least $0.306$ $\ci{0.200}{0.405}$; three sources release no patient identifiers, so dissimilar images of one patient are not excluded (Supplementary Table~\ref{tab:s_duplicates}).

With an open-ended request (36 cases), \ms{} reported the finding on $47.2\%$ of the 36 original images and on none of the 36 erased images (free-text ESG $0.472$ $\ci{0.306}{0.639}$), against $0.306$ to $0.361$ for the reference models.
No difference ($0.111$ to $0.167$) had an interval that excluded zero, so the pre-specified criterion was not met (Supplementary Table~\ref{tab:s_open}).
The ESG of \ms{} rose with lesion area, from $0.545$ in the lowest quartile to $0.909$ in the highest, and its stated confidence ranked its errors less well than that of the reference models (Supplementary Tables~\ref{tab:s_size} and~\ref{tab:s_sensitivity}).

\subsection{Detection of the edit does not account for most of the result}\label{subsec:res_alternatives}

\ms{} was trained on erased and control-edited images built by the same procedure as the test, so it might answer ``absent'' whenever an image looks edited (Table~\ref{tab:alternatives}, Fig.~\ref{fig:tools}).
Edits can be detected: a classifier trained for the purpose separated control-edited from unedited test images with an AUC of $0.826$.

\textit{Matched control edit.}
A model that responds only to the edit has an ESG of zero, because ESG subtracts the share of control-edited images on which the lesion is denied (answered ``absent'' or referred).
\ms{} denied the lesion on $14.6\%$ of the 219 control-edited images and on $90.4\%$ of the 219 erased images.
Control edits also mattered in training.
Among first-stage models, one trained with erasures only denied $43.2\%$ of 125 intact lesions after a control edit made with TELEA, a tool not used in training; with control edits added (a post hoc arm) the figure was $16.0\%$, and with all views $23.2\%$ (Fig.~\ref{fig:tools}b).
The control thus reduced the effect of an unfamiliar tool and did not remove it.

\textit{Amount of lesion removed.}
Answers followed how much of the lesion was removed.
With LaMa, the first-stage model denied $4.2\%$ of 143 validation lesions after a distant control edit, $7.7\%$ after a ring edit at the lesion border, $7.0\%$ when the lesion pixels were pasted back into the erased image, $19.6\%$ when half of the mask was erased and $86.7\%$ when all of it was (Supplementary Table~\ref{tab:s_site}).

\textit{Tools not used in training.}
With the erased and control-edited images of 125 main-test cases remade on the same masks, the ESG of \ms{} was $0.752$ and $0.720$ with the two training tools (LaMa and tissue transplant) and $0.664$ and $0.640$ with TELEA and Stable Diffusion, above each reference model with each tool (Fig.~\ref{fig:tools}a).
The smallest advantage was over \gem{} with TELEA ($0.664$ against $0.560$), and these twelve comparisons are exploratory.
A development version of the model, trained with LaMa edits only, had been run with all four tools, after which tissue transplant was added to training (Supplementary Table~\ref{tab:s_tools}).
With Stable Diffusion \ms{} reported $17.6\%$ of erased lesions, against $7.2\%$ with LaMa ($p = 0.006$).

\textit{Edit-site control.}
Erasures lie on the lesion and control edits away from it, so an edit at the lesion site might itself change the answer.
Like the analysis above, this control was run on the first-stage model and validation cases, not on \ms{} and the test set.
With LaMa a ring-shaped edit around the intact lesion raised denial from $4.2\%$ to $7.7\%$ (difference $3.5$ percentage points $\ci{-2.1}{9.1}$); with TELEA the model denied $20.3\%$ after the ring edit against $12.6\%$ after the distant edit (difference $7.7$ percentage points $\ci{0.0}{15.4}$), which met the pre-specified flag for a site effect.
In a post hoc comparison with \base{} on the 112 cases found by both, the excess of ring over distant edit was $1.8$ percentage points $\ci{-8.0}{11.6}$ larger for the trained model; the interval is too wide to exclude a site effect of the size the flag was set for.
Training about doubled denial after a TELEA edit at either site ($12.5\%$ against $6.2\%$ at the ring; $8.9\%$ against $4.5\%$ away from it; Supplementary Tables~\ref{tab:s_site} and~\ref{tab:s_logodds}).

\textit{Erasures that an edit detector scores as unedited.}
An erasure was called low-visibility when a trained edit detector scored it no higher than the median control edit (111 of 219).
On these the ESG of \ms{} was $0.640$ $\ci{0.550}{0.721}$, higher than that of each reference model (all $p < 0.001$; differences in Table~\ref{tab:alternatives}), and the pre-specified criterion was met.
\ms{} reported $16.2\%$ of the low-visibility erasures against $2.8\%$ of the other 108: either some ``absent'' answers follow edit traces or the least visible erasures retain more of the lesion, which this design cannot separate (Supplementary Table~\ref{tab:s_visibility}).

Taken together, the answers of \ms{} followed mainly whether the lesion region had been erased, with some sensitivity to unfamiliar edit traces; the test without any edit is in Section~\ref{subsec:res_pairs}.

\subsection{External datasets and new test sets}\label{subsec:res_external}

\begin{table*}[!t]
\caption{Unedited colonoscopy pairs: a polyp frame and a polyp-free frame of the same video or clip, both asked ``Is there a polyp in this image?''. REAL-Colon: 103 pairs from 46 videos of four centres; the answers of \msn{} are cross-validated (each pair is answered by a copy trained without that pair's video). LDPolypVideo: 116 pairs from 160 clips of one hospital not used in training or model selection. Found: the polyp frame was not answered ``absent'' and was not referred, so a request for a better image counts as found. Absent or referred: the answer to the polyp-free frame. Natural ESG: the share of polyp-free frames answered ``absent'' or referred minus the same share for polyp frames. CI: 95\% confidence interval from $2{,}000$ bootstrap resamples of whole videos or clips. Differences are \msn{} minus the row model, paired over the same pairs, with two-sided $p$ values. \msn{} is the model of the first training seed; the majority vote of its three seeds scored $0.586$ $\ci{0.491}{0.673}$ on LDPolypVideo. The differences of \ms{} from the other models on REAL-Colon are in Supplementary Table~\ref{tab:s_realcolon}. --: not applicable.}
\label{tab:natural}
\centering
\footnotesize
\setlength{\tabcolsep}{3.5pt}
\providecommand{\cistack}[2]{\begin{tabular}[t]{@{}c@{}}#1\\#2\end{tabular}}
\begin{adjustbox}{max width=\textwidth}
\begin{tabular}{l ccccc ccccc}
\toprule
 & \multicolumn{5}{c}{REAL-Colon ($103$ pairs)} & \multicolumn{5}{c}{LDPolypVideo ($116$ pairs)} \\
\cmidrule(lr){2-6}\cmidrule(lr){7-11}
Model & \shortstack{Polyp\\frames\\found} & \shortstack{Polyp-free\\frames absent\\or referred} & \shortstack{Natural ESG\\{}[95\% CI]} & \shortstack{\msn{}\\minus model\\{}[95\% CI]} & $p$ & \shortstack{Polyp\\frames\\found} & \shortstack{Polyp-free\\frames absent\\or referred} & \shortstack{Natural ESG\\{}[95\% CI]} & \shortstack{\msn{}\\minus model\\{}[95\% CI]} & $p$ \\
\midrule
\base{} & $38.8\%$ & $85.4\%$ & \cistack{$0.243$}{$\ci{0.143}{0.350}$} & \cistack{$0.291$}{$\ci{0.167}{0.407}$} & $<0.001$ & $40.5\%$ & $95.7\%$ & \cistack{$0.362$}{$\ci{0.276}{0.448}$} & \cistack{$0.250$}{$\ci{0.155}{0.353}$} & $<0.001$ \\[2pt]
\ms{} & $37.9\%$ & $86.4\%$ & \cistack{$0.243$}{$\ci{0.149}{0.340}$} & \cistack{$0.291$}{$\ci{0.176}{0.402}$} & $<0.001$ & $45.7\%$ & $96.6\%$ & \cistack{$0.422$}{$\ci{0.328}{0.517}$} & \cistack{$0.190$}{$\ci{0.112}{0.267}$} & $<0.001$ \\[2pt]
\msn{} & $64.1\%$ & $89.3\%$ & \cistack{$0.534$}{$\ci{0.394}{0.661}$} & -- & -- & $62.1\%$ & $99.1\%$ & \cistack{$0.612$}{$\ci{0.526}{0.698}$} & -- & -- \\
\midrule
\gem{} & $68.0\%$ & $78.6\%$ & \cistack{$0.466$}{$\ci{0.336}{0.581}$} & \cistack{$0.068$}{$\ci{-0.069}{0.202}$} & $0.347$ & $71.6\%$ & $81.0\%$ & \cistack{$0.526$}{$\ci{0.422}{0.621}$} & \cistack{$0.086$}{$\ci{-0.017}{0.190}$} & $0.117$ \\[2pt]
\opus{} & $70.9\%$ & $73.8\%$ & \cistack{$0.447$}{$\ci{0.317}{0.579}$} & \cistack{$0.087$}{$\ci{-0.049}{0.230}$} & $0.259$ & $62.9\%$ & $83.6\%$ & \cistack{$0.466$}{$\ci{0.362}{0.569}$} & \cistack{$0.147$}{$\ci{0.043}{0.250}$} & $0.010$ \\[2pt]
\gpt{} & $64.1\%$ & $68.0\%$ & \cistack{$0.320$}{$\ci{0.212}{0.437}$} & \cistack{$0.214$}{$\ci{0.071}{0.343}$} & $0.003$ & $55.2\%$ & $75.0\%$ & \cistack{$0.302$}{$\ci{0.198}{0.405}$} & \cistack{$0.310$}{$\ci{0.181}{0.431}$} & $<0.001$ \\
\bottomrule
\end{tabular}
\end{adjustbox}
\end{table*}

On three further test sets the erased-lesion claims of \ms{} were lower than those of each reference model (no test of this measure was recorded on these sets), and its ESG was higher on two of them (Supplementary Table~\ref{tab:s_external}).
On the external datasets (169 cases from four datasets not used in training) it reported $11.8\%$ of erased lesions against $53.8\%$ to $66.4\%$, with an ESG of $0.597$ against $0.277$ to $0.412$ (all $p \le 0.002$).
The fresh internal and fresh external sets had not been used with any model before the training of \ms{} was fixed.
On the fresh internal set (76 cases, 36 with a lesion) its ESG was $0.472$ against $0.000$ to $0.083$ (all $p \le 0.001$).
On the fresh external set (316 cases from four further datasets) it reported $8.3\%$ of erased lesions against $31.5\%$ to $42.3\%$; its ESG ($0.393$) was higher in point estimate than that of \opus{} ($0.292$) and \gpt{} ($0.268$) and not above that of \gem{} ($0.435$).

On the external datasets and the fresh external set \ms{} was correct on fewer lesion images than the reference models: $72.7\%$ against $94.5\%$ to $96.6\%$, and $47.6\%$ against $63.4\%$ (\gpt) to $74.1\%$ (\gem).
Its net utility was not shown to differ on the external datasets ($0.530$ against $0.544$, $0.544$ and $0.424$ for \gem, \opus{} and \gpt) and was below that of \gem{} on the fresh external set ($-0.053$ against $0.310$; difference $-0.363$ $\ci{-0.496}{-0.233}$).

The deficit was largest for pneumothorax and also present in colonoscopy and brain MRI (ETIS-Larib: lesion images correct $69\%$ against $97\%$ for \gem; BraTS meningioma: $53\%$ against $62\%$; Supplementary Table~\ref{tab:s_external}). \ms{} answered ``absent'' on all 48 pneumothorax films of the fresh external set, which holds them as 256-pixel copies (\gem{} found 30), and on all 100 pneumothorax films of the new SIIM-ACR patients.
Its first training stage had held four pneumothorax images and the validation set none; questions naming pneumothorax made up 54 of about 7{,}000 chest X-ray training views of that stage, $80\%$ of them with the answer ``absent'', and the models after the first two stages each found $40\%$ of 42 films at original resolution (post hoc; Supplementary Table~\ref{tab:s_pneumothorax}).
With films of 150 patients with and 150 without pneumothorax added to every stage (\msp, post hoc), 53 of 100 were found and 5 of 100 films without pneumothorax were called present, with a main-test ESG of $0.740$ against $0.758$ (difference $-0.018$ $\ci{-0.073}{0.037}$).

\subsection{Unedited image pairs: natural ESG rose with unedited training images}
\label{subsec:res_pairs}

On 103 unedited REAL-Colon pairs, each a polyp frame and a polyp-free frame of the same video asked the same question (Section~\ref{subsec:pairs}), \ms{} showed no advantage. Its natural ESG was $0.243$ $\ci{0.149}{0.340}$, the same as that of its base model (difference $0.000$ $\ci{-0.103}{0.095}$) and lower than that of \gem{} ($0.466$; $-0.223$ $\ci{-0.346}{-0.102}$) and \opus{} ($0.447$; $-0.204$ $\ci{-0.333}{-0.071}$); the pre-specified criterion was not met (Table~\ref{tab:natural}; Supplementary Table~\ref{tab:s_realcolon}). \ms{} reported a polyp on 2 of 103 polyp-free frames, against 15 for its base model, but it asked for a better image on 12 more, and the shares of polyp-free frames answered ``absent'' or referred were 86.4\% and 85.4\%. It reported the polyp on 36 of 103 polyp frames, against 40 for its base model.

\begin{figure*}[!t]
\centering
\includegraphics[width=\textwidth]{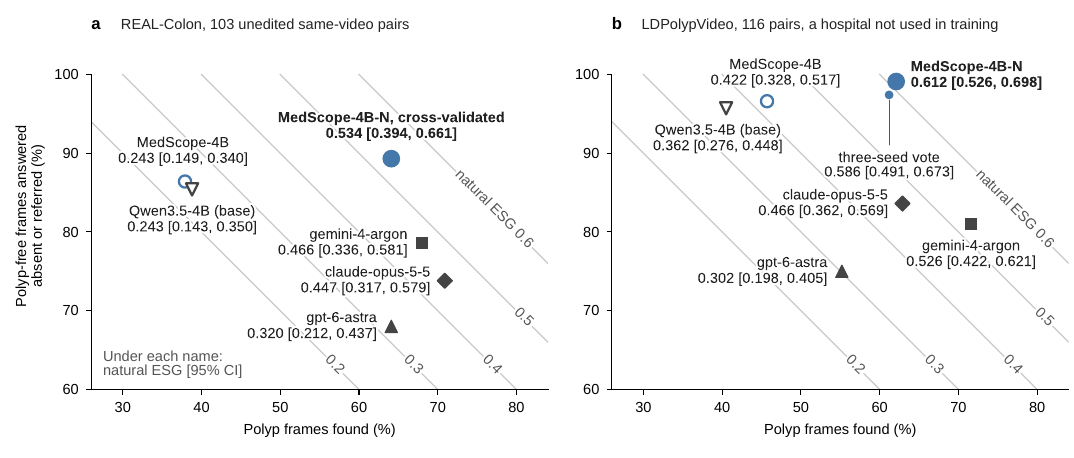}
\caption{Unedited colonoscopy pairs, by model. (a) REAL-Colon, 103 pairs from 46 videos; the answers of \msn{} are cross-validated. (b) LDPolypVideo, 116 pairs from 160 clips of a hospital not used in training or model selection. Each point places a model by the share of polyp frames it found (not answered ``absent'' and not referred, so a request for a better image counts as found) and the share of polyp-free frames it answered ``absent'' or referred. Natural ESG is the sum of the two shares minus one; grey diagonals join points of equal natural ESG, and the value under each name is the natural ESG with its 95\% confidence interval ($2{,}000$ bootstrap resamples of whole videos or clips). The small filled circle in (b) is the three-seed majority vote of \msn{}. Values are in Table~\ref{tab:natural}.}
\label{fig:natural}
\end{figure*}

In a two-by-two comparison of first-stage models, trained with or without counterfactual views and with or without unedited pairs (one seed; Supplementary Table~\ref{tab:s_twobytwo}), natural ESG was $0.563$ with unedited pairs only, $0.505$ with both kinds, $0.223$ with counterfactual views only and $0.214$ with neither. Unedited pairs raised it by $0.350$ $\ci{0.227}{0.474}$ without counterfactual views and by $0.282$ $\ci{0.156}{0.405}$ with them. In the pre-specified primary comparison, counterfactual views were not shown to add to it once unedited pairs were in training ($-0.058$ $\ci{-0.168}{0.042}$, $p = 0.311$). Without unedited pairs, the model with counterfactual views answered more polyp-free frames ``absent'' or referred (86.4\% against 68.9\%; a polyp reported on 3 against 27 of 103) and found fewer polyp frames (35.9\% against 52.4\%). On the main test, ESG was $0.744$ with both kinds and $0.260$ with unedited pairs only (difference $0.484$ $\ci{0.408}{0.558}$), and the model with both kinds made more lesion-free claims than the model with counterfactual views only (16.3\% against 9.4\%; difference 6.9 percentage points $\ci{3.0}{11.1}$).

\msn{} is a post hoc model: \msp{} with unedited colonoscopy frames and chest films added to its last training stage (Section~\ref{subsec:models}). On 116 pairs from LDPolypVideo, a hospital not used in training, it answered ``absent'' or referred on 99.1\% of the polyp-free frames, against 75.0\% to 83.6\% for the reference models, while finding 62.1\% of the polyp frames (reference models 55.2\% to 71.6\%; Fig.~\ref{fig:natural}b). Its natural ESG there was $0.612$ $\ci{0.526}{0.698}$, higher than that of \ms{} ($0.422$; difference $0.190$ $\ci{0.112}{0.267}$, $p < 0.001$), \opus{} ($0.466$; $0.147$ $\ci{0.043}{0.250}$, $p = 0.010$) and \gpt{} ($0.302$; $0.310$ $\ci{0.181}{0.431}$, $p < 0.001$). On the REAL-Colon pairs, already used to test \ms{}, its cross-validated natural ESG was $0.534$ $\ci{0.394}{0.661}$, higher than that of \ms{} ($0.291$ $\ci{0.176}{0.402}$, $p < 0.001$) and \gpt{} ($0.214$ $\ci{0.071}{0.343}$, $p = 0.003$) and not shown to differ from that of \opus{} ($0.087$ $\ci{-0.049}{0.230}$, $p = 0.259$). A difference from \gem{} was not shown on either set ($0.086$ $\ci{-0.017}{0.190}$, $p = 0.117$, and $0.068$ $\ci{-0.069}{0.202}$, $p = 0.347$), so by the pre-specified rule both results are mixed. The other two training seeds scored $0.578$ and $0.578$ on LDPolypVideo (Supplementary Table~\ref{tab:s_ldpolyp}). With only the pairs whose two frames lie at least 25 or at least 50 frames apart (115 and 98 pairs), its natural ESG was $0.609$ and $0.622$, and no difference from another model changed sign (Supplementary Table~\ref{tab:s_framegap}). With every video of the tested centre withheld from training, a model with only the colonoscopy frames added scored $0.495$ $\ci{0.370}{0.612}$ on the REAL-Colon pairs; frames closer to the polyp and chest X-ray follow-up pairs (a secondary analysis, rebuilt post hoc) are in Supplementary Tables~\ref{tab:s_loco}, \ref{tab:s_context} and~\ref{tab:s_cxrpairs}.

\begin{table*}[!t]
\caption{\ms{}, the post hoc model \msn{} and \gem{} on each test set: point estimates. Intervals, paired differences, \msp{} and the three-seed vote are in Supplementary Tables~\ref{tab:s_nmodel}, \ref{tab:s_nseeds} and~\ref{tab:s_chest}. Test sets: $n$ is the number of cases, or of films for the two chest-film sets. Correct: lesion images correct, the share of original and control-edited lesion images answered correctly. Claims: erased-lesion claims, the share of erased images on which the finding was reported. ESG: evidence sensitivity gap. Youden: Youden index on the unedited films of the two chest-film sets, the share of pneumothorax films called present plus the share of films without pneumothorax answered ``absent'' or referred, minus one. On the new SIIM-ACR patients, Correct includes control-edited films; \msn{} called 54 of 100 original pneumothorax films present. The CheXthought films have no edited version, and the reference models were not run on the SIIM-ACR films. --: not applicable, or model not run on the set.}
\label{tab:nmodel}
\centering
\footnotesize
\setlength{\tabcolsep}{3.5pt}
\begin{adjustbox}{max width=\textwidth}
\begin{tabular}{l cccc cccc cccc}
\toprule
 & \multicolumn{4}{c}{\ms{}} & \multicolumn{4}{c}{\msn{}} & \multicolumn{4}{c}{\gem{}} \\
\cmidrule(lr){2-5}\cmidrule(lr){6-9}\cmidrule(lr){10-13}
Test set ($n$) & Correct & Claims & ESG & Youden & Correct & Claims & ESG & Youden & Correct & Claims & ESG & Youden \\
\midrule
Main test ($507$) & $84.9\%$ & $9.6\%$ & $0.758$ & -- & $84.7\%$ & $9.1\%$ & $0.744$ & -- & $87.4\%$ & $41.1\%$ & $0.470$ & -- \\
External datasets ($169$) & $72.7\%$ & $11.8\%$ & $0.597$ & -- & $77.7\%$ & $10.9\%$ & $0.672$ & -- & $95.0\%$ & $58.8\%$ & $0.370$ & -- \\
Fresh internal set ($76$) & $68.1\%$ & $19.4\%$ & $0.472$ & -- & $68.1\%$ & $25.0\%$ & $0.417$ & -- & $63.9\%$ & $61.1\%$ & $0.056$ & -- \\
Fresh external set ($316$) & $47.6\%$ & $8.3\%$ & $0.393$ & -- & $60.7\%$ & $22.0\%$ & $0.381$ & -- & $74.1\%$ & $31.5\%$ & $0.435$ & -- \\
New SIIM-ACR patients ($200$) & $0.0\%$ & $0.0\%$ & $0.000$ & $0.000$ & $53.0\%$ & $25.0\%$ & $0.270$ & $0.500$ & -- & -- & -- & -- \\
CheXthought test patients ($600$) & -- & -- & -- & $0.000$ & -- & -- & -- & $0.660$ & -- & -- & -- & $0.460$ \\
\bottomrule
\end{tabular}
\end{adjustbox}
\end{table*}

\subsection{\texorpdfstring{\msn}{MedScope-4B-N} on the earlier tests and on chest films}
\label{subsec:res_nmodel}

On the main test \msn{} was not shown to differ from \ms{}: its ESG was $0.744$ against $0.758$ (difference $-0.014$ $\ci{-0.068}{0.040}$, $p = 0.680$), so the pre-specified criterion for a loss, an ESG lower by more than $0.05$, was not met, although the interval does not exclude a fall of that size (Table~\ref{tab:nmodel}). Each of its three training seeds ($0.744$, $0.740$ and $0.726$) had a higher ESG than each reference model (smallest difference $0.256$ $\ci{0.171}{0.338}$; Supplementary Tables~\ref{tab:s_nseeds} and~\ref{tab:s_nseeds_ref}). A majority vote of the three seeds, which needs three model runs per image, had an ESG of $0.753$ and lesion-free claims of 9.0\%. On the external datasets the ESG of \msn{} was higher than that of \ms{} ($0.672$ against $0.597$; difference $0.076$ $\ci{0.009}{0.143}$, $p = 0.033$) and lower than that of \msp{} ($0.739$; $-0.067$ $\ci{-0.126}{-0.009}$, $p = 0.025$).

On the fresh external set \msn{} was correct on more lesion images than \ms{} (60.7\% against 47.6\%; difference 13.1 percentage points $\ci{6.9}{19.7}$). It reported 22.0\% of erased lesions, against 11.3\% for \msp{} (difference 10.7 percentage points $\ci{5.4}{16.2}$) and 8.3\% for \ms{} (13.7 percentage points $\ci{7.7}{20.2}$); the increase was largest on the SIIM-ACR chest films (33\% against 10\% for \msp{}; Supplementary Table~\ref{tab:s_source}), and the training mix that validation did not select reported 17.3\% (Supplementary Table~\ref{tab:s_mix}). This narrowed its margin over \gem{} on that set (31.5\%), and its ESG ($0.381$ against $0.435$), lesion images correct (60.7\% against 74.1\%) and net utility ($0.114$ against $0.310$) were lower than those of \gem{} in point estimate; these contrasts were not tested.

On unedited chest films \msn{} had a higher Youden index than each reference model. On 300 pneumothorax films and 300 films without pneumothorax of CheXthought test patients, whose source also supplied part of its training films, its Youden index was $0.660$ $\ci{0.591}{0.726}$, against $0.460$ for \gem, $0.207$ for \opus{} and $0.377$ for \gpt{} (differences $0.200$ $\ci{0.122}{0.284}$, $0.453$ $\ci{0.344}{0.560}$ and $0.283$ $\ci{0.196}{0.375}$; each $p < 0.001$). Pneumothorax detection on new SIIM-ACR patients, used before to test the earlier models, was not raised over \msp{} by the first seed (54 against 53 of 100 films; difference 1.0 percentage point $\ci{-8.0}{10.0}$, $p = 0.920$); the other two seeds found 75 and 70 and reported the finding on more erased films (59.0\% and 43.0\%, against 25.0\%). Accuracy on standard question answering is in Supplementary Table~\ref{tab:s_vqa}.

\subsection{Alternative training designs and decision rules}
\label{subsec:res_designs}

None of seventeen alternative designs met its criterion fixed in advance (Supplementary Tables~\ref{tab:s_designs}, \ref{tab:s_designs_later} and~\ref{tab:s_inspect}; Supplementary Fig.~\ref{fig:s_designs}). They changed the training objective, the adapter, the agent design, the mix of editing tools in training, the reward for inspecting the image or the answer score at decision time. Matching the control edit to the erasure in training was inconclusive under its rule, and first-stage models trained with counterfactual views were not shown to exceed models trained without them and then by preference optimisation (ESG on the fresh external set $0.381$ against $0.355$, means of three seeds). Among designs with a recorded difference from their comparator, ESG changed by $-0.054$ to $0.041$; one adapter design lost the effect (ESG $0.000$). Several designs had one seed, so small gains are not excluded. None of 270 decision rules passed on both halves of the validation set.

\section{Discussion}\label{sec:discussion}

We erased lesions from medical images and made a matched control edit elsewhere in each image; nine comparison models, correct on $82.9\%$ to $88.8\%$ of lesion images, still reported the finding on $41.1\%$ to $61.6\%$ of 219 erased images. \ms, trained on such images, reported it on $9.6\%$, but on the fresh external set it was correct on fewer lesion images than \gem{} ($47.6\%$ against $74.1\%$), and its advantage did not carry over to unedited colonoscopy pairs (natural ESG $0.243$, as for its base model). \msn, a post hoc model trained also on unedited images, had a natural ESG of $0.612$ $\ci{0.526}{0.698}$ at a hospital not used in training, above two of the three reference models; a difference from \gem{} ($0.086$ $\ci{-0.017}{0.190}$) was not shown, and on the fresh external set it reported more erased lesions than \ms{} ($22.0\%$ against $8.3\%$). On the 20 of 36 erasures that a radiologist or a gastroenterologist accepted as free of the finding, the comparison models still reported it on $43.3\%$ $\ci{31.7}{56.1}$ and \ms{} on 1 of 20.

These results do not show that the comparison models would report a lesion on half of the lesion-free images met in practice; they show that removing the lesion region often left the answer unchanged. On unedited lesion-free images the same models reported a finding on $9.0\%$ to $30.9\%$, and on a blank image with the same question on none, so the question alone did not produce the reports. What in the edited image led to them (the organ, the instrument, the surrounding tissue or residual lesion) was not identified, and some may be correct readings of residual evidence: the specialists called the finding present on 16 of 36 erasures. The persistence, however, did not depend on such images; it held on the erasures that the specialists accepted. ESG therefore measures how far an answer depends on the lesion region, not error against a clinical reference standard. The persistence was seen in each of the nine comparison models and, for the three reference models, in every imaging type and under every prompting condition tested. Accuracy on unaltered images would not have revealed it: models that differed by less than 6 percentage points in lesion images correct differed by about 20 percentage points in erased-lesion claims.

Detection of the edit is the main alternative explanation for the result of \ms, which was trained on images built like those of the test. A model that answers ``absent'' whenever an image shows traces of editing does so on the control edit too, and its ESG is zero. The answers also followed the amount of lesion removed, from $4.2\%$ denied after a distant control edit to $86.7\%$ when the whole mask was erased (Section~\ref{subsec:res_alternatives}). The advantage of \ms{} persisted, at a lower ESG, with editing tools not used in training and on erasures that an edit detector scored as unedited, with some sensitivity to unfamiliar edit traces (Section~\ref{subsec:res_alternatives}). The unedited pairs do not settle the question for \ms, whose fewer reports of a polyp on polyp-free frames than its base model were offset by more requests for a better image (Section~\ref{subsec:res_pairs}). The data fit best answers that follow mainly whether the lesion region had been erased, with a contribution from edit traces that these tests cannot exclude.

Training on counterfactual views changed the answers on edited images and did not improve detection on unedited ones. It raised the main-test ESG of first-stage models from $0.279$ to $0.741$ (means over three seeds), and on a second base model from $0.374$ to $0.721$. On unedited colonoscopy pairs it raised the share of polyp-free frames answered ``absent'' or referred and lowered the share of polyp frames found, and once unedited pairs were in training it was not shown to add anything there (Section~\ref{subsec:res_pairs}). Detection on unedited images changed with unedited training images: \msn{} answered ``absent'' or referred on $99.1\%$ of polyp-free frames at the hospital not used in training (reference models $75.0\%$ to $83.6\%$), and its Youden index for pneumothorax on unedited films exceeded that of each reference model. Changes of the training objective, the adapter or the agent design were not shown to do better: none of seventeen alternative designs and none of 270 decision rules met its criterion (Section~\ref{subsec:res_designs}). One explanation, not tested here, is that an erased view shows a familiar context without its lesion, which may discourage answers made from context while adding no example of a lesion.

These findings agree with earlier evidence that accuracy does not establish image use. \citet{jin2024hidden} found by physician reading that $35.5\%$ of the correct answers of one model came with a flawed rationale, most often in image comprehension ($27.2\%$). Accuracy fell from $84\%$ to $28\%$ for one model when a vignette misled \citep{buckley2026text}. Four of five models answered $33.7\%$ to $37.7\%$ of 175 image-dependent questions correctly without the image, against $20\%$ by chance \citep{gu2025illusion}, and four systems kept about half of their correct answers when the radiograph was exchanged for one with the opposite label \citep{lotfinia2026image}. Our results differ in one respect: given a blank image, every comparison model asked for a better image, so the reports did not survive removal of the image. The difference may lie in the design, as our questions carry no clinical vignette and our protocol allows such a request. Removing or exchanging the whole image cannot detect reports that arise only when an image of the right kind is shown.

Erasure with a matched control can establish whether an answer depends on what the lesion region shows, across models, images and editing tools. It cannot establish that the finding is absent for a specialist, because secondary signs outside the mask, such as the mass effect of a brain tumour or the posterior acoustic shadowing of a breast mass, may remain. It also suits only findings that a mask can enclose. A diffuse finding may survive erasure (on QaTa-COV19 the ESG of \ms{} was $0.05$), and a pneumothorax, recognised by a pleural line and absent lung markings beyond it, may not be removed by filling its mask. Unedited same-patient pairs avoid these problems but exist only where a dataset marks the finding frame by frame.

This study has five main limitations. (i) Not every erasure was accepted by readers: the specialists called the finding present on 16 of 36 LaMa erasures and both trainees on 13 of 32, so the main-test rates include images with residual evidence; on the 16 erasures that a specialist called present, \ms{} reported the finding on 3, so part of its low rate may be denial of residual evidence. The specialist analysis rests on one reader per imaging type and 20 accepted erasures; further radiologists are reading. Two of the four readers who have finished are co-authors; all read without labels. The lesion-free labels are imperfect too: the trainees called a finding present on 23 of 64 reads of lesion-free images, and on 26 of the 288 lesion-free images at least seven of the nine comparison models reported a finding. In colonoscopy, REAL-Colon marks only resected polyps, and an LDPolypVideo polyp-free frame may lie as few as 5 frames from a polyp frame, so some polyp-free frames may show a polyp; the comparisons there were unchanged with pairs at least 50 frames apart. ESG compares two edits of one image and does not use these labels. (ii) Confirmation on unedited images covers colonoscopy frames and pneumothorax films, with three of the nine comparison models; \texttt{claude-sonnet-5-5}, whose main-test ESG was higher than that of \opus, was not run on them. \msn{} was not shown to exceed \gem{} on colonoscopy pairs (lower limit of the difference at LDPolypVideo $-0.017$), its pneumothorax films came from a source used in its training, and its detection on new SIIM-ACR patients varied between seeds (54, 75 and 70 of 100). (iii) The main test had been used with development versions of the trained model. On the fresh external set, run with no model before \ms{} was fixed, it reported $8.3\%$ of erased lesions but was correct on $47.6\%$ of lesion images, against $74.1\%$ for \gem. (iv) The tests were yes/no questions on single images of focal lesions from public datasets, to which the comparison models may have been exposed. Lesion-free images were not always from the anatomical sites of the lesion images, which affects lesion-free claims but not ESG. Each comparison model was run once, no task-specific detector was included, and the open-ended test (36 cases) did not confirm the advantage; all nine comparison models showed the persistence, which bounds the effect of a single run. (v) \msn{} was designed after earlier results were known, although its tests were fixed before it was trained, and the analysis plan is our own dated record. Net utility rests on a cost matrix that we wrote and used as the training reward, so it favours the trained model; ESG and the claim rates do not depend on it. No model was evaluated prospectively.

Readings by further radiologists will show whether the result on accepted erasures holds with more than one reader per imaging type. Unedited same-patient pairs with a verified reference standard are needed beyond colonoscopy; the chest X-ray follow-up pairs used here were a post hoc secondary analysis. The consequence for patients of a reported finding that the image does not support can be measured only prospectively, in clinical use.

\section{Conclusion}\label{sec:conclusion}

On public images and yes/no questions, nine comparison models reported the finding on $41.1\%$ to $61.6\%$ of images from which the lesion region had been erased, and on no blank image. A small open-weight model trained on matched counterfactual views reported it on $9.6\%$ while correct on $84.9\%$ of lesion images, which shows that the dependence of an answer on the lesion region can be measured and changed by training. The persistence held on erasures that specialists accepted as free of the finding. The change did not improve detection on unedited images; a post hoc model trained also on unedited images did improve there, without being shown to exceed the best reference model. More readers per imaging type and unedited pairs beyond colonoscopy are needed before the measure can be read clinically.

\section*{Data and code availability}
{\raggedright
All images come from public datasets. The copies used for training and for the main test are Kvasir-SEG from \url{https://datasets.simula.no/kvasir-seg}, and HyperKvasir (\texttt{SimulaMet-HOST/HyperKvasir}), SLAKE (\texttt{BoKelvin/SLAKE}), BUSI (\texttt{MedOtter/BUSI}), ChestX-Det (\texttt{MedOtter/ChestX-Det}) and TCGA-LGG (\texttt{MedOtter/TCGA-LGG-Mask}) on Hugging Face. SIIM-ACR, used for the fresh external set and in the training of \msp{} and \msn, came from Kaggle (\path{jesperdramsch/siim-acr-pneumothorax-segmentation-data}). The other evaluation sets were obtained from the sources cited in Table~\ref{tab:datasets} and Supplementary Table~\ref{tab:s_datasets}. Because the licences of the sources differ, no image is redistributed. Every test set is rebuilt from its source with the scripts of the project repository (\url{https://github.com/bhaskargurram-ai/medscope}). The repository also holds the code, the dated analysis plan (Supplementary Table~\ref{tab:s_plan}), the seeds, and the hardware and software versions. For the unedited pairs, the repository gives the video, clip and frame numbers of REAL-Colon and LDPolypVideo, without images; the LDPolypVideo repository states no licence. CheXpert Plus images and CheXthought annotations were used under their research-use agreements; only image identifiers and model outputs are released for them. To obtain the answers of the comparison models, test images were sent to the application programming interfaces of OpenAI, Anthropic and Google; these included REAL-Colon and LDPolypVideo frames and, for the three reference models, CheXthought test films.
\par}

{\raggedright
The models and MedScope-Bench are released on Hugging Face. MedScope-Bench (\url{https://huggingface.co/datasets/bhaskar-ai/MedScope-Bench}) holds the lists of views, without images, and the answers and transcripts of the model runs, including the comparison-model runs (31 October to 26 November 2026). \ms{} is released under its own name (\url{https://huggingface.co/bhaskar-ai/MedScope-4B}). Run from the public release with the released evaluation code, it gave a main-test ESG of $0.763$, against $0.758$ in the run reported here, with 1,102 of 1,115 decisions identical. \msp{} is released as \texttt{MedScope-4B-v4.2} (\url{https://huggingface.co/bhaskar-ai/MedScope-4B-v4.2}). \msn{}, trained under the name \texttt{MedScope-4B-v6}, and the answers of the runs on unedited pairs and chest films will be added to the release. The answer keys of the reader study are withheld until the specialist round closes. Earlier versions of this study were posted as arXiv:2611.13006 \citep{gurram2026medscope}.
\par}

\section*{CRediT authorship contribution statement}
\textbf{Bhaskar Gurram:} Conceptualization, Methodology, Software, Formal analysis, Investigation, Data curation, Visualization, Writing -- original draft, Project administration.
\textbf{Yoshitha Challagulla:} Software, Validation, Writing -- review \& editing.
\textbf{Jyothirmaye Challagulla:} Investigation (reader in the trainee round, internal medicine), Writing -- review \& editing.
\textbf{Samir C. Grover:} Investigation (reader in the specialist round), Writing -- review \& editing.

\section*{Declaration of competing interest}
B.G.\ and Y.C.\ are employed as AI engineers in industry; this work was done independently, outside their employment, and received no funding, data, computing or other resources from their employer.

\section*{Funding}
This research did not receive any specific grant from funding agencies in the public, commercial, or not-for-profit sectors.

\section*{Ethics statement}
All images come from public, de-identified datasets; no patient was contacted and no new patient data were collected. Physician readers took part voluntarily and gave consent on the reading page before starting, and their answers are stored under reader codes. Two readers are also authors (J.C., trainee round; S.C.G., specialist round). Like all readers, they read without access to image labels, kinds of edit, model outputs or other readers' answers.

\section*{Acknowledgements}
We thank the second reader of the trainee round (radiology), who prefers to remain anonymous.

\bibliography{medscope}
\bibliographystyle{plainnat}

\newpage
\onecolumn
\appendix
\setcounter{table}{0}\renewcommand{\thetable}{S\arabic{table}}
\setcounter{figure}{0}\renewcommand{\thefigure}{S\arabic{figure}}
\section{Analysis plan, models and methods in detail}
\label{sec:s_plan}

This section gives every analysis of the study with its status (Table~\ref{tab:s_overview}), the dated analysis plan (Table~\ref{tab:s_plan}), the evaluation sets in full (Table~\ref{tab:s_datasets}), the trained models (Table~\ref{tab:s_models}), the cost matrix (Table~\ref{tab:s_cost}), a bound on what a decision rule can add, the construction of the test views and pairs, the training details, further statistical details, the reader study protocol, and the prompts and model settings. Shares are given as percentages and differences between shares in percentage points; ESG, natural ESG, the Youden index, net utility and AUC (area under the receiver operating characteristic curve) are on their own scales.

\subsection{Overview of all analyses}

Table~\ref{tab:s_overview} lists every analysis of the study, in the order of the Results, with the design element that answers it, its test data, its status, the part of Table~\ref{tab:s_plan} that gives its decision rule, and where it is reported. An analysis is \emph{pre-specified} when its test set, comparison and decision rule were fixed and dated before the data it applies to were seen; \emph{descriptive} when its test set and comparison were fixed in advance without a decision rule; and \emph{post hoc} when it was added after earlier results were known. The plan was written during the study, from 30 October to 26 November 2026, each rule after the results that preceded it; each rule was written to a dated file in the public project repository (\url{https://github.com/bhaskargurram-ai/medscope}), whose file history is the record. It is our own record and not a third-party registry.

Confirmatory analyses are those in which no model of this study had been run on the test data when the rule was dated: \ms{} on the fresh internal and fresh external sets and on the REAL-Colon pairs, \msn{} on the LDPolypVideo pairs, the first test on the new SIIM-ACR patients (a variant of \ms{} with pneumothorax films in its last stage, 4 November 2026), and the two reader rounds. Four pre-specified analyses re-analyse main-test answers that existed, and whose pooled results were known, when the rule was fixed: the low-visibility erasures, the question types, the near-duplicate exclusion and the re-scoring on the trainees' answers; for these the rule was fixed before the analysis was run. Post hoc are: the models \msp{} and \msn, the last-stage variant and the two single-addition models (their selection rules and tests were fixed before they were trained or tested); the family of 27 primary comparisons; the interaction analysis of the edit-site control; the analysis of the pneumothorax failure of \ms{} (training coverage and image resolution); the plain prompt and the highest image-resolution setting; first-stage training with control edits only and with lesion insertions only; resampling of patients or image groups in place of cases; the comparison of image hashes across sets; the corrected version of the chest X-ray follow-up pairs; and a comparison of the comparator of the consistency loss with the first-stage model, reported with the alternative designs.

\begin{table}[!p]
\caption{Every analysis of the study, in the order of the Results: the question, the design element that answers it, the test data, the status, the part of Supplementary Table~\ref{tab:s_plan} that gives its decision rule (--: no decision rule), and where it is reported. Part 1 of 3. Status: pre-specified, descriptive or post hoc as defined in the text; ``existing answers'': a re-analysis of main-test answers that existed when the rule was fixed; ``confirmatory'': no model of this study had been run on the test data when the rule was dated. Sections are those of the main text. ESG: evidence sensitivity gap.}
\label{tab:s_overview}
\centering
\footnotesize
\setlength{\tabcolsep}{3pt}
\renewcommand{\arraystretch}{1.1}
\begin{tabular}{@{}L{0.22\textwidth}L{0.21\textwidth}L{0.16\textwidth}L{0.14\textwidth}L{0.05\textwidth}L{0.15\textwidth}@{}}
\toprule
\textbf{Question} & \textbf{Design element} & \textbf{Test data} & \textbf{Status} & \textbf{Rule} & \textbf{Reported in} \\
\midrule
\multicolumn{6}{@{}l}{\textit{Answers of the comparison models (Section~\ref{subsec:res_main})}} \\
Do models report the finding after erasure, and still after a control edit? & Erased image against matched control edit; nine comparison models, six open-weight models and \ms & Main test: 507 cases (219 with a lesion, 288 lesion-free), 1,115 views & Pre-specified$^{a}$; family of 27 comparisons post hoc & 1 & Table~\ref{tab:main}; Fig.~\ref{fig:main}; Tables~\ref{tab:s_modality} and~\ref{tab:s_primary} \\
Is the reported finding produced by the question alone? & Blank grey image, and an image of another imaging type, with the same question & 183 main-test lesion cases & Pre-specified & 2 & Table~\ref{tab:s_blank} \\
Does the result hold when the question names the finding? & Cases split by question type & 114 finding-specific and 105 generic lesion cases & Pre-specified; existing answers & 2 & Table~\ref{tab:s_qtype} \\
Does it depend on the prompt, the model settings or the wording? & Five further prompting and inference conditions for the three reference models; question wordings not used in training & 150 and 373 main-test cases & Descriptive; plain prompt and highest image-resolution setting post hoc & -- & Table~\ref{tab:s_prompts} \\
\midrule
\multicolumn{6}{@{}l}{\textit{Reader assessment of the erased images (Section~\ref{subsec:res_readers})}} \\
Do specialists still report the finding on erased images, and can they tell edited from unedited images? & Specialist reader round & 212 main-test images & Pre-specified; confirmatory & 3 & Table~\ref{tab:readers}; Table~\ref{tab:s_expert} \\
The same for trainee physicians & Earlier trainee round & 128 main-test images & Pre-specified; confirmatory & 3 & Table~\ref{tab:readers}; Table~\ref{tab:s_trainee} \\
Do the comparison models still report erasures that neither trainee called present? & Re-scoring of the model answers on these erasures & 10 erasures & Pre-specified; existing answers & 3 & Section~\ref{subsec:res_readers} \\
Do the comparison models still report erasures that the specialists accepted? & Re-scoring of the model answers on these erasures & 20 erasures & Post hoc rule; existing answers & 3 & Section~\ref{subsec:res_readers}; Table~\ref{tab:s_specialist} \\
\midrule
\multicolumn{6}{@{}l}{\textit{A model trained on counterfactual views (Sections~\ref{subsec:res_trained} and~\ref{subsec:res_designs})}} \\
Which training stage becomes \ms? & Selection rule on validation data & Validation set: 540 cases, 1,473 views & Pre-specified & 1 & Section~\ref{subsec:models}; Table~\ref{tab:s_models} \\
Do the counterfactual views cause the result? & First-stage models with and without counterfactual views (three seeds), with the share of ``absent'' answers matched, and on MedGemma 1.5 4B & Main test & Descriptive & -- & Fig.~\ref{fig:tools}; Table~\ref{tab:s_ablations} \\
Does it hold for other training seeds, and from the public release? & Training of \ms{} repeated with two further seeds; released model run again & Main test & Pre-specified & 2 & Table~\ref{tab:s_seeds} \\
Could test images that resemble training images explain it? & The quarter of test cases most similar to a training image removed in each source & 378 main-test cases kept & Pre-specified; existing answers & 2 & Section~\ref{subsec:res_trained} \\
Does it hold for an open-ended request? & Request to describe any abnormal findings, scored by a text-only judge & 36 main-test cases & Pre-specified & 2 & Section~\ref{subsec:res_trained} \\
Does it depend on lesion size, and does stated confidence rank the errors? & Quartiles of lesion area; risk-coverage curve & Main test & Descriptive & -- & Section~\ref{subsec:res_trained} \\
Do other objectives, adapters, agent designs, tool mixes or rewards do better? & Seventeen alternative designs, each against a matched comparator & Tools not used in training; fresh external set; validation set; REAL-Colon pairs & Pre-specified & 5, 6 & Section~\ref{subsec:res_designs}; Table~\ref{tab:s_designs}; Fig.~\ref{fig:s_designs} \\
Can a decision rule applied to model outputs do better? & 270 decision rules, chosen on one half of the validation cases and confirmed on the other & Validation set & Pre-specified & 6 & Section~\ref{subsec:res_designs}; Table~\ref{tab:s_designs} \\
\bottomrule
\multicolumn{6}{@{}p{0.97\textwidth}@{}}{$^{a}$\,The training procedure and selection rule of \ms, the test set and the measures were fixed before the model's training data were built. Development versions of the trained model had been run on this test set and on the external datasets (Section~\ref{sec:s_training}). A planned analysis of whether ESG with its matched control selects better models needs a new reader-assessed cohort and was not run (Table~\ref{tab:s_plan}).} \\
\end{tabular}
\end{table}

\begin{table}[!p]
\ContinuedFloat
\caption{(continued) Part 2 of 3.}
\centering
\footnotesize
\setlength{\tabcolsep}{3pt}
\renewcommand{\arraystretch}{1.1}
\begin{tabular}{@{}L{0.22\textwidth}L{0.21\textwidth}L{0.16\textwidth}L{0.14\textwidth}L{0.05\textwidth}L{0.15\textwidth}@{}}
\toprule
\textbf{Question} & \textbf{Design element} & \textbf{Test data} & \textbf{Status} & \textbf{Rule} & \textbf{Reported in} \\
\midrule
\multicolumn{6}{@{}l}{\textit{Detection of the edit (Section~\ref{subsec:res_alternatives})}} \\
Would a model that answers ``absent'' to any edited image score well? & Matched control edit (ESG is zero for such a model); unedited lesion-free images & Main test & By design & -- & Table~\ref{tab:alternatives} \\
Does the result depend on the editing tool used in training? & Erased and control-edited images remade with two tools not used in training & 125 cases common to all tools; 219 lesion cases with TELEA, 191 with Stable Diffusion & Descriptive & -- & Fig.~\ref{fig:tools}; Table~\ref{tab:s_tools} \\
Which training views are needed? & First-stage models with erasures only, with control edits added, with lesion insertions added, with both & Main test; tools not used in training & Descriptive; control edits only and lesion insertions only post hoc & -- & Fig.~\ref{fig:tools}; Table~\ref{tab:s_ablations} \\
Is the answer driven by an edit at the lesion's site, or by how much of the lesion is removed? & Edit-site control (ring edit around the intact lesion), half erasure, lesion pixels pasted back; probes of hidden states; other frozen encoders & 179 validation lesion cases (143 whose lesion the first-stage model found) & Pre-specified; interaction analysis post hoc & 2 & Table~\ref{tab:alternatives}; Table~\ref{tab:s_site} \\
Does it hold on erasures that an edit detector scores as unedited? & Low-visibility erasures & 111 of 219 main-test erasures & Pre-specified; existing answers & 2 & Table~\ref{tab:alternatives}; Table~\ref{tab:s_visibility} \\
\midrule
\multicolumn{6}{@{}l}{\textit{External datasets and new test sets (Section~\ref{subsec:res_external})}} \\
Does the result hold on data not used in training? & \ms{} run once on each set & External datasets (169 cases); fresh internal (76); fresh external (316) & Pre-specified; fresh sets confirmatory$^{b}$ & 1 & Table~\ref{tab:s_external} \\
Why did \ms{} miss pneumothorax? & Counts of pneumothorax in training and validation; films at original resolution & 42 fresh external films; training and validation sets & Post hoc & -- & Section~\ref{subsec:res_external} \\
Do pneumothorax films in training restore pneumothorax detection? & Last-stage variant; \msp & 200 new SIIM-ACR patients & Models post hoc; rules pre-specified; first test confirmatory & 1 & Section~\ref{subsec:res_external}; Table~\ref{tab:s_models} \\
Does \ms{} keep accuracy on standard question answering? & Test splits of three question-answering datasets & 1,051 items & Descriptive & -- & Table~\ref{tab:s_vqa} \\
Requests for a better image by kind of degradation & Degraded copies on all four edited test sets & Degraded copies & Descriptive; planned and not reported & -- & This table \\
\midrule
\multicolumn{6}{@{}l}{\textit{Unedited image pairs (Section~\ref{subsec:res_pairs})}} \\
Does the advantage of \ms{} carry over to unedited same-patient pairs? & Polyp frame against polyp-free frame of the same video & REAL-Colon: 103 pairs from 46 videos & Pre-specified; confirmatory & 3 & Table~\ref{tab:natural}; Fig.~\ref{fig:natural} \\
With unedited pairs in training, does it hold on patients not used in training, and with no centre shared? & \msp{} with colonoscopy pairs; cross-validated answers; leave-one-centre-out training & REAL-Colon: 103 pairs & Model post hoc; guard and tests pre-specified & 1, 3 & Tables~\ref{tab:s_additions}, \ref{tab:s_loco} \\
Against polyp-free frames from the same stretch of the video? & Context-matched frames & REAL-Colon: 96 pairs & Descriptive & -- & Table~\ref{tab:s_context} \\
What does each kind of training data contribute? & Two-by-two: with or without counterfactual views, with or without unedited pairs & REAL-Colon: 103 pairs; main test & Pre-specified & 3 & Section~\ref{subsec:res_pairs}; Table~\ref{tab:s_twobytwo} \\
Does \msn{} hold at a hospital not used in training, and on REAL-Colon? & \msn; cross-validated answers on REAL-Colon & LDPolypVideo: 116 pairs; REAL-Colon: 103 pairs & Model post hoc; tests pre-specified; LDPolypVideo confirmatory & 3 & Table~\ref{tab:natural}; Tables~\ref{tab:s_ldpolyp}, \ref{tab:s_realcolon} \\
Do pairs with nearby frames drive the LDPolypVideo result? & Pairs restricted to frames at least 25 or 50 frames apart & 115 and 98 pairs & Descriptive & -- & Table~\ref{tab:s_framegap} \\
Do same-patient chest X-ray pairs show the same? & Follow-up pairs from NIH ChestX-ray14 & 44 pairs (planned version); 183 of 200 (corrected version) & Descriptive; corrected version post hoc & -- & Table~\ref{tab:s_cxrpairs} \\
\bottomrule
\multicolumn{6}{@{}p{0.97\textwidth}@{}}{$^{b}$\,No model of this study had been run on the two fresh sets before \ms{} was fixed; the external datasets had been used with development versions of the trained model.} \\
\end{tabular}
\end{table}

\begin{table}[!p]
\ContinuedFloat
\caption{(continued) Part 3 of 3.}
\centering
\footnotesize
\setlength{\tabcolsep}{3pt}
\renewcommand{\arraystretch}{1.1}
\begin{tabular}{@{}L{0.22\textwidth}L{0.21\textwidth}L{0.16\textwidth}L{0.14\textwidth}L{0.05\textwidth}L{0.15\textwidth}@{}}
\toprule
\textbf{Question} & \textbf{Design element} & \textbf{Test data} & \textbf{Status} & \textbf{Rule} & \textbf{Reported in} \\
\midrule
\multicolumn{6}{@{}l}{\textit{\msn{} on the earlier tests and on chest films (Section~\ref{subsec:res_nmodel})}} \\
Which training mix, and does it pass the validation guard? & Two candidate mixes chosen on validation data & Validation set; SIIM-ACR validation patients & Model post hoc; rule pre-specified & 1 & Table~\ref{tab:s_mix} \\
Do CheXthought films in training raise pneumothorax detection? & \msp{} with CheXthought films & 200 new SIIM-ACR patients & Model post hoc; guard and test pre-specified & 1 & Table~\ref{tab:s_additions} \\
Does \msn{} keep the main-test result, across seeds, and does the vote help? & \msn{} on the main test; three seeds and their vote & Main test & Model post hoc; tests pre-specified & 4 & Table~\ref{tab:nmodel}; Table~\ref{tab:s_nseeds} \\
Does it raise pneumothorax detection, and what does it cost on the fresh external set? & \msn{} against \msp & 200 new SIIM-ACR patients; fresh external set & Model post hoc; tests pre-specified & 4 & Table~\ref{tab:nmodel}; Table~\ref{tab:s_nmodel} \\
How does it do elsewhere? & \msn{} on CheXthought test films, external datasets, fresh internal set, standard question answering, low-visibility erasures and question types & 600 CheXthought films; the other sets as above & Model post hoc; descriptive & -- & Table~\ref{tab:s_nmodel}; Tables~\ref{tab:s_chest}, \ref{tab:s_vqa} \\
\bottomrule
\end{tabular}
\end{table}

\subsection{Dated analysis plan}

Table~\ref{tab:s_plan} lists every analysis that had a decision rule, with the date on which the rule was fixed, the rule and its outcome. Outcomes are given in the terms of the rule: met, not met, or mixed where the rule defines an intermediate result. ``Post hoc model'' means that the model was designed after earlier results were known; where its selection rule or its test was fixed before the model was trained or tested, the table says so.

\newcommand{\planhead}{\toprule
\textbf{Analysis: what was asked} & \textbf{Rule fixed} & \textbf{Status} & \textbf{Decision rule} & \textbf{Outcome} \\
\midrule}
\newenvironment{plantab}{\tabular{@{}L{0.205\textwidth}L{0.12\textwidth}L{0.105\textwidth}L{0.285\textwidth}L{0.215\textwidth}@{}}}{\endtabular}

\begin{table}[!p]
\caption{Analyses with a decision rule fixed in advance: what was asked, when the rule was fixed, the status of the analysis, the rule, and the outcome. Part 1 of 6: the trained models and their selection. All dates are in 2026. ESG: evidence sensitivity gap. Intervals are 95\% bootstrap confidence intervals. The three reference models are \gem, \opus{} and \gpt.}
\label{tab:s_plan}
\centering
\footnotesize
\setlength{\tabcolsep}{3pt}
\renewcommand{\arraystretch}{1.1}
\begin{plantab}
\planhead
Which training stage becomes \ms? & 1 Nov, before its training data were built. A clause on the development version of the earlier series was added on 2 Nov, before any validation result of stages 2 and 3. & Pre-specified & Take stage 3 if its validation net utility is at most 0.02 below that of stage 2 and its lesion-free claims are at most 2 percentage points above; otherwise apply the same test to stage 2 against stage 1. Keep the earlier development version if the chosen stage is more than 0.02 below it in validation net utility. & Met: stage 3 chosen (net utility 0.613 against 0.610 for stage 2; lesion-free claims 7.7\% against 12.7\%; earlier development version 0.593). \\
Main test: is \ms{} ahead of each of the nine comparison models on the three primary outcomes? & Model, test set and measures: 1 Nov. The family of 27 comparisons was defined after the results were known. & Pre-specified; multiplicity family post hoc & Bonferroni correction over ESG, erased-lesion claims and net utility against each of the nine comparison models. & All 27 comparisons at $p \le 0.001$. \\
Fresh internal and fresh external sets: does the result hold on data on which no model had been run? & 1 Nov & Pre-specified; confirmatory & Each set is run once after the model is fixed, with the measures of the main test. & Fresh internal: ESG 0.472 against 0.000 to 0.083 for the three reference models. Fresh external: erased-lesion claims 8.3\% against 31.5\% to 42.3\%; net utility below \gem{} ($-0.363$ $\ci{-0.496}{-0.233}$); pneumothorax found on 0 of 48 films. \\
Does repeating only the last stage with pneumothorax films pass the selection rule? (variant not used further) & 4 Nov, before any of its results & Post hoc model; rule pre-specified & Validation net utility at most 0.02 below that of \ms, lesion-free claims at most 2 percentage points above, and more pneumothoraces found than \ms{} on SIIM-ACR validation patients. & Not met (lesion-free claims 10.0\% against 7.7\%). \\
Does \msp{} pass its validation rule? & 15 Nov, before any of its test results & Post hoc model; rule pre-specified & Validation net utility at most 0.02 below that of \ms, lesion-free claims at most 2 percentage points above, and at least as many SIIM-ACR validation pneumothoraces found as the last-stage variant. & Met (net utility 0.632 against 0.613; lesion-free claims 9.1\% against 7.7\%; pneumothoraces found 65\% against 48\%). \\
\msp{} with colonoscopy pairs: with unedited pairs in training, does the advantage hold on patients not used in training? & 21 Nov, before any training frame was drawn & Post hoc model; guard and test pre-specified & Guard: validation net utility at most 0.02 below that of \ms{} and lesion-free claims at most 2 percentage points above. Test (cross-validated answers on the 103 REAL-Colon pairs): natural ESG above each of the three reference models with the interval above zero, holds; above none, does not hold; otherwise mixed. & Guard met (0.654; 9.1\%). Mixed: natural ESG 0.553 $\ci{0.430}{0.670}$, above \gpt{} only. \\
\msp{} with CheXthought films: do these films raise pneumothorax detection without more claims on films without pneumothorax? & 22 Nov, before any CheXthought image was downloaded & Post hoc model; guard and test pre-specified & Guard: validation net utility at most 0.02 below that of \msp{} and lesion-free claims at most 2 percentage points above. Raised if, on the 200 new SIIM-ACR patients, the share of pneumothorax films answered present exceeds that of \msp{} with the interval above zero, and films without pneumothorax answered present are not more frequent by more than 5 percentage points. & Guard met (0.628; 9.5\%). Raised: 69 against 53 of 100 ($+16.0$ percentage points $\ci{+7.0}{+26.0}$); 7 against 5 of 100 films without pneumothorax. \\
\msn: which training mix, and does it pass the validation guard? & 24 Nov, before the model was trained and before any LDPolypVideo file was downloaded & Post hoc model; rule pre-specified & Take the mix that repeats erased images, unless the plain mix is higher by more than 0.005 in net utility over the main validation views and the SIIM-ACR validation patients together. Guard: on the main validation views, net utility at most 0.02 below that of \ms{} and lesion-free claims at most 2 percentage points above. & Plain mix chosen (0.547 against 0.452). Guard met (net utility 0.642, limit 0.593; lesion-free claims 9.1\%, limit 9.7\%). \\
\bottomrule
\end{plantab}
\end{table}

\begin{table}[!p]
\ContinuedFloat
\caption{(continued) Part 2 of 6: analyses of the main result and of the alternative explanations. AUC: area under the receiver operating characteristic curve.}
\centering
\footnotesize
\setlength{\tabcolsep}{3pt}
\renewcommand{\arraystretch}{1.1}
\begin{plantab}
\planhead
Does the main result hold for two further training seeds of \ms? & 20 Nov & Pre-specified & Holds if the main-test ESG of every seed is above that of each of the three reference models with the paired interval above zero. & Met (ESG 0.758, 0.744 and 0.731; every comparison $p < 0.001$). \\
Blank-image control: is the reported finding produced by the question alone? & 22 Nov & Pre-specified & For each model: if erased-lesion claims exceed claims on a blank image with the interval above zero, its claims are not explained by the question alone. & Met for every model: claims on the blank image 0\%; erased minus blank 42.1 to 56.8 percentage points for the nine comparison models and 10.9 $\ci{6.6}{15.8}$ for \ms. \\
Question type: does the main result hold on questions that name the finding? & 22 Nov & Pre-specified; existing answers & Holds if, on finding-specific questions, the ESG of \ms{} is above that of each of the three reference models with the interval above zero. & Met (0.728 $\ci{0.649}{0.807}$ against 0.298 to 0.439 for the nine comparison models). \\
Low-visibility erasures: does the advantage hold on erasures that an edit detector scores as unedited? & 20 Nov & Pre-specified; existing answers & Edit visibility does not explain the advantage if, on these cases, the ESG of \ms{} is above that of each of the three reference models with the paired interval above zero. & Met (0.640 $\ci{0.550}{0.721}$; differences 0.270, 0.450 and 0.378, each $p < 0.001$). \\
Open-ended request: does the advantage hold without a yes/no question? & 20 Nov & Pre-specified & Holds if the free-text ESG of \ms{} is above that of each of the three reference models with the paired interval above zero. & Not met (0.472 against 0.306 to 0.361; no interval excluded zero; 36 cases). \\
Does the public release reproduce the main result? & 20 Nov & Pre-specified & Reproduced if main-test ESG is within 0.02 of 0.758 and at least 97\% of per-view decisions match the original run. & Met (ESG 0.763; 98.8\% of decisions identical). \\
Near-duplicate exclusion: could test images that resemble training images explain the result? & 15 Nov & Pre-specified; existing answers & Not explained if, after the 25\% of test cases most similar to a training image are removed in every source, the ESG advantage over each of the three reference models keeps its sign with the interval above zero. & Met (smallest advantage 0.306 $\ci{0.200}{0.405}$; 378 cases kept). \\
Edit-site control: does an edit at the lesion's site, with the lesion intact, produce the answer ``absent''? & 14 Nov, before any score was computed. The planned score failed its validity check (63\% agreement with the model's answers against the required 80\%) and was replaced by generated answers the same day, before the rule was applied. & Pre-specified; interaction analysis post hoc & A site effect is flagged if the share of intact lesions denied after a ring edit is at least 10\%, at least 5 percentage points above the share after a control edit elsewhere, and at least 5 percentage points above that of the base model. & Flag met for TELEA only (20.3\%; ring minus control edit elsewhere 7.7 percentage points $\ci{0.0}{15.4}$). Post hoc: the part specific to the site, relative to the base model, was 1.8 percentage points $\ci{-8.0}{11.6}$, too imprecise to exclude a site effect of the size the flag was set for. \\
Probes: do hidden states separate erased from intact lesions better than the answers do? & 14 Nov & Pre-specified & A read-out gap exists if the best linear probe exceeds the model's own answers by at least 0.05 in AUC (erased against ring-edited views, TELEA). & Not found (at most 0.028). \\
The bound of \ref{sec:s_bound}, measured: do the answers use the separation that the hidden states hold? & 15 Nov & Pre-specified & No read-out gap if a lower bound on the separation in the hidden states is within 0.05 of the model's own difference in ``absent or refer'' answers. & No gap (LaMa: 0.665 and 0.642 against 0.648; TELEA: 0.492 and 0.615 against 0.642). \\
Is the limit specific to the image encoder? & 15 Nov & Pre-specified & Another frozen encoder holds more if it separates erasures from control edits better by at least 0.05 in AUC with the paired interval above zero. & None did. \\
\bottomrule
\end{plantab}
\end{table}

\begin{table}[!p]
\ContinuedFloat
\caption{(continued) Part 3 of 6: reader assessment and unedited image pairs. CI: confidence interval.}
\centering
\footnotesize
\setlength{\tabcolsep}{3pt}
\renewcommand{\arraystretch}{1.1}
\begin{plantab}
\planhead
Trainee round: do readers still report the finding on erased images? & By 8 Nov, before any answer was collected & Pre-specified; confirmatory & The erasures are accepted if the upper limit of the 95\% interval of the share of erased images with the finding reported is at most 10\%. & Not met (35 of 64 reads; both trainees on 13 of 32 erased images, exact 95\% CI 23.7 to 59.4). \\
Do the comparison models still report the finding on erasures that neither trainee called present? & 14 Nov, before any model was scored in this way & Pre-specified; existing answers & The main result holds on these images if the pooled share for the nine comparison models has a lower 95\% limit above 20\%. & Not met (26\%, 95\% CI 10\% to 42\%; 10 erasures). \\
Specialist round: do specialists still report the finding on erased images? & 15 Nov; closing rule 18 Nov, before any answer was examined & Pre-specified; confirmatory & For each editing tool, the erasures are accepted if the upper limit of the exact 95\% CI of the share of erased images that the reader majority calls present is at most 10\%. The round closes when two radiologists have read all 212 images or on 10 January 2027. & Round open for further radiologists; a gastroenterologist and a radiologist have read all 212 images (Table~\ref{tab:readers}). Interim: with each imaging type read by its specialist, 16 of 36 LaMa erasures and 17 of 36 Stable Diffusion erasures were called present. Not met for either tool. \\
Do the comparison models still report the finding on erasures that the specialists accepted? & 26 Nov, after the radiologist had finished and before the re-scoring was run & Post hoc rule; existing answers & Erasures judged by the specialist for the imaging type; accepted if not called present. The main result holds if the pooled share for the nine comparison models has a lower 95\% limit above 20\%; fewer than 10 accepted erasures: descriptive only. & Met (43.3\%, 95\% CI 31.7\% to 56.1\%; 20 erasures; Table~\ref{tab:s_specialist}). \\
Does the advantage of \ms{} carry over to unedited same-patient pairs (REAL-Colon)? & 19 Nov, before any frame was downloaded & Pre-specified; confirmatory & Natural ESG above each of the three reference models with the interval above zero: carries over. Above none: does not. Otherwise: mixed. & Not met: does not carry over (0.243 $\ci{0.149}{0.340}$; above none of the three). \\
With no centre shared between training and test (\msp{} with colonoscopy pairs, leave-one-centre-out), does the advantage hold? & 22 Nov & Post hoc model; test pre-specified & The rule of the row above. & Mixed (0.495 $\ci{0.370}{0.612}$; above \gpt{} only). \\
Two-by-two: do counterfactual views add anything on unedited frames once unedited pairs are in training? & 22 Nov & Pre-specified & Natural ESG with both kinds of data minus natural ESG with unedited pairs only. Interval above zero: they add. Interval includes zero: not shown. Interval below zero: they cost. & Not shown ($-0.058$ $\ci{-0.168}{+0.042}$). \\
\msn{} at a hospital not used in training (LDPolypVideo): is its natural ESG above the three reference models? & 24 Nov, before any file of the dataset was downloaded & Post hoc model; test pre-specified; confirmatory & The three-way rule above, applied to the first seed. Also: \msn{} minus \ms{} with the interval above zero means that training on unedited pairs carries over to the new hospital. Fewer than 40 pairs: descriptive only. & Mixed (0.612 $\ci{0.526}{0.698}$; above \opus{} and \gpt; difference from \gem{} 0.086 $\ci{-0.017}{0.190}$, not shown). Above \ms{} (0.190 $\ci{0.112}{0.267}$). 116 pairs. The other two seeds: 0.578 and 0.578. \\
\msn{} on the REAL-Colon pairs, cross-validated & 24 Nov & Post hoc model; test pre-specified & The three-way rule above. & Mixed (0.534 $\ci{0.394}{0.661}$; above \gpt{} only). \\
\bottomrule
\end{plantab}
\end{table}

\begin{table}[!p]
\ContinuedFloat
\caption{(continued) Part 4 of 6: \msn{} on the earlier test sets.}
\centering
\footnotesize
\setlength{\tabcolsep}{3pt}
\renewcommand{\arraystretch}{1.1}
\begin{plantab}
\planhead
Does the advantage of \msn{} hold for two further training seeds? & 24 Nov & Post hoc model; test pre-specified & Holds if the main-test ESG of every seed is above that of each of the three reference models with the paired interval above zero. & Met (ESG 0.744, 0.740 and 0.726). \\
Is the three-seed vote the reported operating point? & 24 Nov & Post hoc model; rule pre-specified & The vote is the reported operating point if its main-test lesion-free claims are not above the mean of the seeds and its main-test ESG is at most 0.02 below their mean. & Met (lesion-free claims 9.0\% against a mean of 11.6\%; ESG 0.753 against a mean of 0.737). The single model is nevertheless reported as primary, because the confirmatory test on LDPolypVideo was specified for the first seed; the vote is given beside it. \\
Does \msn{} give up the main-test result? & 24 Nov & Post hoc model; test pre-specified & A main-test ESG more than 0.05 below that of \ms{} counts as a trade. & No trade (0.744 against 0.758; $-0.014$ $\ci{-0.068}{+0.040}$). \\
Does \msn{} raise pneumothorax detection on the 200 new SIIM-ACR patients over \msp? & 24 Nov & Post hoc model; test pre-specified (patients used before with the earlier models) & Raised if the difference in pneumothorax films answered present has an interval above zero, and films without pneumothorax answered present are not more frequent by more than 5 percentage points. & Not raised (54 against 53 of 100; $+1.0$ percentage point $\ci{-8.0}{+10.0}$). \\
Fresh external set: are the erased-lesion claims of \msn{} above those of \msp? & 24 Nov & Post hoc model; test pre-specified & \msn{} minus \msp, paired over cases. Interval above zero: the cost remains. Otherwise: no cost shown. & The cost remains (22.0\% against 11.3\%; $+10.7$ percentage points $\ci{+5.4}{+16.2}$, $p < 0.001$). \\
\bottomrule
\end{plantab}
\end{table}

\begin{table}[!p]
\ContinuedFloat
\caption{(continued) Part 5 of 6: alternative training objectives and adapter settings. Each design was compared with a matched comparator. Tools not used in training are TELEA and Stable Diffusion. Rows marked $\dagger$ in parts 5 and 6 make up the twelve alternative designs of the first series and rows marked $\ddagger$ in part 6 the five designs tested later, seventeen in all (Section~\ref{subsec:res_designs}); a row with two designs counts twice.}
\centering
\footnotesize
\setlength{\tabcolsep}{3pt}
\renewcommand{\arraystretch}{1.1}
\begin{plantab}
\planhead
Must the control edit match the erasure in shape and tool?$^{\dagger}$ & 6 Nov & Pre-specified & Matching matters if, with a matched control in training, intact lesions are denied after a control edit less often by at least 5 percentage points, with the interval excluding zero, on the fresh external set and on tools not used in training. & Inconclusive under the rule; matching was not shown to matter. \\
Is counterfactual training better than preference optimisation applied after training without counterfactual views?$^{\dagger}$ & 6 Nov (three seeds: 6 Nov) & Pre-specified & Better if ESG and net utility are both higher on the fresh external set with intervals excluding zero, in all three seed pairs. & No difference shown (ESG 0.381 (SD 0.018) against 0.355 (SD 0.015) over three seeds). \\
Does preference optimisation on top of counterfactual training help?$^{\dagger}$ & 6 Nov & Pre-specified & Passes if, on the fresh external set and on tools not used in training, lesion images correct rise by at least 3 percentage points with the interval excluding zero and erased-lesion claims do not rise by more than 3 percentage points. A further training stage is added only if it passes. & Not met (tools not used in training: erased-lesion claims 24.4\% against 16.3\%). The further stage was not trained. \\
Does a consistency loss on paired views lower erased-lesion claims on tools not used in training?$^{\dagger}$ & 7 Nov, after a favourable one-seed trial on the same test sets & Pre-specified & Lower in every one of three seed pairs with the pooled interval excluding zero; ESG not lower by more than 0.03; guards: main-test lesion-free claims not higher by more than 3 percentage points, fresh-external ESG and net utility not lower by more than 0.03. & Not met ($+1.1$ percentage points $\ci{-0.9}{+3.0}$; main-test guard not met, $+5.2$ percentage points). \\
Does ESG with its matched control select better models than measures without it? (new reader-assessed cohort) & 8 Nov & Pre-specified & The model selected with ESG has a net utility higher by at least 0.05, with the interval excluding zero, on a second source or editing tool as well. & Not run: it needs a new reader-assessed cohort, which was not built. \\
Do adapters of rank 64 help?$^{\dagger}$ & 11 Nov & Pre-specified, one seed & On the fresh external set, lesion images correct higher by at least 5 percentage points or more pneumothorax films found, without erased-lesion claims rising by more than 3 percentage points on the main test or on tools not used in training. & Not met (main-test ESG 0.731 against 0.735). \\
Do adapters on the vision encoder help?$^{\dagger}$ (two learning rates; the second was declared as the only repeat) & 11 Nov & Pre-specified, one seed each & The rule of the row above. & Not met (higher learning rate: ESG 0.000 on every set; lower learning rate: fresh-external lesion images correct 48.8\% against 56.0\%). \\
Does reinforcement learning on the cost-matrix reward help?$^{\dagger}$ & 11 Nov & Pre-specified, one seed & On tools not used in training: erased-lesion claims not higher than the comparator's, net utility higher by at least 0.05, lesion images correct not lower by more than 3 percentage points; main-test lesion-free claims not higher by more than 3 percentage points. & Not met (net utility about $+0.056$; erased-lesion claims about $+2.2$ percentage points). \\
\bottomrule
\end{plantab}
\end{table}

\begin{table}[!p]
\ContinuedFloat
\caption{(continued) Part 6 of 6: agent designs, decision rules, editing-tool mixes and the designs tested later. Rows marked $\dagger$ belong to the twelve alternative designs of the first series and rows marked $\ddagger$ to the five designs tested later (see part 5).}
\centering
\footnotesize
\setlength{\tabcolsep}{3pt}
\renewcommand{\arraystretch}{1.1}
\begin{plantab}
\planhead
Does a multi-agent system with an erase-and-ask-again verifier help?$^{\dagger}$ (two versions; the second was declared as the only repeat) & 13 Nov & Pre-specified, one seed each & On tools not used in training: net utility higher by at least 0.05, erased-lesion claims not higher, lesion images correct not lower by more than 3 percentage points; guards on main-test lesion-free claims and fresh-external lesion images correct. & Not met twice (first: fresh-external lesion images correct 37.8\% against 56.0\%; second: erased-lesion claims $+6.1$ percentage points $\ci{+2.7}{+9.6}$). \\
Can any of 270 decision rules applied to model outputs raise ESG? & 13 Nov, before each of the two sets of rules was run & Pre-specified & A rule passes on one half of the validation cases and again on the other: net utility $+0.03$ or more, ESG $+0.02$ or more, lesion images correct not lower by more than 3 percentage points, lesion-free claims not higher by more than 3 percentage points. & None met (19 of 128 rules on recorded outputs passed the first half, none the second; none of 142 rules with new test-time evidence passed the first half). \\
Does training with a family of four editing tools transfer to TELEA?$^{\dagger}$ (two designs) & 14 Nov & Pre-specified, one seed each & TELEA ESG $+0.03$ or more and denial of intact lesions after a TELEA control edit lower by 3 percentage points or more; LaMa ESG within 0.02, and lesion images correct and lesion-free claims within 3 percentage points. & Not met (TELEA ESG $+0.017$ $\ci{-0.034}{0.067}$ and $+0.011$ $\ci{-0.056}{0.084}$). \\
Does a term that swaps the whole image for a similar lesion-free image help?$^{\ddagger}$ & 15 Nov & Pre-specified, one seed & TELEA ESG $+0.03$ or more; LaMa ESG within 0.02, and lesion images correct and lesion-free claims within 3 percentage points. & Not met (TELEA ESG $-0.028$ $\ci{-0.073}{0.017}$). \\
Do vision-encoder adapters with layer-wise learning-rate decay find more lesions?$^{\ddagger}$ & 15 Nov & Pre-specified, one seed & Unedited lesion images correct higher by 3 percentage points or more; LaMa and TELEA ESG not lower by more than 0.02; lesion-free claims not higher by more than 3 percentage points. & Not met ($+1.7$ percentage points $\ci{-3.4}{6.7}$). \\
Does subtracting a question-only prior at decision time help?$^{\ddagger}$ & 15 Nov & Pre-specified & AUC for erased against ring-edited TELEA views higher by at least 0.03 with the interval above zero. & Not met ($+0.003$ $\ci{-0.018}{0.024}$). \\
Does rewarding inspection before the answer help on unedited pairs?$^{\ddagger}$ (two designs) & 22 Nov & Pre-specified, one seed each & On 38 REAL-Colon pairs not used in training by any design: natural ESG higher by at least 0.05, at least three more polyp frames answered present with a correct box, validation net utility not lower by more than 0.02 and lesion-free claims not higher by more than 2 percentage points. & Not met (natural ESG 0.500 and 0.263 against 0.632). \\
\bottomrule
\end{plantab}
\end{table}

\clearpage

\subsection{Evaluation sets}
\label{sec:s_data}

Table~\ref{tab:s_datasets} gives every evaluation set with its cases per source. The SIIM-ACR films of the fresh external set come from a 256-pixel copy of the dataset; for a post hoc check, 42 of its 48 pneumothorax films were found at original resolution. The plan named ChestX-Det and TCGA-LGG for the fresh internal set; no unused TCGA-LGG test-split patient remained, so the set holds ChestX-Det cases only. The hash comparison across sets covered the training and validation sets, the main test, the external datasets and the two fresh sets (6,849 unedited images); the sets added later are kept apart by source or by patient as stated in the table.

\begin{table}[!htb]
\caption{Evaluation sets, in full. For each set: its role, the number of cases, the source datasets (cases per source in parentheses), and whether any image of that source was used to train a model of this study. ``Other patients'' means that training and test images of the source come from different patients (or, where the source releases no patient identifiers, from different groups of near-identical images).}
\label{tab:s_datasets}
\centering
\footnotesize
\setlength{\tabcolsep}{3pt}
\renewcommand{\arraystretch}{1.08}
\begin{tabular}{@{}L{0.13\textwidth}L{0.17\textwidth}L{0.17\textwidth}L{0.27\textwidth}L{0.19\textwidth}@{}}
\toprule
\textbf{Evaluation set} & \textbf{Role} & \textbf{Cases} & \textbf{Source datasets} & \textbf{Source used in training?} \\
\midrule
Validation & Model selection; choice of decision rules; edit-site control & 540 cases, 1,473 views & The five sources of the main test & Yes (other patients) \\
Main test & Primary comparison & 507 cases (219 with a lesion, 288 lesion-free), 1,115 views & SLAKE (107) \citep{liu2021slake}; Kvasir-SEG with HyperKvasir (174) \citep{jha2020kvasir,borgli2020hyperkvasir}; ChestX-Det (91) \citep{liu2020chestxdet}; BUSI (57) \citep{al2020dataset}; TCGA-LGG (78) \citep{buda2019association} & Yes (other patients) \\
Tools not used in training & Main-test erasures and control edits remade with TELEA and Stable Diffusion & 125 cases common to all tools; 219 lesion cases with TELEA, 191 with Stable Diffusion & As main test & Yes (other patients); the two tools: no \\
External datasets & Datasets not used in training & 169 cases (119 with a lesion, 50 lesion-free) & CVC-ClinicDB (40) \citep{bernal2015cvc}; QaTa-COV19 (20) \citep{degerli2022qata}; BUS-BRA (45) \citep{gomez2024busbra}; BraTS-Africa 2023 (64) \citep{adewole2023bratsafrica} & No \\
Fresh internal & New patients of a training source; no model run on it before \ms{} was fixed & 76 cases (36 with a lesion, 40 lesion-free) & ChestX-Det \citep{liu2020chestxdet} & Yes (other patients) \\
Fresh external & New datasets; no model run on it before \ms{} was fixed & 316 cases & BraTS 2023 meningioma (82) \citep{labella2023meningioma}; BUS-UCLM (99) \citep{vallez2025busuclm}; SIIM-ACR pneumothorax, 256-pixel copy (96) \citep{siim2019pneumothorax}; ETIS-Larib (39) \citep{silva2014etis} & \ms: no. \msp{} and \msn: SIIM-ACR only (other patients)$^{a}$ \\
New SIIM-ACR patients & Pneumothorax detection on unedited films; first used for the models that preceded \msp & 200 patients (100 with, 100 without pneumothorax) & SIIM-ACR pneumothorax \citep{siim2019pneumothorax} & \ms: no. \msp{} and \msn: yes (other patients) \\
CheXthought test films & Pneumothorax detection on unedited films & 600 films (300 with, 300 without pneumothorax) & CheXthought \citep{sharma2026chexthought} on CheXpert Plus images \citep{chambon2024chexpertplus} & \ms{} and \msp: no. \msn: yes (other patients) \\
REAL-Colon pairs & Unedited same-patient pairs & 103 pairs from 46 of 60 videos, four centres & REAL-Colon \citep{biffi2024realcolon} & \ms{} and \msp: no. \msn: yes; answers cross-validated by video \\
LDPolypVideo pairs & Unedited same-patient pairs at a hospital not used in training & 116 pairs from 160 clips, one hospital & LDPolypVideo \citep{ma2021ldpolyp} & No \\
Chest X-ray follow-up pairs & Unedited same-patient pairs (secondary) & 44 pairs (planned version); 183 scored of 200 built (corrected version, post hoc) & NIH ChestX-ray14 \citep{wang2017chestx} & Yes: ChestX-Det is drawn from it. The planned exclusion of ChestX-Det patients matched no patient in either build \\
Standard question answering & Accuracy on ordinary medical visual question answering & 1,051 items & Test splits of VQA-RAD (451) \citep{lau2018vqarad}, SLAKE (300) \citep{liu2021slake} and PathVQA (300) \citep{he2020pathvqa} & Yes (training splits, last training stage) \\
\bottomrule
\multicolumn{5}{@{}p{0.97\textwidth}@{}}{$^{a}$\,SIIM-ACR draws on NIH ChestX-ray14, as ChestX-Det does; the hash comparison across sets found one image shared by the fresh external set and the validation set, and none shared with training.} \\
\end{tabular}
\end{table}

\subsection{Trained models}

Table~\ref{tab:s_models} lists the trained models. Three are named in the main text. \ms{} is the primary model. \msp{} adds pneumothorax films to every stage. \msn{} adds unedited colonoscopy frames and unedited pneumothorax films to the last stage of \msp. Before \msn{} was defined, each of its two additions was tested on its own; these two models, and a first variant that added pneumothorax films to the last stage only, appear in this table and in Table~\ref{tab:s_additions} and were not used further. No test result was used to choose among models: \ms{} and the training mix of \msn{} were chosen on validation data by rules fixed in advance (Table~\ref{tab:s_plan}).

\begin{table}[!htb]
\caption{Trained models. Each model is described by what its training adds to an earlier model. Validation values are net utility and lesion-free claims on the main validation set (540 cases, 1,473 views). All dates are in 2026.}
\label{tab:s_models}
\centering
\footnotesize
\setlength{\tabcolsep}{3pt}
\renewcommand{\arraystretch}{1.1}
\begin{tabular}{@{}L{0.16\textwidth}L{0.285\textwidth}L{0.11\textwidth}L{0.225\textwidth}L{0.15\textwidth}@{}}
\toprule
\textbf{Model} & \textbf{Training} & \textbf{Seeds} & \textbf{Status} & \textbf{Validation: net utility; lesion-free claims} \\
\midrule
\ms & \base{} with a low-rank adapter; three stages on automatically built counterfactual views (Section~\ref{subsec:models}) & 3 & Primary model. Procedure and selection rule fixed on 1 Nov, before its training data were built. & 0.613; 7.7\% \\
Last-stage variant (not used further) & The last stage of \ms{} repeated with films of 150 patients with and 150 without pneumothorax (SIIM-ACR) & 1 & Post hoc. Rule fixed on 4 Nov; not met. & 0.610; 10.0\% \\
\msp & All three stages of \ms{} repeated with the same data, settings and seed, and the same SIIM-ACR films in every stage & 1 & Post hoc remedy for the pneumothorax failure. Rule fixed on 15 Nov; met. & 0.632; 9.1\% \\
\msp{} with colonoscopy pairs (single addition) & The last stage of \msp{} repeated with 1,727 polyp frames and 1,881 polyp-free frames from REAL-Colon & 1; three cross-validation copies; four leave-one-centre-out copies & Post hoc. Guard and test fixed on 21 Nov; guard met. & 0.654; 9.1\% \\
\msp{} with CheXthought films (single addition) & The last stage of \msp{} repeated with 1,500 films with and 1,500 without pneumothorax from CheXthought & 1 & Post hoc. Guard and test fixed on 22 Nov; guard met. & 0.628; 9.5\% \\
\msn & The last stage of \msp{} repeated with both additions (plain mix) & 3; three cross-validation copies of the first seed & Designed post hoc. Mix rule, guard and tests fixed on 24 Nov; guard met. & 0.642; 9.1\% \\
\msn, alternative mix (not chosen) & As \msn, with erased images repeated until their share equals that in \msp & 1 & Candidate mix under the rule fixed on 24 Nov; not chosen. & 0.621; 10.0\% \\
\bottomrule
\end{tabular}
\end{table}

The two candidate mixes of \msn{} differ in the share of erased images in the last training stage. In \msp{} the last stage holds 6,217 views, of which 1,756 (28.2\%) are erased images and 420 (6.8\%) lesion-free images. Adding the unedited images in full gives 12,292 views with 1,223 erased images (9.9\%) and 3,801 lesion-free images (30.9\%). The alternative mix repeats erased images to the share they have in \msp: 15,428 views with 4,359 erased images (28.3\%) and 3,801 lesion-free images (24.6\%). The target share was computed by the training code from the settings of \msp{} and was not chosen. Over the main validation views and the SIIM-ACR validation patients together, net utility was 0.547 for the plain mix and 0.452 for the alternative, so the rule chose the plain mix. The alternative mix was then run on the main test, the fresh external set and the new SIIM-ACR patients only (Table~\ref{tab:s_mix}).

Cross-validated answers on REAL-Colon use three copies of a model, each trained without one third of the videos. The videos were divided by their number within each centre (videos 1 to 5, 6 to 10 and 11 to 15), so each copy was trained on 40 videos, and the division was fixed before any training frame was drawn. For the leave-one-centre-out analysis, four copies of \msp{} with colonoscopy pairs were each trained without all videos of one centre (45 training videos each). The models of the two-by-two analysis are first-stage models: the first stage with and without counterfactual views, each continued or not with the REAL-Colon training frames; the continued models were trained three times on the same division of the videos.

\subsection{Cost matrix and alternative cost matrices}

Net utility is the mean reward per view under the cost matrix of Table~\ref{tab:s_cost}. We fixed the matrix on 30 October 2026, before any model reported here was trained; it was not elicited from clinicians. The rows for erased and for lesion-free images are identical, so detecting an edit earns nothing. The second training stage of \ms{} kept the sampled answer with the highest reward under this matrix, and the comparison models were not shown it.

\begin{table}[!htb]
\caption{Cost matrix: reward per view by kind of view and final action. A wrong answer on an original or control-edited image costs $-4.0$ when the finding had been marked in advance as critical (brain tumour, malignant breast lesion, pneumothorax or mass on chest X-ray, and CT or MRI findings labelled cancer or tumour). A correct answer on an original or control-edited image earns a further $+0.5$ when its box has an intersection over union of at least 0.3 with the lesion and covers at most half of the image. An output that cannot be parsed scores $-1.0$, and each tool call beyond the first costs $0.05$.}
\label{tab:s_cost}
\centering
\small
\begin{tabular}{@{}lcccc@{}}
\toprule
\textbf{View} & \textbf{Correct answer} & \textbf{Wrong answer} & \textbf{Refer} & \textbf{Request for a better image} \\
\midrule
Original image; control edit & $+1.0$ & $-2.0$ ($-4.0$ if critical) & $-0.3$ & $-0.3$ \\
Erased image; lesion-free image & $+1.0$ (absent) & $-2.0$ (finding reported) & $+0.5$ & $-0.5$ \\
Degraded copy & $+0.25$ & $-2.0$ & $0.0$ & $+0.5$ \\
\bottomrule
\end{tabular}
\end{table}

To test whether the ranking of models depends on these weights, all models were scored again under five matrices: the matrix above with the penalty for a wrong answer set to $-2$, $-1$ and $-0.5$, and two variants of the referral reward (referral on lesion-free images paid $0$; referral paid $-0.3$ on every view). \ms{} ranked first among all compared models under all five. Always referring scores $+0.110$ under the matrix above, $-0.118$ when referral on lesion-free images pays $0$, and $-0.300$ when referral pays $-0.3$ on every view.

\subsection{A bound on what a decision rule can add}
\label{sec:s_bound}

A decision rule that acts on a fixed set of model outputs cannot create evidence sensitivity that those outputs do not contain. Let $S$ denote any signals computed from an image and its question, for example answer probabilities, confidence scores or the outputs of several agents, and let $\pi$ be any deterministic or randomised rule that maps $S$ to a final action. Let $\mu_{\mathrm{e}}$ and $\mu_{\mathrm{c}}$ be the distributions of $S$ over erased images and over control edits. Then
\begin{equation}
\label{eq:s_bound}
\mathrm{ESG}(\pi) \;\le\; \mathrm{TV}(\mu_{\mathrm{e}}, \mu_{\mathrm{c}}) \;=\; \sup_{A}\, \bigl|\mu_{\mathrm{e}}(A) - \mu_{\mathrm{c}}(A)\bigr| ,
\end{equation}
where TV is the total-variation distance and the supremum runs over sets of signal values. The statement follows from the data-processing inequality for total variation: the event ``$\pi(S)$ is absent or refer'' is a (possibly randomised) function of $S$, so the difference of its probabilities under $\mu_{\mathrm{e}}$ and $\mu_{\mathrm{c}}$, which is the ESG of $\pi$, cannot exceed the distance between the two distributions.

Two consequences were used in the study. First, a rule that refers more often can raise net utility, because referral is rewarded on erased and lesion-free images, without raising ESG. The 128 decision rules that recombine recorded outputs are of this kind. The 142 rules that add new evidence at test time (altered copies of the image, close-ups, a second model, answer-token probabilities) are not bound by the inequality for the original signals, which is why they were tested separately. Second, the right-hand side can be bounded from below: the hard decision of a linear probe on held-out cases is one event $A$, so its true-positive rate minus its false-positive rate is a lower bound on the distance. This is the measured bound of Table~\ref{tab:s_plan}; it is linear, and the true distance may be larger.

\subsection{Construction of the test views and pairs}
\label{sec:s_views}

\emph{Automated checks of edited views.} The contrast between the erased region and a surrounding ring of 6 to 14 pixels had to fall to 35\% of its value before the edit or less (or to 1.0, if that is larger), and an edit detector had to score the erased region below 0.5. A control edit could change no lesion pixel. Near-duplicate images across splits were removed by 64-bit difference hashing and pixel correlation. Views were filtered until a texture classifier separated erasures from control edits with an AUC of at most 0.60; the value on the main test was 0.56. Degraded copies were made by blurring, added noise, eightfold down-sampling and up-sampling, strong JPEG compression, motion blur, or substitution of an image of another imaging type.

\emph{Blank-image control, low-visibility erasures and question types.} The blank-image control uses the 183 main-test lesion cases whose yes/no question is answered ``yes'' on the original image. The blank view is a uniform mid-grey image of the same size. The swapped view is a lesion-free main-test image of another imaging type, chosen by a fixed hash of the case; it exists for 141 of the 183 cases, because swaps within one imaging type were removed after a coding fault. For the low-visibility analysis, a ResNet-18 was trained for five epochs to separate control-edited from unedited training views, both of which show the lesion, and then scored every main-test view. An erasure is low-visibility when its score is at or below the median score of the main-test control edits (111 of 219 erasures); existing answers were used and no model was run again. A question is generic when it asks about an ``abnormality'' or about a ``healthy'' or ``normal'' appearance, and finding-specific otherwise (114 and 105 lesion cases; 128 and 160 lesion-free cases).

\emph{REAL-Colon pairs.} A track is a run of boxed frames at least 1~s long, with gaps under 2~s merged. For each of a video's three longest tracks, the polyp frame comes from the middle half of the track, with the largest box at least 1\% of the frame. The paired frame comes from the same video, 10 to 30~s before the track starts and at least 10~s from any boxed frame; it was taken after the track when fewer than three frames qualified before it (28 of the 103 pairs; the written plan said ``no frame'', and the selection code fixed with the plan, which was followed, said ``fewer than three''). On each side, the sharpest of 12 evenly spaced candidates with a mean brightness of 40 to 215 was kept. The context-matched frame lies within 10~s of the ends of the track and 3 to 10~s from the nearest boxed frame, and was found for 96 pairs. Two limits were stated in advance: the two frames of a pair differ in more than the polyp (view, lighting, bowel preparation), and a paired frame may hold a small polyp without a box, because the dataset boxes the polyps that were removed.

\emph{LDPolypVideo pairs.} The dataset holds 160 clips from the First Affiliated Hospital of Anhui Medical University, with 40,186 labelled frames (33,875 with a polyp and 6,311 polyp-free); both of its parts were used, and nothing was trained on them. The clips state no frame rate, so the rules were written in frames, with one pair per clip. A track is a run of polyp frames at least 10 frames long with gaps of at most 5 frames, and the longest track of the clip is used. The polyp frame is chosen as for REAL-Colon. The polyp-free frame is marked polyp-free and lies at least 5 frames from any frame that is not. Of the 160 clips, 116 gave a pair; 39 had no polyp-free frame at least 5 frames from a polyp frame, and 5 had no polyp frame with a box of at least 1\% of the frame. The plan required at least 40 pairs for the decision rule to apply. The repository of the dataset states no licence, and its images are not redistributed.

\emph{Chest films.} CheXthought films are frontal views with the question ``Is there a pneumothorax in this chest X-ray?''. A film is positive when the CheXpert Plus label for pneumothorax is positive and at least one CheXthought radiologist states a pneumothorax, without expressing doubt, and marks it with click points. A film is negative when the label is negative or blank and no radiologist states or suspects a pneumothorax. A box counts as correct when its intersection over union with the box around the click points is at least 0.3. For the follow-up pairs from NIH ChestX-ray14, the lesion film is one whose only label is pneumothorax, pleural effusion, nodule or mass, and the paired film is the nearest study of the same patient and view position labelled ``No Finding'', with one pair per patient and up to 50 pairs per finding; the labels are mined from reports. The planned exclusion of patients in any ChestX-Det split matched no patient in either version. In the planned version a check for near-copies wrongly removed 156 of 200 pairs, which left 44; the corrected version fixed that check and gave 200 pairs, of which 183 were scored. The corrected version is post hoc.

\subsection{Training details}
\label{sec:s_training}

\emph{Development versions.} Before the training procedure of \ms{} was fixed, the following versions of the trained model had been evaluated on the main test: an adapter trained only to refer; first-stage models with and without counterfactual views, at two seeds; a model after rejection-sampling fine-tuning; and that model continued with general question-answering data, which was the trained model of the earlier series. The last two were also run on the external datasets, and the earlier series was run with all four editing tools on 125 main-test cases, after which tissue transplant was added to training. The training of \ms{} was changed in response to weaknesses these versions showed: false findings on lesion-free images, a regression on SLAKE CT and MRI, lower detection of real lesions at the external datasets, weak chest X-ray and small-lesion results, and a possible dependence on one editing tool. The record does not give the number of versions run.

\emph{Data preparation.} Training used 1,508 SLAKE and colonoscopy cases and 1,792 chest X-ray, breast ultrasound and brain MRI cases, with 427 and 690 lesion-free images; the validation set holds 540 cases (1,473 views). Excluded before splitting were 179 of 642 SLAKE masks (grey levels not listed in the mask key), 623 diffuse chest X-ray findings and 179 lesions outside the size limits in ChestX-Det, 88 BUSI lesions too large to erase, and 364 TCGA-LGG slices with little brain. The lesion-free colonoscopy images are 427 HyperKvasir frames of normal landmarks (caecum, retroflexed rectum and ileum). Every patient or image group that holds a main-test image was removed from the training and validation splits of \ms.

\emph{Adapter and training examples.} The adapter is a low-rank adapter (LoRA) \citep{hu2022lora} of rank 16 ($\alpha = 32$, dropout 0.05) on all linear layers of the language model; the vision encoder is frozen. Each training example is a sequence of turns built from the labels alone: a \texttt{zoom} call on a jittered box around the lesion (always when the lesion covers less than 2\% of the image), a \texttt{window} call on 30\% of greyscale images, and then the expected final action with short template reasoning. Half of the training questions were reworded with four paraphrases; three further paraphrases were kept for testing.

\emph{Stages.} Stage 1 is supervised fine-tuning on 16,000 views for one pass (3,706 optimiser steps) at a learning rate of $2 \times 10^{-4}$. In stage 2, four answers per view are sampled at temperature 1.2, the successful answer with the highest reward is kept, views that the model mostly failed are repeated three times, and the model is fine-tuned on these at a learning rate of $5 \times 10^{-5}$. Stage 3 is fine-tuning at the same learning rate on 5,000 views that mix counterfactual views with questions from the training splits of VQA-RAD, SLAKE and PathVQA; views whose expected answer is ``absent'' make up at least 35\% of the mix. Seeds change the initialisation of the adapter, the order of the data and the random choices in the training examples; the seeds of a model share their training data. For the three-seed vote of \msn, the answer of each seed on a view is classed as present, absent, refer or other (a request for a better image, or an output that cannot be parsed). The vote is the class that two or three seeds chose. When all three differ, a referral is taken if one seed refers, and otherwise a request for a better image.

\emph{Hardware and software.} Each stage of \ms{} took about 4 hours on one NVIDIA A100 graphics processor with 40~GB of memory, with PyTorch 2.11, Transformers 5.22.1 and PEFT 0.21.2. The final runs of \msn{} used PyTorch 2.14.1, Transformers 5.24.1 and PEFT 0.22.0 on several graphics processors; on an A100 with 80~GB, two jobs at a time, the last stages of its two candidate mixes took 5 and 6.5 hours.

\emph{Pneumothorax films in \msp.} The SIIM-ACR patients were divided at random, with a fixed seed, into training patients (150 with and 150 without pneumothorax), validation patients (98 cases, 242 views) and the 200 new patients used for testing; none is in the fresh external set.

\emph{Unedited images in \msn.} From every REAL-Colon video, polyp frames were drawn from up to ten tracks: one frame per 2~s of track, at most six per track, with the largest box at least 0.5\% of the frame. About as many polyp-free frames were drawn per video: near a track by the rule of the test pairs, context-matched, or at least 30~s from any boxed frame; a video without tracks gave twelve. This yielded 1,727 polyp frames, 772 of them with a polyp under 2\% of the frame, and 1,881 polyp-free frames. CheXthought films are frontal views only. The zoom box of a positive film is the bounding box of the radiologists' click points, padded by 5\% of the image on each side and at least 15\% of the image in width and height. Patients were divided by a fixed hash into a training pool (80\%) and a test pool (20\%); training used 1,500 positive and 1,500 negative films and testing 300 and 300, at a long side of at most 768 pixels. CheXpert Plus images and CheXthought annotations are used under their research-use agreements and are not redistributed.

\subsection{Statistics}
\label{sec:s_stats}

The $p$ value of a paired comparison is twice the smaller share of the 2,000 resampled differences on either side of zero, so the smallest non-zero value it can take is 0.001; a value of zero is printed as $p < 0.001$. Resampling patients or image groups in place of cases (post hoc; 416 groups on the main test, 155 at the external datasets, 75 in the fresh internal set and 230 in the fresh external set) widened intervals by at most 0.01 on the main test and 0.03 on the fresh external set. All 27 primary comparisons kept $p < 0.001$, and one other comparison crossed 0.05: fresh-external ESG of \ms{} against \opus, $p = 0.037$ with cases and $p = 0.060$ with patients resampled. Outside the 27 primary comparisons, $p$ values are not corrected for multiplicity and are read with their intervals.

\subsection{Reader study protocol}
\label{sec:s_readers}

\emph{Trainee round.} Two trainee physicians, one in internal medicine and one in radiology (recorded as one year each), read 128 main-test images: 32 each of unedited lesion images, LaMa erasures, LaMa control edits and unedited lesion-free images. The images were shown one at a time in random order on a web page without labels. For each image the reader answered whether the image appeared to be artificially edited, and whether the finding named in the clinical question was present (yes, no or cannot tell). Readers gave no reasons. They had no access to the labels, the kinds of edit or each other's answers. The criterion, fixed before any answer was collected, was that the upper limit of the 95\% interval of the share of erased images with the finding reported is at most 10\%.

\emph{Re-scoring of model answers on the trainees' answers.} This analysis was fixed on 14 November 2026, before any model was scored in this way. An erasure is \emph{clean for both readers} when neither trainee answered that the finding was present, and \emph{called present by both} when both did. For every model run on the main test, the existing answers on these images were used; no model was run again. The nine comparison models were pooled, with a bootstrap over images (2,000 resamples). The rule is in Table~\ref{tab:s_plan}.

\emph{Specialist round.} The round was fixed on 15 November 2026. The set was planned as 216 images, 36 of each of six kinds; 212 were used, because only 32 Stable Diffusion control edits were available from distinct cases. Radiologists were required to be board certified with at least five years of practice after certification, and a gastroenterologist or endoscopist was sought for colonoscopy. Each reader received a personal link with an opaque code. The invitation stated that the study evaluates whether automatically edited medical images still show a finding, that the 212 images come from public research datasets and carry no patient information, that each image has two questions (whether it has been digitally edited, and the clinical question shown), and that answers should rest on the image alone. Consent to the use of the answers under a reader code was recorded on the page before the first image. Readers recorded their specialty, whether they are board certified and the years since certification, and whether they had taken part in the earlier round. Each answer was saved as it was given, and the latest answer per image counts. The order of images was shuffled for each reader. An optional free-text note could be added to any image.

The closing rule was fixed on 18 November 2026, before any answer was examined: the round closes when two board-certified radiologists have read all 212 images, or on 10 January 2027, whichever comes first; the date is not moved after any answer is seen; and with fewer than two radiologists the round is reported as incomplete, with what was collected. The gastroenterologist read all 212 images. As planned, these answers count towards the criterion for colonoscopy images only; answers on other imaging types are reported as outside the reader's specialty and are kept out of the radiologist majority. On the same day, before any answer was opened, an interim analysis of the gastroenterologist's colonoscopy answers was added to the plan. One planned part of it, the scoring of models on the erasures this reader called clean, was not run for that interim analysis. A board-certified radiologist (8 years of practice) then read all 212 images. A look at the radiologist's answers before the round closes was not planned and is reported as interim; the scoring of models on the erasures accepted by the specialist for each imaging type was fixed on 26 November 2026, before it was run, and is in Supplementary Table~\ref{tab:s_specialist}. Readers who complete the reading and review the manuscript are offered authorship; all readers read without access to labels, kinds of edit or other readers' answers. The answer keys of both rounds are withheld while the specialist round is open.

\subsection{Prompts and model settings}
\label{sec:s_prompts}

\emph{System prompt.} Every model, in every analysis that uses the agent protocol, received the following system prompt, followed by the image and the line ``Question:'' with the question of the case. The prompt addresses the model as MedScope, the name also used for the benchmark and the trained models.

\begin{quote}
\footnotesize\ttfamily\raggedright
You are MedScope, a careful medical imaging assistant used for research. You look at a medical image and answer a question about it from what the image actually shows: you report a finding only when you can see it, and you say so when it is not there.\par
\smallskip
On each turn, think briefly in plain text, then write exactly ONE action as a tagged JSON object and stop. After a tool call, stop and wait: you will be shown the result before you continue.\par
\smallskip
Example of a tool turn:\par
The lesion may be small, so I will look closer at the upper left. <tool>{\char123}"name": "zoom", "box": [100, 150, 450, 500]{\char125}</tool>\par
Example of a final turn:\par
The mass is clearly visible. <final>{\char123}"action": "answer", "answer": "Yes", "box": [180, 220, 320, 360], "confidence": 0.85{\char125}</final>\par
\smallskip
Tools (at most 3 calls per case; boxes are integers 0-1000 on the ORIGINAL full image, as [x1, y1, x2, y2]):\par
<tool>{\char123}"name": "zoom", "box": [x1, y1, x2, y2]{\char125}</tool>\ \ look closer at a region\par
<tool>{\char123}"name": "window", "center": 128, "width": 100{\char125}</tool>\ \ adjust brightness/contrast (0-255 scale)\par
\smallskip
Final actions (exactly one, then stop):\par
<final>{\char123}"action": "answer", "answer": "...", "box": [x1, y1, x2, y2], "confidence": 0.0-1.0{\char125}</final>\par
\hspace*{1.5em}when the image lets you answer. If the finding is visible, the box marks it on the original image. If the finding asked about is not there, say so and leave out the box: "No" to "Is there ...?", "Yes" to "Is ... healthy?" or "Is ... normal?", "No abnormality" to an open question.\par
<final>{\char123}"action": "clarify", "question": "..."{\char125}</final>\par
\hspace*{1.5em}when the image is unusable (too blurred, noisy, low resolution, or not the anatomy asked about).\par
<final>{\char123}"action": "refer", "reason": "..."{\char125}</final>\par
\hspace*{1.5em}when the image is usable but you cannot decide either way.\par
\smallskip
Keep answers short, using the wording a radiologist would use (for example "Yes", "No", "Liver", "Left lung"). Never report a finding you cannot see.
\end{quote}

\emph{Prompting conditions.} Five further conditions were applied to the three reference models on 150 main-test cases. The cautionary prompt appends to the system prompt: ``Important: a question may name a finding that is not in the image. Do not assume the question is correct. Before answering, check carefully whether the finding is actually visible. If you cannot clearly see it, say it is not there (`No'). Reporting a finding that is not visible is a serious error.'' The condition with solved examples adds the cautionary text and four solved training cases of other patients before the case. The plain prompt replaces the system prompt by ``You are a medical imaging assistant. Look at the image and answer the question. Keep the answer short.'' and offers no tools. The remaining two conditions use the provider's highest reasoning setting and highest image-resolution setting. For the open-ended request, the original and the erased image of each case were shown with the plain prompt and the request ``Describe any abnormal findings in this image.'', with one reply and no tools. The cases are the 36 with a finding-specific question among the 150 cases of the prompting conditions. A text-only judge (\gpt), given neither the image nor the name of the model, marked whether each reply states that the finding named in the yes/no question of the case is present. The free-text ESG is the share of original images whose reply reports the finding minus the share of erased images whose reply still reports it.

\emph{Comparison models.} The nine comparison models were called through the providers' interfaces under the identifiers given in Section~\ref{subsec:models}, with each provider's default settings (reasoning switched on where that is the default), and each was run once per set. The first comparison models run, on 150 main-test cases, were \texttt{gpt-6-sol} and \gem. All nine were then run on the complete main test on 31 October 2026. The three reference models first appear in the record later the same day, when they were run on the external datasets and under the cautionary prompt; the record gives no criterion for their choice, and the plan of 2 November 2026 names them as the set for the later analyses. They include the comparison model with the highest main-test ESG (\gem, 0.470); \texttt{claude-sonnet-5-5} (0.402) and \texttt{gpt-6-sol} (0.342) had a higher main-test ESG than \opus{} (0.329) and \gpt{} (0.306). All nine were run again for the blank-image control on 22 November 2026, and the three reference models were run on the later sets and conditions between 31 October and 26 November 2026 (the fresh sets on 3 November, the REAL-Colon pairs on 21 November, the CheXthought films on 24 November and the LDPolypVideo pairs on 26 November) under the same identifiers. Images were sent as PNG files with a long side of at most 768 pixels. The output limit was 16,000 tokens (64,000 at the highest reasoning setting). A request that failed was repeated up to six times and then scored as an output that cannot be parsed; the share of such outputs per model was recorded only for the first run of \texttt{gpt-6-sol} and \gem{} on 150 cases (none). The highest reasoning setting is reasoning effort ``high'' (OpenAI), effort ``max'' with adaptive thinking (Anthropic) and thinking level ``high'' (Google). The highest image-resolution setting is image detail ``high'' (OpenAI) and media resolution ``high'' (Google); Anthropic offers no such setting. The same system prompt, images, tool results and turn structure were used as for the open-weight models. Frames of REAL-Colon and LDPolypVideo and the CheXthought test films were sent to the three providers for these runs. Because the datasets are public, nothing can be said about whether the comparison models were trained on them.

\emph{Open-weight models.} The six open-weight models are \base{} (the base of the trained models, without fine-tuning), MedGemma 1.5 4B \citep{sellergren2025medgemma}, Lingshu-7B \citep{xu2025lingshu}, HuatuoGPT-Vision-7B \citep{chen2024huatuogpt}, Qwen2.5-VL-7B \citep{bai2025qwen25vl} and Qwen3-VL-8B. They were run locally in 16-bit floating point (bfloat16) with greedy decoding, without any repetition penalty from their own configuration, under the same protocol. The trained models were run in the same way. Free-text replies that carried no action tag were read by the finding asked about first (for example, ``no pneumothorax'' counts as absent); before this reading was adopted, such replies made up 35\% of the replies of Qwen2.5-VL-7B and all replies of Lingshu-7B and HuatuoGPT-Vision-7B.

\section{Additional results: counterfactual test sets}\label{sec:s_results_a}

The tables in this section follow the order of Sections~\ref{subsec:res_main} to~\ref{subsec:res_external}.
Measures are those of Section~\ref{subsec:measures}: lesion images correct (share of original and control-edited images answered correctly), erased-lesion claims (share of erased images on which the finding was reported), lesion-free claims (share of unedited lesion-free images on which a finding was reported), the evidence sensitivity gap (ESG) and net utility.
Unless a caption says otherwise, brackets are 95\% bootstrap intervals over cases (2{,}000 resamples), differences are paired by case, and ``denied'' means answered ``absent'' or referred.
Lesion images correct is given as a proportion and claims as percentages; a difference between two claim rates is a difference of proportions unless it is marked as percentage points.
CI: confidence interval; AUC: area under the receiver operating characteristic curve; SD: standard deviation.
The three reference models are \gem, \opus{} and \gpt.
A first-stage model is a model after the first of the three training stages.
\providecommand{\Chd}[2]{\parbox[b]{#1}{\centering\strut #2\strut}}

\subsection{Main test}

\begin{table}[!htbp]
\caption{Main test by imaging type (507 cases), for \ms{} and the three reference models. Cases are lesion cases plus lesion-free cases. Claims are percentages. The values are point estimates; no intervals were computed for the strata.}
\label{tab:s_modality}
\centering
\small
\begin{adjustbox}{max width=\linewidth}
\begin{tabular}{llccccc}
\toprule
Imaging type (cases) & Model & Lesion images correct & Erased-lesion claims (\%) & Lesion-free claims (\%) & ESG & Net utility \\
\midrule
Colonoscopy ($83 + 91$) & \ms & $0.96$ & $1$ & $0$ & $0.95$ & $1.06$ \\
 & \gem & $1.00$ & $31$ & $11$ & $0.69$ & $0.83$ \\
 & \opus & $0.96$ & $40$ & $37$ & $0.57$ & $0.53$ \\
 & \gpt & $0.95$ & $58$ & $37$ & $0.40$ & $0.40$ \\
\addlinespace
Brain MRI ($36 + 42$) & \ms & $0.88$ & $14$ & $0$ & $0.75$ & $0.75$ \\
 & \gem & $0.89$ & $47$ & $0$ & $0.42$ & $0.56$ \\
 & \opus & $0.93$ & $61$ & $14$ & $0.31$ & $0.51$ \\
 & \gpt & $0.83$ & $31$ & $10$ & $0.50$ & $0.43$ \\
\addlinespace
Breast ultrasound ($42 + 15$) & \ms & $0.88$ & $14$ & $7$ & $0.74$ & $0.78$ \\
 & \gem & $0.89$ & $48$ & $13$ & $0.40$ & $0.53$ \\
 & \opus & $0.92$ & $62$ & $40$ & $0.29$ & $0.40$ \\
 & \gpt & $0.98$ & $83$ & $60$ & $0.12$ & $0.28$ \\
\addlinespace
Chest X-ray ($41 + 50$) & \ms & $0.70$ & $17$ & $20$ & $0.51$ & $0.40$ \\
 & \gem & $0.65$ & $49$ & $6$ & $0.20$ & $0.21$ \\
 & \opus & $0.60$ & $66$ & $20$ & $-0.05$ & $-0.12$ \\
 & \gpt & $0.79$ & $66$ & $26$ & $0.15$ & $0.14$ \\
\addlinespace
SLAKE CT and MRI ($17 + 90$) & \ms & $0.53$ & $12$ & $12$ & $0.47$ & $0.22$ \\
 & \gem & $0.74$ & $41$ & $12$ & $0.35$ & $0.38$ \\
 & \opus & $0.76$ & $59$ & $13$ & $0.24$ & $0.32$ \\
 & \gpt & $0.68$ & $47$ & $22$ & $0.29$ & $0.17$ \\
\bottomrule
\end{tabular}
\end{adjustbox}
\end{table}

\begin{table}[!htbp]
\caption{Training seeds of \ms{} (pre-specified). The whole three-stage training was repeated with two further seeds; the first seed is \ms. Upper part: every measure on the main test and ESG and net utility on the other sets. Lower part: main-test differences from each reference model, paired by case. SD: standard deviation over the three seeds (variation between trainings); brackets: 95\% bootstrap intervals over cases (variation between test cases).}
\label{tab:s_seeds}
\centering
\small
\begin{adjustbox}{max width=\linewidth}
\begin{tabular}{llcccc}
\toprule
Test set (cases) & Measure & First seed (\ms) & Second seed & Third seed & Mean (SD) \\
\midrule
Main test (507) & Lesion images correct & $0.849$ & $0.861$ & $0.840$ & $0.850$ ($0.010$) \\
 & Erased-lesion claims (\%) & $9.6$ & $11.4$ & $10.5$ & $10.5$ ($0.9$) \\
 & Lesion-free claims (\%) & $7.6$ & $17.0$ & $9.0$ & $11.2$ ($5.1$) \\
 & ESG & $0.758$ $\ci{0.701}{0.815}$ & $0.744$ $\ci{0.680}{0.805}$ & $0.731$ & $0.744$ ($0.014$) \\
 & Net utility & $0.732$ & $0.672$ & $0.698$ & $0.701$ ($0.030$) \\
External datasets (169) & ESG & $0.597$ & $0.605$ $\ci{0.505}{0.704}$ & $0.605$ & $0.602$ ($0.005$) \\
 & Net utility & $0.530$ & $0.542$ & $0.500$ & $0.524$ ($0.022$) \\
Fresh internal (76) & ESG & $0.472$ & $0.472$ $\ci{0.314}{0.641}$ & $0.361$ & $0.435$ ($0.064$) \\
 & Net utility & $0.402$ & $0.386$ & $0.283$ & $0.357$ ($0.065$) \\
Fresh external (316) & ESG & $0.393$ & $0.333$ $\ci{0.253}{0.410}$ & $0.375$ & $0.367$ ($0.031$) \\
 & Net utility & $-0.053$ & $-0.099$ & $-0.045$ & $-0.066$ ($0.029$) \\
\midrule
\multicolumn{6}{l}{\textit{Main test, seed minus reference model}} \\
ESG & minus \gem & $0.288$ $\ci{0.208}{0.367}$ & $0.274$ $\ci{0.190}{0.359}$ & $0.260$ $\ci{0.178}{0.341}$ & \\
 & minus \opus & $0.429$ $\ci{0.345}{0.507}$ & $0.416$ $\ci{0.333}{0.495}$ & $0.402$ $\ci{0.311}{0.484}$ & \\
 & minus \gpt & $0.452$ $\ci{0.364}{0.539}$ & $0.438$ $\ci{0.350}{0.526}$ & $0.425$ $\ci{0.332}{0.514}$ & \\
Erased-lesion claims & minus each of the three & $-0.315$ to $-0.521$ & $-0.297$ to $-0.475$ & $-0.306$ to $-0.484$ & \\
Net utility & minus each of the three & $0.168$ to $0.477$ & $0.107$ to $0.363$ & $0.134$ to $0.390$ & \\
Lesion-free claims & minus \gem & $-0.014$ & $0.080$ $\ci{0.034}{0.130}$ & none (both $0.090$) & \\
\bottomrule
\end{tabular}
\end{adjustbox}
\\[2pt]\parbox{\linewidth}{\footnotesize Note. Erased-lesion claims and lesion-free claims in the lower part are differences of proportions. The erased-lesion-claim and net-utility ranges of the first seed are over all nine comparison models, whose intervals are in Supplementary Table~\ref{tab:s_primary}. For the first seed all 27 differences from the nine comparison models have $p < 0.001$; for the second and third seeds all ESG and erased-lesion-claim differences from the reference models have $p < 0.001$ and their net-utility differences $p \le 0.015$. Lesion-free claims minus \gem: $p = 0.59$ for the first seed and $p = 0.002$ for the second; lesion-free claims of the second and third seeds were lower than those of \opus{} and \gpt{} ($p < 0.001$ for the third). The pre-specified criterion, every seed above each reference model on main-test ESG with the interval above zero, was met.}
\end{table}

\begin{table}[!htbp]
\caption{The 27 primary comparisons: \ms{} minus each comparison model on the main test (507 cases), paired by case. Brackets are 95\% intervals from 2,000 bootstrap resamples of cases. Erased-lesion claims are differences in percentage points. Every difference has $p < 0.001$, below the Bonferroni-corrected level of $0.05/27$.}
\label{tab:s_primary}
\centering
\small
\begin{adjustbox}{max width=\linewidth}
\begin{tabular}{lccc}
\toprule
Comparison model & ESG & Erased-lesion claims (points) & Net utility \\
\midrule
\texttt{gpt-6-sol} & $0.416$ $\ci{0.341}{0.489}$ & $-42.0$ $\ci{-49.1}{-34.6}$ & $0.472$ $\ci{0.389}{0.554}$ \\
\gpt & $0.452$ $\ci{0.364}{0.539}$ & $-49.3$ $\ci{-56.2}{-42.6}$ & $0.424$ $\ci{0.329}{0.522}$ \\
\texttt{gpt-6.1-sol} & $0.434$ $\ci{0.354}{0.520}$ & $-49.3$ $\ci{-56.6}{-42.3}$ & $0.456$ $\ci{0.360}{0.552}$ \\
\texttt{claude-fable-5-1} & $0.416$ $\ci{0.339}{0.496}$ & $-42.5$ $\ci{-50.0}{-35.5}$ & $0.477$ $\ci{0.384}{0.566}$ \\
\opus & $0.429$ $\ci{0.345}{0.507}$ & $-44.3$ $\ci{-51.5}{-36.8}$ & $0.374$ $\ci{0.283}{0.466}$ \\
\texttt{claude-sonnet-5-5} & $0.356$ $\ci{0.284}{0.431}$ & $-38.8$ $\ci{-45.9}{-31.8}$ & $0.351$ $\ci{0.267}{0.437}$ \\
\gem & $0.288$ $\ci{0.208}{0.367}$ & $-31.5$ $\ci{-38.5}{-24.3}$ & $0.168$ $\ci{0.083}{0.254}$ \\
\texttt{gemini-3.1-pro-preview} & $0.374$ $\ci{0.298}{0.448}$ & $-42.0$ $\ci{-49.1}{-35.0}$ & $0.426$ $\ci{0.337}{0.519}$ \\
\texttt{gemini-3.8-flash} & $0.489$ $\ci{0.407}{0.566}$ & $-52.1$ $\ci{-59.6}{-44.6}$ & $0.470$ $\ci{0.380}{0.560}$ \\
\bottomrule
\end{tabular}
\end{adjustbox}
\end{table}

\begin{table}[!htbp]
\caption{Blank-image control (pre-specified): 183 main-test lesion cases whose question is answered ``yes'' on the original image. Blank image: a uniform grey image of the same size. Swapped image: an unedited lesion-free main-test image of another imaging type (available for 141 of the 183 cases). Entries are the share of images on which the model reported the finding, answered ``absent'', or asked for a better image. The last column is the paired difference between the share of erased images and the share of blank images on which the finding was reported, in percentage points.}
\label{tab:s_blank}
\centering
\small
\begin{adjustbox}{max width=\linewidth}
\begin{tabular}{lcccc}
\toprule
 & Blank image (\%) & Swapped image (\%) & Erased image (\%) & \Chd{2.6cm}{Erased minus blank (percentage points)} \\
Model & reported / absent / better image & reported / absent / better image & reported & \\
\midrule
\ms & $0$ / $100$ / $0$ & $2.1$ / $21.3$ / $76.6$ & $10.9$ & $10.9$ $\ci{6.6}{15.8}$ \\
\base & $0$ / $18.6$ / $81.4$ & $2.1$ / $18.4$ / $79.4$ & $40.4$ & $40.4$ $\ci{33.3}{47.5}$ \\
Nine comparison models (range) & $0$ / $0$ / $100$ (every model) & $0$ to $2.1$ / $0$ to $17.0$ / $82$ to $100$ & $42.1$ to $56.8$ & $42.1$ to $56.8$, all $p < 0.001$ \\
\midrule
\multicolumn{5}{l}{\textit{Erased image, finding reported (\%), by comparison model}} \\
\multicolumn{5}{L{15.5cm}}{\texttt{gpt-6-sol} $48.1$; \gpt{} $54.1$; \texttt{gpt-6.1-sol} $54.6$; \texttt{claude-fable-5-1} $52.5$; \opus{} $54.1$; \texttt{claude-sonnet-5-5} $48.1$; \gem{} $42.1$; \texttt{gemini-3.1-pro-preview} $52.5$; \texttt{gemini-3.8-flash} $56.8$.} \\
\bottomrule
\end{tabular}
\end{adjustbox}
\end{table}

\begin{table}[!htbp]
\caption{Main test by question type (pre-specified). A question is generic if it asks about an ``abnormality'' or whether the image is ``healthy'' or ``normal'', and finding-specific otherwise: 114 finding-specific and 105 generic lesion cases, with 128 and 160 lesion-free cases. (A) \ms, \msn{} and the three reference models, with the paired difference in ESG of each trained model from each reference model. (B) ESG of the other comparison models and of the base model on finding-specific questions (in the note below the table). (C) ESG on finding-specific questions by imaging type, with the reference models against which the difference from \ms{} had $p < 0.05$ (uncorrected).}
\label{tab:s_qtype}
\centering
\small
\begin{adjustbox}{max width=\linewidth}
\begin{tabular}{llccccc}
\toprule
 & & & & & \multicolumn{2}{c}{ESG: trained model minus model in the row} \\
\cmidrule(lr){6-7}
(A) Question type & Model & ESG & \Chd{1.9cm}{Erased-lesion claims (\%)} & \Chd{1.7cm}{Lesion-free claims (\%)} & \ms & \msn \\
\midrule
Finding-specific & \ms & $0.728$ $\ci{0.649}{0.807}$ & $9.6$ & $7.8$ & & $-0.018$ $\ci{-0.088}{0.061}$ \\
(114 lesion cases) & \msn & $0.711$ $\ci{0.632}{0.789}$ & $9.6$ & $9.4$ & & \\
 & \gem & $0.439$ $\ci{0.333}{0.553}$ & $37.7$ & $8.6$ & $0.289$ $\ci{0.175}{0.412}$ & $0.272$ $\ci{0.149}{0.395}$ \\
 & \opus & $0.342$ $\ci{0.254}{0.430}$ & $52.6$ & $28.9$ & $0.386$ $\ci{0.280}{0.500}$ & $0.368$ $\ci{0.263}{0.482}$ \\
 & \gpt & $0.351$ $\ci{0.254}{0.447}$ & $50.9$ & $26.6$ & $0.377$ $\ci{0.246}{0.509}$ & $0.360$ $\ci{0.237}{0.491}$ \\
\addlinespace
Generic & \ms & $0.790$ $\ci{0.714}{0.867}$ & $9.5$ & $7.5$ & & $-0.010$ $\ci{-0.086}{0.067}$ \\
(105 lesion cases) & \msn & $0.781$ $\ci{0.695}{0.857}$ & $8.6$ & $8.1$ & & \\
 & \gem & $0.505$ $\ci{0.410}{0.590}$ & $44.8$ & $9.4$ & $^{\dagger}$ & $0.276$ $\ci{0.162}{0.400}$ \\
 & \opus & $0.314$ $\ci{0.210}{0.419}$ & $55.2$ & $19.4$ & $^{\dagger}$ & $0.467$ $\ci{0.362}{0.571}$ \\
 & \gpt & $0.257$ $\ci{0.171}{0.343}$ & $67.6$ & $28.7$ & $^{\dagger}$ & $0.524$ $\ci{0.400}{0.629}$ \\
\midrule
\parbox[b]{3.4cm}{\raggedright\strut (C) Finding-specific questions by imaging type: ESG\strut} & \Chd{2.4cm}{Lesion cases, specific / generic} & \ms & \gem & \opus & \gpt & \Chd{3.6cm}{\ms{} higher with $p < 0.05$ against} \\
\midrule
Breast ultrasound & 42 / 0 & $0.738$ & $0.405$ & $0.286$ & $0.119$ & all three \\
Colonoscopy & 24 / 59 & $0.917$ & $0.792$ & $0.625$ & $0.625$ & \opus, \gpt \\
Brain MRI & 18 / 18 & $0.778$ & $0.500$ & $0.389$ & $0.611$ & \gem, \opus \\
Chest X-ray & 20 / 21 & $0.550$ & $0.150$ & $0.100$ & $0.300$ & \opus \\
SLAKE CT and MRI & 10 / 7 & $0.500$ & $0.200$ & $0.300$ & $0.300$ & none \\
\bottomrule
\end{tabular}
\end{adjustbox}
\\[2pt]\parbox{\linewidth}{\footnotesize (A) All differences from reference models have $p < 0.001$; no difference between the two trained models was shown ($p = 0.719$ and $p = 0.896$). The pre-specified criterion, \ms{} above each reference model on finding-specific questions with the interval above zero, was met. $^{\dagger}$\,For \ms{} on generic questions only the range of the differences from the nine comparison models is recorded: $0.286$ to $0.552$, all $p < 0.001$. (B) Finding-specific questions, ESG of the other models: \texttt{claude-sonnet-5-5} $0.386$; \texttt{gpt-6-sol} $0.351$; \texttt{gpt-6.1-sol} $0.333$; \texttt{claude-fable-5-1} $0.333$; \texttt{gemini-3.1-pro-preview} $0.316$; \texttt{gemini-3.8-flash} $0.298$ (\ms{} minus this model $0.430$ $\ci{0.316}{0.544}$); \base{} $0.263$. Over the nine comparison models, ESG was $0.298$ to $0.439$ and erased-lesion claims $37.7\%$ to $53.5\%$ on finding-specific questions, and $0.238$ to $0.505$ and $44.8\%$ to $70.5\%$ on generic questions. (C) The ESG columns in (C) are those of the four models on finding-specific questions; the strata are small and the tests uncorrected. Breast ultrasound: \ms{} above all nine comparison models, each $p < 0.001$. Colonoscopy: $p = 0.012$ (\opus) and $p = 0.025$ (\gpt); difference from \gem{} $0.125$ $\ci{-0.083}{0.333}$; \texttt{claude-sonnet-5-5} had the same ESG as \ms{} ($0.917$). Brain MRI: difference from \gpt{} $p = 0.068$. Chest X-ray: $p = 0.030$ (\opus) and $p = 0.077$ (\gem). In colonoscopy the nine comparison models reported the finding on $0\%$ to $45.8\%$ of the 24 erased polyps asked about with a polyp-specific question (mean of the nine $26.9\%$) and on $35.6\%$ to $69.5\%$ of the 59 asked about with a generic question (mean $52.4\%$).}
\end{table}

\begin{table}[!htbp]
\caption{Prompting and inference conditions for the three reference models on the same 150 main-test cases, as ESG / erased-lesion claims (\%). Default: the agent protocol of Section~\ref{subsec:models}. Cautionary prompt: the default prompt with an added paragraph stating that a question may name a finding that is not in the image and that reporting a finding that is not visible is a serious error. Solved examples: the cautionary prompt preceded by four solved training cases (a lesion image, the same image erased, a control-edited image and a lesion-free image). Highest reasoning and highest image resolution: the provider's highest setting. Plain prompt: one free-text reply without tools. No intervals were computed per cell. \opus{} was not run at the highest image-resolution setting. Question wordings not used in training are in Table~\ref{tab:s_sensitivity}.}
\label{tab:s_prompts}
\centering
\small
\begin{adjustbox}{max width=\linewidth}
\begin{tabular}{lcccccc}
\toprule
Model & Default & Cautionary prompt & Solved examples & Highest reasoning & Highest image resolution & Plain prompt \\
\midrule
\gem & $0.441$ / $41.2$ & $0.412$ / $42.6$ & $0.456$ / $36.8$ & $0.456$ / $39.7$ & $0.485$ / $38.2$ & $0.338$ / $52.9$ \\
\opus & $0.338$ / $54.4$ & $0.382$ / $44.1$ & $0.309$ / $52.9$ & $0.397$ / $51.5$ & -- & $0.412$ / $38.2$ \\
\gpt & $0.338$ / $52.9$ & $0.397$ / $47.1$ & $0.500$ / $38.2$ & $0.324$ / $58.8$ & $0.397$ / $51.5$ & $0.250$ / $66.2$ \\
\bottomrule
\end{tabular}
\end{adjustbox}
\\[2pt]\parbox{\linewidth}{\footnotesize On these cases \ms{} under the default protocol had ESG $0.750$ and erased-lesion claims of $7.4\%$. It had a higher ESG and net utility than each reference model under each condition for which a test was recorded ($p \le 0.012$). With the highest reasoning setting no reference model was shown to differ from its own default.}
\end{table}

\begin{table}[!htbp]
\caption{Open-ended request (pre-specified): the 36 main-test cases with a finding-specific question among the 150 cases of Table~\ref{tab:s_prompts}. Each model was asked ``Describe any abnormal findings in this image.'' (one reply, no tools) on the original and on the erased image, and a text-only judge (\gpt, not shown the image or the name of the model) marked whether the reply reports the finding. Free-text ESG: share of original images minus share of erased images on which the finding was reported.}
\label{tab:s_open}
\centering
\small
\begin{adjustbox}{max width=\linewidth}
\begin{tabular}{lcccc}
\toprule
Model & Original images: finding reported (\%) & Erased images: finding reported (\%) & Free-text ESG & \ms{} minus model \\
\midrule
\ms & $47.2$ & $0.0$ (0 of 36) & $0.472$ $\ci{0.306}{0.639}$ & \\
\gem & $69.4$ & $38.9$ & $0.306$ & $0.167$ $\ci{-0.028}{0.361}$, $p = 0.135$ \\
\opus & $86.1$ & $52.8$ & $0.333$ $\ci{0.139}{0.528}$ & $0.139$ $\ci{-0.083}{0.333}$, $p = 0.232$ \\
\gpt & $83.3$ & $47.2$ & $0.361$ & $0.111$ $\ci{-0.056}{0.278}$, $p = 0.250$ \\
\base & $80.6$ & $52.8$ & $0.278$ & \\
\bottomrule
\end{tabular}
\end{adjustbox}
\\[2pt]\parbox{\linewidth}{\footnotesize The pre-specified criterion, \ms{} above each reference model with the interval above zero, was not met.}
\end{table}

\begin{table}[!htbp]
\caption{Further analyses on the main test (507 cases). Each row changes one element of the scoring, the test or the resampling and gives the result for \ms{} and, where available, the comparison. The analysis without degraded views and the highest reasoning setting of Table~\ref{tab:s_prompts} were fixed before any test result of \ms{} existed; the reproduction was pre-specified; the remaining rows are descriptive.}
\label{tab:s_sensitivity}
\centering
\small
\begin{adjustbox}{max width=\linewidth}
\begin{tabular}{L{6.4cm}L{9.6cm}}
\toprule
Analysis & Result \\
\midrule
Only explicit ``absent'' answers counted towards ESG (referral not counted) & \ms{} $0.758$; nine comparison models $0.237$ to $0.470$ (\gem{} $0.470$, \texttt{claude-sonnet-5-5} $0.397$, \texttt{gemini-3.1-pro-preview} $0.384$, \texttt{claude-fable-5-1} $0.342$, \opus{} $0.324$, \gpt{} $0.311$, \texttt{gpt-6.1-sol} $0.297$, \texttt{gemini-3.8-flash} $0.269$, \texttt{gpt-6-sol} $0.237$) \\
\addlinespace
Net utility without the 170 degraded views & \ms{} $0.774$; \gem{} $0.579$, \opus{} $0.361$, \gpt{} $0.310$; differences $0.195$, $0.413$ and $0.464$, each $p < 0.001$ \\
\addlinespace
Degraded views (170) & \ms{} asked for a better image on 170 of 170 \\
\addlinespace
Finding reported with a box on the lesion (intersection over union at least $0.3$) & \ms{} $0.785$ of lesion images \\
\addlinespace
Open-weight models & \ms{} had a higher ESG and a higher net utility than each of the six (all $p < 0.001$). Its erased-lesion claims ($9.6\%$) were not the lowest: Qwen2.5-VL-7B reported $5.5\%$ and was correct on $0.144$ of lesion images, and Qwen3-VL-8B reported $12.8\%$ (Table~\ref{tab:main}) \\
\addlinespace
Question wordings not used in training (373 cases) & Net utility of \ms{} changed by $0.006$ $\ci{-0.033}{0.041}$, $p = 0.738$, against the original wording \\
\addlinespace
All tool calls refused at test time & ESG of \ms{} $0.753$ against $0.758$ ($p = 0.918$); net utility lower by $0.092$ ($p = 0.003$); lesion images correct lower by $0.053$ ($p = 0.001$) \\
\addlinespace
26 of the 288 lesion-free images removed whose label at least seven of the nine comparison models disputed & The conclusions for \ms{} were unchanged; its estimates without the 26 images were not recorded \\
\addlinespace
Whole patients or groups of near-identical images resampled, not cases & Intervals widened by at most $0.01$ (main test) and $0.03$ (fresh external set); all 27 primary comparisons kept $p < 0.001$; one secondary comparison changed side of $0.05$ (fresh external ESG against \opus: $p = 0.037$ to $p = 0.060$) \\
\addlinespace
Reproduction from the public release (pre-specified: ESG within $0.02$ and at least $97\%$ of decisions identical) & ESG $0.763$ against $0.758$; 1{,}102 of 1{,}115 decisions identical ($98.8\%$): met \\
\addlinespace
Ranking of errors by stated confidence & Selective risk (error rate among the answers kept) of \ms{} $0.151$ at $100\%$ coverage and $0.107$ at $70\%$ coverage (area under the risk-coverage curve $0.065$); reference models $0.114$ to $0.137$ and $0.023$ to $0.055$ (area $0.020$ to $0.033$) \\
\bottomrule
\end{tabular}
\end{adjustbox}
\end{table}

\begin{table}[!htbp]
\caption{Main test by quartile of lesion area (219 lesion cases): ESG of \ms{} and the three reference models, and erased-lesion claims of \ms. Quartile 1 holds the smallest lesions. Descriptive; no intervals were computed.}
\label{tab:s_size}
\centering
\small
\begin{adjustbox}{max width=\linewidth}
\begin{tabular}{lcccc}
\toprule
 & Quartile 1 & Quartile 2 & Quartile 3 & Quartile 4 \\
\midrule
\ms, ESG & $0.545$ & $0.741$ & $0.836$ & $0.909$ \\
\ms, erased-lesion claims (\%) & $18.2$ & $13.0$ & $5.5$ & $1.8$ \\
\gem, ESG & $0.400$ & $0.556$ & $0.418$ & $0.509$ \\
\opus, ESG & $0.200$ & $0.352$ & $0.345$ & $0.418$ \\
\gpt, ESG & $0.273$ & $0.333$ & $0.400$ & $0.218$ \\
\bottomrule
\end{tabular}
\end{adjustbox}
\end{table}

\clearpage
\subsection{Reader assessment}

\begin{table}[!htbp]
\caption{Specialist round (pre-specified): 212 main-test images, each from a different case. Results of the two readers who have read all images: a board-certified gastroenterologist (20 years of practice; median reading time $6.7$~s per image) and a board-certified radiologist (8 years of practice). (A) Gastroenterologist, colonoscopy images (61): number of images on which the finding was called present, with exact 95\% CI for erased images. (B) Gastroenterologist, other imaging types (outside the specialty; descriptive), and edit detection. (C) Radiologist, radiology images (brain MRI, breast ultrasound, chest X-ray and SLAKE; 151), and all images.}
\label{tab:s_expert}
\centering
\small
\begin{adjustbox}{max width=\linewidth}
\begin{tabular}{L{7.4cm}L{8.4cm}}
\toprule
\multicolumn{2}{l}{\textit{(A) Colonoscopy images: finding called present}} \\
\midrule
Unedited lesion images & 8 of 9 \\
Control-edited images (lesion intact), LaMa / Stable Diffusion & 8 of 8 / 19 of 20 (together 27 of 28) \\
Erased images, LaMa & 3 of 7 ($42.9\%$; $9.9$ to $81.6$) \\
Erased images, Stable Diffusion & 1 of 10 ($10.0\%$; $0.3$ to $44.5$) \\
Erased images, both tools & 4 of 17 ($23.5\%$; $6.8$ to $49.9$) \\
Erased images by question: generic (``Is there an abnormality?'') / about a polyp & 3 of 8 / 1 of 9 \\
Unedited lesion-free images & 0 of 7 \\
Erased images called edited, LaMa / Stable Diffusion & 0 of 7 / 5 of 10 \\
Edited against unedited images, AUC; erased against control-edited, AUC & $0.68$; $0.68$ \\
\midrule
\multicolumn{2}{l}{\textit{(B) Other imaging types (outside the reader's specialty) and all images}} \\
\midrule
Erased images of the other imaging types called present & 20 of 55 ($36.4\%$); ``cannot tell'' on 22 \\
Edited against unedited images, AUC, by imaging type & brain MRI $0.56$; breast ultrasound $0.54$; chest X-ray $0.48$; SLAKE CT and MRI $0.51$ \\
Edited against unedited images, AUC, all images; erased against control-edited & $0.54$; $0.59$ \\
Agreement on the finding with the two trainee readers (39 images read in both rounds) & 21 of 39 (Cohen's $\kappa$ $0.24$) and 24 of 39 ($\kappa$ $0.39$) \\
\midrule
\multicolumn{2}{l}{\textit{(C) Radiologist: radiology images, finding called present}} \\
\midrule
Unedited lesion images & 25 of 27 ($92.6\%$) \\
Control-edited images (lesion intact), LaMa / Stable Diffusion & 25 of 28 / 10 of 12 \\
Erased images, LaMa & 13 of 29 ($44.8\%$; $26.4$ to $64.3$) \\
Erased images, Stable Diffusion & 16 of 26 ($61.5\%$; $40.6$ to $79.8$) \\
Erased images by source, both tools: brain MRI / breast ultrasound / chest X-ray / SLAKE & 8 of 13 / 9 of 18 / 10 of 17 / 2 of 7 \\
Unedited lesion-free images & 5 of 29 ($17.2\%$) \\
Edited against unedited images, AUC; erased against control-edited, AUC & $0.64$; $0.53$ \\
All 212 images: erased images called present, LaMa / Stable Diffusion & 17 of 36 / 19 of 36; colonoscopy (outside the specialty) 7 of 17 \\
All 212 images: edited against unedited, AUC; erased against control-edited & $0.67$; $0.50$ \\
\bottomrule
\end{tabular}
\end{adjustbox}
\\[2pt]\parbox{\linewidth}{\footnotesize The pre-specified criterion (upper bound of the share of erased images called present at most $10\%$ for each tool) was not met: with each imaging type read by its specialist, 16 of 36 LaMa and 17 of 36 Stable Diffusion erasures were called present. The model answers on the erasures that the specialists accepted are in Supplementary Table~\ref{tab:s_specialist}. The round remains open for further radiologists.}
\end{table}

\begin{table}[!htbp]
\caption{Earlier reader round: two trainee physicians (internal medicine and radiology) each read the same 128 main-test images in random order, 32 each of original, erased (LaMa), control-edited and unedited lesion-free images, without knowing which were edited. (A) Per reader. (B) Erased and lesion-free images called present by imaging type (6 or 7 images of each kind per type).}
\label{tab:s_trainee}
\centering
\small
\begin{adjustbox}{max width=\linewidth}
\begin{tabular}{lccccccccc}
\toprule
 & \multicolumn{5}{c}{Image called edited} & \multicolumn{4}{c}{Finding called present} \\
\cmidrule(lr){2-6}\cmidrule(lr){7-10}
(A) Reader & Erased & \Chd{1.3cm}{Control-edited} & Unedited & \Chd{1.9cm}{AUC, edited against unedited} & \Chd{2.2cm}{AUC, erased against control-edited} & Original & Erased & \Chd{1.3cm}{Control-edited} & \Chd{1.2cm}{Lesion-free} \\
\midrule
Internal medicine trainee & 12 of 32 & 3 of 32 & 11 of 64 & $0.54$ & $0.64$ & 29 of 32 & 19 of 32 & 31 of 32 & 16 of 32 \\
Radiology trainee & 19 of 32 & 12 of 32 & 28 of 64 & $0.55$ & $0.61$ & 26 of 32 (3) & 16 of 32 (8) & 27 of 32 & 7 of 32 (9) \\
\bottomrule
\end{tabular}
\end{adjustbox}

\vspace{4pt}
\begin{adjustbox}{max width=\linewidth}
\begin{tabular}{lcccc}
\toprule
 & \multicolumn{2}{c}{Erased images called present} & \multicolumn{2}{c}{Lesion-free images called present} \\
\cmidrule(lr){2-3}\cmidrule(lr){4-5}
(B) Imaging type & Internal medicine trainee & Radiology trainee & Internal medicine trainee & Radiology trainee \\
\midrule
Colonoscopy & 1 of 6 & 2 of 6 & 2 of 6 & 0 of 6 \\
Breast ultrasound & 6 of 7 & 4 of 7 & 5 of 7 & 1 of 7 \\
Chest X-ray & 5 of 6 & 4 of 6 & 4 of 6 & 3 of 6 \\
Brain MRI & 3 of 7 & 5 of 7 & 1 of 7 & 1 of 7 \\
SLAKE CT and MRI & 4 of 6 & 1 of 6 & 4 of 6 & 2 of 6 \\
\midrule
All & 19 of 32 & 16 of 32 & 16 of 32 & 7 of 32 \\
\bottomrule
\end{tabular}
\end{adjustbox}
\\[2pt]\parbox{\linewidth}{\footnotesize Numbers in parentheses in (A) are ``cannot tell'' answers. Both readers together: 35 of 64 reads of erased images ($54.7\%$) and 23 of 64 reads of lesion-free images ($35.9\%$) called present; both readers called 13 of 32 erased images present ($40.6\%$; exact 95\% CI $23.7$ to $59.4$), so the pre-specified criterion (upper bound at most $10\%$) was not met. Agreement between the readers: Cohen's $\kappa$ $0.10$ on whether an image was edited (128 images) and $0.57$ on whether the finding was present (105 images on which neither reader answered ``cannot tell''). Of the 13 erased images that both readers called present, 4 carried the generic question ``Is there an abnormality?'' and 9 a finding-specific question, and 11 were chest X-rays, breast ultrasound or brain MRI images. The readers gave no reasons for their answers.}
\end{table}

\begin{table}[!htbp]
\caption{Model answers re-scored on erased images grouped by the trainee readers' calls (pre-specified; 32 erased images of the earlier reader round; answers already given on the main test). Entries are the share of images, or the number of images, on which the finding was reported.}
\label{tab:s_rescoring}
\centering
\small
\begin{adjustbox}{max width=\linewidth}
\begin{tabular}{L{5.6cm}ccc}
\toprule
 & \Chd{3.6cm}{Neither reader called the finding present (10 images)} & \Chd{3.4cm}{At most one reader did (19 images, including the 10)} & \Chd{2.6cm}{Both readers did (13 images)} \\
\midrule
Nine comparison models, pooled & $26\%$ (95\% CI $10\%$ to $42\%$) & $36\%$ & $82\%$ \\
Open-weight models and trained variants, pooled (43 evaluation runs) & $12\%$ & $19\%$ & $40\%$ \\
\ms & 0 of 10 & not computed & 4 of 13 \\
\bottomrule
\end{tabular}
\end{adjustbox}
\\[2pt]\parbox{\linewidth}{\footnotesize The pre-specified reading required the lower bound of the comparison models' rate on the images neither reader called present to be above $20\%$; it was not met (10 images). Restricted to comparison models that had reported the lesion on the original image of the same case, the rates were $30\%$ (19 of 63 model-image pairs) on images neither reader called present, $54\%$ on images the readers disagreed on and $85\%$ on images both called present. \ms{} found 56 of 63 lesions that the readers confirmed and reported a finding on 0 of 16 images that the readers confirmed as lesion-free. The 10 images neither reader called present include 4 colonoscopy images. Over all erased images of the main test, the nine comparison models pooled reported the erased finding on $45\%$ of 83 colonoscopy images (and found $96\%$ of the unedited polyps), $46\%$ of brain MRI, $71\%$ of breast ultrasound, $60\%$ of chest X-ray and $49\%$ of SLAKE images.}
\end{table}

\begin{table}[!htbp]
\caption{Model answers re-scored on the specialist round (144 images whose views are in the main test; Stable Diffusion views are not). Each imaging type was judged by its specialist: the radiologist for brain MRI, breast ultrasound, chest X-ray and SLAKE, the gastroenterologist for colonoscopy. Accepted erasure: the specialist did not call the finding present; strictly accepted: the specialist answered ``no''. Entries are the number of images on which the model reported the finding, except for confirmed lesions, where they are lesions found.}
\label{tab:s_specialist}
\centering
\small
\begin{adjustbox}{max width=\linewidth}
\begin{tabular}{lccccc}
\toprule
Model & \Chd{1.8cm}{Accepted erasures (20)} & \Chd{1.8cm}{Strictly accepted (15)} & \Chd{2.2cm}{Erasures called present (16)} & \Chd{1.8cm}{Confirmed lesions (66)} & \Chd{2.0cm}{Confirmed lesion-free (31)} \\
\midrule
\texttt{gpt-6-sol} & 5 & 2 & 10 & 59 & 7 \\
\gpt & 11 & 6 & 10 & 62 & 8 \\
\texttt{gpt-6.1-sol} & 10 & 5 & 10 & 64 & 6 \\
\texttt{claude-fable-5-1} & 10 & 7 & 11 & 59 & 9 \\
\opus & 12 & 9 & 11 & 60 & 4 \\
\texttt{claude-sonnet-5-5} & 7 & 5 & 11 & 64 & 0 \\
\gem & 5 & 3 & 11 & 63 & 0 \\
\texttt{gemini-3.1-pro-preview} & 10 & 8 & 10 & 58 & 6 \\
\texttt{gemini-3.8-flash} & 8 & 6 & 9 & 62 & 4 \\
\midrule
Nine comparison models, pooled & $43.3\%$ ($31.7$ to $56.1$) & $37.8\%$ ($23.7$ to $53.3$) & $64.6\%$ ($44.4$ to $83.4$) & & $15.8\%$ ($9.3$ to $22.2$) \\
\ms & 1 & 1 & 3 & 59 & 1 \\
\msn & 1 & 1 & 3 & 61 & 3 \\
\bottomrule
\end{tabular}
\end{adjustbox}
\\[2pt]\parbox{\linewidth}{\footnotesize Pooled: mean over the nine models of the per-image share, with a 95\% interval from 2,000 resamples of images; because every model read the same images, the interval reflects the images, not the models. The rule, written before this analysis was run and after the radiologist had finished, took the persistence to hold if the pooled lower bound on accepted erasures was above $20\%$, the bound used in Supplementary Table~\ref{tab:s_rescoring}; it was met ($31.7\%$). The nine models pooled minus \ms{} on the 20 accepted erasures: $38.3$ percentage points (95\% CI $22.8$ to $52.8$); the same for \msn. Accepted erasures by source: brain MRI 2, breast ultrasound 5, chest X-ray 4, colonoscopy 4, SLAKE 5. Interim: one specialist per imaging type.}
\end{table}

\clearpage
\subsection{Tests of the edit-detection explanation}

\begin{table}[!htbp]
\caption{Editing tools. (A) ESG / erased-lesion claims (\%) on the 125 main-test cases whose erased and control-edited views were remade on the same masks with four tools: LaMa and tissue transplant (used to build training views) and TELEA and Stable Diffusion (not used in training). First-stage models (first seed) differ in the views they were trained with; a lesion insertion is a training-only view in which a segmented lesion is copied into a lesion-free training image. (B) First-stage model trained with all views, on all 219 lesion cases (Stable Diffusion views exist for 191 of them): where ESG is lost with each tool. No intervals were computed per cell; differences between \ms{} and the reference models in (A) were $0.10$ to $0.48$ in ESG, each $p \le 0.049$ without correction for the twelve comparisons, which are exploratory.}
\label{tab:s_tools}
\centering
\small
\begin{adjustbox}{max width=\linewidth}
\begin{tabular}{lcccc}
\toprule
 & \multicolumn{2}{c}{Tools used in training} & \multicolumn{2}{c}{Tools not used in training} \\
\cmidrule(lr){2-3}\cmidrule(lr){4-5}
(A) Model (125 cases) & LaMa & Tissue transplant & TELEA & Stable Diffusion \\
\midrule
\ms & $0.752$ / $7.2$ & $0.720$ / $8.0$ & $0.664$ / $7.2$ & $0.640$ / $17.6$ \\
\gem & $0.536$ / $32.8$ & $0.488$ / $32.8$ & $0.560$ / $8.8$ & $0.392$ / $40.0$ \\
\opus & $0.376$ / $45.6$ & $0.264$ / $56.8$ & $0.296$ / $41.6$ & $0.352$ / $46.4$ \\
\gpt & $0.312$ / $55.2$ & $0.240$ / $64.0$ & $0.256$ / $60.8$ & $0.256$ / $61.6$ \\
\midrule
\multicolumn{5}{l}{\textit{First-stage models, by training views}} \\
Erased views, control edits and lesion insertions (as for \ms) & $0.720$ / $12.0$ & $0.744$ / $8.0$ & $0.696$ / $7.2$ & $0.568$ / $24.8$ \\
Erased views and control edits & $0.712$ / $17.6$ & $0.736$ / $9.6$ & $0.768$ / $7.2$ & $0.704$ / $16.0$ \\
Erased views and lesion insertions & $0.696$ / $6.4$ & $0.664$ / $4.8$ & $0.512$ / $5.6$ & $0.600$ / $11.2$ \\
Erased views only & $0.688$ / $7.2$ & $0.656$ / $4.8$ & $0.544$ / $2.4$ & $0.712$ / $4.8$ \\
No counterfactual views & $0.272$ / $69.6$ & $0.200$ / $76.8$ & $0.144$ / $83.2$ & $0.232$ / $73.6$ \\
\bottomrule
\end{tabular}
\end{adjustbox}

\vspace{4pt}
\begin{adjustbox}{max width=\linewidth}
\begin{tabular}{lccccc}
\toprule
(B) Tool (219 cases; 191 for Stable Diffusion) & \Chd{1.7cm}{Erased image denied} & \Chd{2.1cm}{Control-edited image denied} & ESG & \Chd{4.4cm}{Control-edited image denied, among lesions found on the original} & \Chd{3.2cm}{Erased-lesion claims with the box at the erased site} \\
\midrule
LaMa (used in training) & $0.881$ & $0.146$ & $0.735$ & $0.027$ (5 of 186) & 9 of 22 \\
TELEA (not used) & $0.918$ & $0.219$ & $0.699$ & $0.129$ (24 of 186; 13 breast ultrasound) & 11 of 13 \\
Stable Diffusion (not used) & $0.743$ & $0.152$ & $0.592$ & $0.043$ (7 of 162) & 41 of 45 \\
\bottomrule
\end{tabular}
\end{adjustbox}
\\[2pt]\parbox{\linewidth}{\footnotesize \gem's erased-lesion claims fell from $32.8\%$ with LaMa to $8.8\%$ with TELEA. With LaMa, 33 of 219 lesions were missed on the original image. Erased-lesion claims of \ms{} with Stable Diffusion were $10.4$ percentage points above those with LaMa ($p = 0.006$); Stable Diffusion erasures have been assessed by one specialist reader, within specialty for colonoscopy only (Table~\ref{tab:s_expert}), so claims at the erased site may reflect residual or regenerated lesion-like tissue. A development version of the model, trained with LaMa edits only, had been evaluated with the same four tools before \ms{} was specified; its ESG with TELEA was $0.536$ against $0.560$ for \gem{} (difference $-0.024$; no difference was shown, and no interval was recorded). Tissue transplant was then added to training, and TELEA and Stable Diffusion were fixed as the tools not used in training, before \ms{} was trained.}
\end{table}

\begin{table}[!htbp]
\caption{Training views: first-stage models on the main test (507 cases). (A) Models trained with and without counterfactual views; mean (SD) over three training seeds where marked, one seed otherwise. ``Absent'' share matched: no counterfactual views, with unedited lesion-free images added until the share of ``absent'' answers equals that of training with counterfactual views. Matched and unmatched control edits: the control edit has, or does not have, the shape and tool of the erasure. Preference optimisation: direct preference optimisation with erasure-based preferences after training without counterfactual views (one of the alternative designs of Section~\ref{subsec:res_designs}). (B) Two-by-two design of the two training-only kinds of view (first seed), with the share of 125 intact lesions denied after a control edit made with TELEA, a tool not used in training.}
\label{tab:s_ablations}
\centering
\small
\begin{adjustbox}{max width=\linewidth}
\begin{tabular}{lccccc}
\toprule
(A) Training views & Lesion images correct & Erased-lesion claims (\%) & Lesion-free claims (\%) & ESG & Net utility \\
\midrule
\multicolumn{6}{l}{\textit{Base model \base}} \\
With counterfactual views (3 seeds) & $0.871$ ($0.022$) & $13.2$ ($3.2$) & $13.9$ ($4.0$) & $0.741$ ($0.019$) & $0.684$ ($0.021$) \\
Without counterfactual views (3 seeds) & $0.955$ ($0.008$) & $67.6$ ($11.9$) & $35.9$ ($3.9$) & $0.279$ ($0.107$) & $0.341$ ($0.090$) \\
Without, ``absent'' share matched & $0.945$ & $73.5$ & $31.6$ & $0.215$ & $0.321$ \\
Erased views only & $0.788$ & $6.8$ & $5.9$ & $0.708$ & $0.667$ \\
Erased views and matched control edits & $0.897$ & $15.1$ & $11.8$ & $0.744$ & $0.727$ \\
Erased views and unmatched control edits & $0.904$ & $15.1$ & $12.5$ & $0.753$ & $0.739$ \\
Without, then preference optimisation (3 seeds) & $0.903$ ($0.014$) & $24.2$ ($5.3$) & $19.4$ ($4.3$) & $0.661$ ($0.037$) & $0.637$ ($0.033$) \\
\midrule
\multicolumn{6}{l}{\textit{Base model MedGemma 1.5 4B}} \\
With counterfactual views & $0.806$ & $7.8$ & $5.9$ & $0.721$ & $0.698$ \\
Without counterfactual views & $0.947$ & $57.5$ & $31.6$ & $0.374$ & $0.409$ \\
Not fine-tuned & $0.674$ & $54.3$ & $24.0$ & $0.283$ & $-0.255$ \\
\bottomrule
\end{tabular}
\end{adjustbox}

\vspace{4pt}
\begin{adjustbox}{max width=\linewidth}
\begin{tabular}{cccccccc}
\toprule
\multicolumn{3}{l}{(B) Views in training} & \multicolumn{4}{c}{Main test (507 cases)} & TELEA control edit (125 cases) \\
\cmidrule(lr){1-3}\cmidrule(lr){4-7}\cmidrule(lr){8-8}
Erased & \Chd{1.4cm}{Control edits} & \Chd{1.4cm}{Lesion insertions} & \Chd{1.8cm}{Lesion images correct} & \Chd{1.9cm}{Erased-lesion claims (\%)} & \Chd{1.7cm}{Lesion-free claims (\%)} & ESG & \Chd{2.6cm}{Intact lesion denied (\%)} \\
\midrule
yes & no & no & $0.788$ & $6.8$ & $5.9$ & $0.708$ & $43.2$ \\
yes & no & yes & $0.806$ & $7.3$ & $11.1$ & $0.735$ & $43.2$ \\
yes & yes & no & $0.897$ & $15.1$ & $11.8$ & $0.744$ & $16.0$ \\
yes & yes & yes & $0.847$ & $11.9$ & $9.4$ & $0.735$ & $23.2$ \\
\bottomrule
\end{tabular}
\end{adjustbox}
\\[2pt]\parbox{\linewidth}{\footnotesize With against without counterfactual views, the ESG difference by seed was $0.525$, $0.539$ and $0.324$ (each $p < 0.001$); on the external datasets ESG was $0.622$ ($0.030$) against $0.389$ ($0.043$). With counterfactual views against the ``absent''-matched model: $0.521$ $\ci{0.440}{0.600}$. MedGemma 1.5 4B with against without counterfactual views: $0.347$ $\ci{0.276}{0.425}$. The second and third training stages changed main-test net utility by $0.040$ ($p = 0.15$) and ESG by $0.023$ ($p = 0.39$) over the first stage and lowered lesion-free claims on validation from $13.6\%$ to $7.7\%$. In (B), control edits raised lesion images correct by $0.109$ without insertions and $0.041$ with them ($p \le 0.018$).}
\end{table}

\begin{table}[!htbp]
\caption{Edit-site control and hidden-state probes on validation cases (179 lesion cases with a yes/no question and 80 lesion-free images), for the first-stage model (first seed) and the base model \base. (A) Share of intact or erased lesions denied after each view, among cases whose lesion the model found on the original image (143 for the first-stage model, 127 for the base model). Distant: the control edit away from the lesion. Ring: the edit-site control, a ring-shaped edit that starts 2 pixels outside the intact lesion. Restored: the lesion pixels pasted back into the erased image. Half erased: half of the lesion mask erased (LaMa only). ESG and AUC are computed with the distant or with the ring edit as the control. Pre-specified flag for a site effect: ring denial at least $0.10$, at least $0.05$ above the distant edit and at least $0.05$ above the base model. (B) Cross-validated AUC (five folds, grouped by case) of logistic-regression probes on hidden states of the language model, trained on LaMa views, at the best of eight layers.}
\label{tab:s_site}
\centering
\small
\begin{adjustbox}{max width=\linewidth}
\begin{tabular}{llccccccccccc}
\toprule
(A) Model & Tool & \Chd{1.2cm}{Found on original} & Distant & Ring & Restored & \Chd{1.0cm}{Half erased} & Erased & \Chd{2.6cm}{Ring minus distant} & \Chd{1.0cm}{ESG, distant} & \Chd{0.8cm}{ESG, ring} & \Chd{1.0cm}{AUC, distant} & \Chd{0.8cm}{AUC, ring} \\
\midrule
First-stage model & LaMa & $0.799$ & $0.042$ & $0.077$ & $0.070$ & $0.196$ & $0.867$ & $0.035$ $\ci{-0.021}{0.091}$ & $0.665$ & $0.648$ & $0.853$ & $0.845$ \\
First-stage model & TELEA & $0.799$ & $0.126$ & $0.203$ & $0.140$ & -- & $0.923$ & $0.077$ $\ci{0.000}{0.154}$ & $0.654$ & $0.642$ & $0.847$ & $0.824$ \\
\base & LaMa & $0.709$ & $0.023$ & $0.047$ & -- & -- & -- & $0.023$ & -- & -- & -- & -- \\
\base & TELEA & $0.709$ & $0.070$ & $0.116$ & -- & -- & -- & $0.047$ & -- & -- & -- & -- \\
\bottomrule
\end{tabular}
\end{adjustbox}

\vspace{4pt}
\begin{adjustbox}{max width=\linewidth}
\begin{tabular}{llcccc}
\toprule
 & & \multicolumn{2}{c}{First-stage model} & \multicolumn{2}{c}{\base} \\
\cmidrule(lr){3-4}\cmidrule(lr){5-6}
(B) Probe separates & Test tool & Last prompt token & Image tokens (mean) & Last prompt token & Image tokens (mean) \\
\midrule
Erased from ring-edited & LaMa & $0.898$ & $0.901$ & $0.829$ & $0.870$ \\
Erased from ring-edited & TELEA & $0.816$ & $0.852$ & $0.808$ & $0.831$ \\
Erased from distant control edit & LaMa & $0.898$ & $0.887$ & $0.827$ & $0.853$ \\
Erased from distant control edit & TELEA & $0.812$ & $0.851$ & $0.771$ & $0.804$ \\
Ring-edited from original & LaMa & $0.618$ & $0.629$ & $0.573$ & $0.617$ \\
Ring-edited from original & TELEA & $0.802$ & $0.761$ & $0.633$ & $0.707$ \\
\bottomrule
\end{tabular}
\end{adjustbox}
\\[2pt]\parbox{\linewidth}{\footnotesize The flag for a site effect was met with TELEA and not with LaMa. Post hoc, on the 112 cases found by both models: after a TELEA ring edit the first-stage model denied $12.5\%$ and the base model $6.2\%$ (difference $0.062$ $\ci{-0.009}{0.134}$; bootstrap over cases, 4{,}000 resamples); after the distant TELEA edit $8.9\%$ and $4.5\%$; the site-specific part, (ring minus distant) for the first-stage model minus the same for the base model, was $0.018$ $\ci{-0.080}{0.116}$. With LaMa the corresponding figures were $4.5\%$ and $7.1\%$ (difference $-0.027$ $\ci{-0.081}{0.027}$). The TELEA ring denials of the first-stage model were 12 of 17 chest X-rays, 0 of 33 breast ultrasound, 2 of 34 colonoscopy and 0 of 24 brain MRI cases; no reader has assessed whether a ring edit next to a small chest lesion alters the lesion itself. In (B), the answers of the first-stage model separated erased from ring-edited TELEA views with an AUC of $0.824$, against $0.852$ and $0.816$ for the probes: the difference ($0.028$ and $-0.008$) is below the pre-specified threshold of $0.05$ for a gap between what the hidden states hold and what the answers use.}
\end{table}

\begin{table}[!htbp]
\caption{Within-case change of a single-turn answer score (log-odds of the ``present'' reply against the ``absent'' reply, in nats) relative to the original image, on validation cases. Descriptive only: the planned use of this score as a classifier across cases did not pass its validity check ($63\%$ agreement with the model's generated answers against the required $80\%$), because the ``present'' reply carries box coordinates and is about 11 nats less probable than the ``absent'' reply on every image. Differences within a case cancel this constant. Paired-view consistency loss: one of the alternative designs of Section~\ref{subsec:res_designs}.}
\label{tab:s_logodds}
\centering
\small
\begin{adjustbox}{max width=\linewidth}
\begin{tabular}{llccccc}
\toprule
Model & Tool & Erased & Distant control edit & Ring edit & Restored & Half erased \\
\midrule
First-stage model, all views & LaMa & $-14.10$ & $-0.09$ & $-0.43$ & $+0.87$ & $-4.05$ \\
First-stage model, all views & TELEA & $-10.76$ & $-2.41$ & $-1.53$ & $+1.34$ & -- \\
\base & LaMa & $-3.58$ & $-0.07$ & $+0.17$ & $+0.69$ & $-0.41$ \\
First stage, no counterfactual views & TELEA & $-10.34$ & $-1.37$ & $-0.94$ & $+1.25$ & -- \\
First stage, erased views only & TELEA & $-11.88$ & $-4.77$ & $-3.94$ & $-0.62$ & -- \\
First stage continued with a paired-view consistency loss & TELEA & $-6.08$ & $-0.31$ & $-0.12$ & $+0.57$ & -- \\
\bottomrule
\end{tabular}
\end{adjustbox}
\end{table}

\begin{table}[!htbp]
\caption{Edit visibility on the main test (219 erased images). (A) Pre-specified analysis. A ResNet-18 trained to separate control-edited from unedited training views scored every test view (test AUC $0.826$ for control-edited against unedited, $0.841$ for erased against unedited, $0.522$ for erased against control-edited). An erasure is low-visibility when it scores no higher than the median control edit (median score $0.509$): 111 low-visibility and 108 other erasures. (B) Repeat of the analysis for \msn. The detector was fitted again for this run (test AUC $0.840$, $0.829$ and $0.497$; median score $0.532$), which gives 117 low-visibility and 102 other erasures and slightly different values for \ms. (C) An earlier, descriptive version: erased images that \gem, asked for the probability that an image was edited, rated at most $0.3$ (61) or higher (158).}
\label{tab:s_visibility}
\centering
\small
\begin{adjustbox}{max width=\linewidth}
\begin{tabular}{llcccc}
\toprule
\Chd{2.2cm}{Erasures: low-visibility / other} & Model & \Chd{2.6cm}{Erased-lesion claims, low-visibility (\%)} & \Chd{2.3cm}{Erased-lesion claims, other (\%)} & \Chd{2.6cm}{ESG, low-visibility} & \Chd{4.4cm}{ESG difference, low-visibility (see note)} \\
\midrule
(A) 111 / 108 & \ms & $16.2$ & $2.8$ & $0.640$ $\ci{0.550}{0.721}$ & \\
 & \gem & $44.1$ & $38.0$ & & $0.270$ $\ci{0.153}{0.387}$ \\
 & \opus & $59.5$ & $48.1$ & & $0.450$ $\ci{0.324}{0.568}$ \\
 & \gpt & $55.0$ & $63.0$ & & $0.378$ $\ci{0.243}{0.514}$ \\
\midrule
(B) 117 / 102 & \msn & $13.7$ & $3.9$ & $0.624$ $\ci{0.538}{0.709}$ & \\
 & \ms & $14.5$ & $3.9$ & $0.650$ $\ci{0.564}{0.726}$ & $-0.026$ $\ci{-0.103}{0.051}$, $p = 0.587$ \\
 & \gem & $42.7$ & $39.2$ & $0.393$ $\ci{0.291}{0.496}$ & $0.231$ $\ci{0.103}{0.359}$ \\
 & \opus & $57.3$ & $50.0$ & $0.222$ $\ci{0.137}{0.316}$ & $0.402$ $\ci{0.291}{0.513}$ \\
 & \gpt & $54.7$ & $63.7$ & $0.265$ $\ci{0.171}{0.359}$ & $0.359$ $\ci{0.231}{0.479}$ \\
\midrule
(C) 61 / 158 & \ms & $15$ & $8$ & & \\
 & Nine comparison models & $41$ to $64$ & & & \\
\bottomrule
\end{tabular}
\end{adjustbox}
\\[2pt]\parbox{\linewidth}{\footnotesize In (A) the differences are \ms{} minus the reference model; in (B) they are \msn{} minus the model in the row. All differences from reference models have $p < 0.001$. The ESG of the reference models on the low-visibility erasures of (A) was not tabulated. Further measurements of how detectable the edits are: an earlier ResNet-18 separated control-edited from original test images with an AUC of $0.832$ $\ci{0.799}{0.868}$; a linear probe on the frozen image encoder of \base{} did so with $0.603$ $\ci{0.584}{0.626}$ and separated erased from control-edited images with $0.860$; asked directly, \gem{} separated control-edited from original images with an AUC of $0.524$ $\ci{0.474}{0.575}$ and erased from original images with $0.753$, and \texttt{gpt-6.1-sol} with $0.546$ $\ci{0.491}{0.598}$ and $0.698$.}
\end{table}

\begin{table}[!htbp]
\caption{Test images that resemble training images (pre-specified). Within each source, main-test cases were ranked by the similarity (DINOv2 image features) of their image to the most similar training image of that source, and the most similar cases were removed. The comparisons were then recomputed from the answers already given. Differences are \ms{} minus the reference model in ESG.}
\label{tab:s_duplicates}
\centering
\small
\begin{adjustbox}{max width=\linewidth}
\begin{tabular}{L{4.4cm}cccccc}
\toprule
Cases removed & \Chd{1.0cm}{Cases kept} & \Chd{2.4cm}{ESG of \ms} & \Chd{2.3cm}{Erased-lesion claims of \ms{} (\%)} & Minus \gem & Minus \opus & Minus \gpt \\
\midrule
None & 507 & $0.758$ $\ci{0.701}{0.815}$ & $9.6$ & $0.288$ $\ci{0.208}{0.367}$ & --$^{\ast}$ & --$^{\ast}$ \\
Most similar $25\%$ of every source & 378 & $0.755$ & & $0.306$ $\ci{0.200}{0.405}$ & $0.435$ $\ci{0.333}{0.530}$ & $0.456$ $\ci{0.342}{0.561}$ \\
Most similar $50\%$ of every source & 252 & $0.795$ & & \multicolumn{3}{c}{each at least $0.375$, intervals above zero} \\
Most similar $50\%$ of the three sources without patient identifiers & 337 & $0.715$ & $13.1$ & $0.323$ & $0.454$ & $0.415$ \\
\bottomrule
\end{tabular}
\end{adjustbox}
\\[2pt]\parbox{\linewidth}{\footnotesize The pre-specified reading, that the advantage remains without the most similar cases, was met. $^{\ast}$\,Given only within the range over the nine comparison models ($0.288$ to $0.489$; Table~\ref{tab:s_seeds}). Intervals for the last row are above zero. Where patient identifiers exist, similarity did not separate images of the same patient from images of different patients (AUC $0.531$ on ChestX-Det, 35 images, and $0.530$ on TCGA-LGG, 486 images), so the exclusion tests look-alike images, not same-patient images. Median similarity to the nearest training image by source: TCGA-LGG $0.889$, BUSI $0.844$, ChestX-Det $0.987$, colonoscopy $0.863$, SLAKE $0.846$.}
\end{table}

\clearpage
\subsection{External datasets and new test sets}

\begin{table}[!htbp]
\caption{\ms{} and the three reference models on the external datasets (169 cases: CVC-ClinicDB 40, BraTS-Africa 2023 64, BUS-BRA 45, QaTa-COV19 20), the fresh internal set (76 ChestX-Det cases from patients not used in training, validation or the main test) and the fresh external set (316 cases: BraTS 2023 meningioma 82, BUS-UCLM 99, SIIM-ACR pneumothorax 96, ETIS-Larib 39). (A) All cases of each set. (B) \ms{} and \gem{} on the fresh external set by source; the ESG of \ms{} on the external datasets by source is in the note. Point estimates; the intervals and tests that were computed are given in Section~\ref{subsec:res_external} and below the table.}
\label{tab:s_external}
\centering
\small
\begin{adjustbox}{max width=\linewidth}
\begin{tabular}{llccccc}
\toprule
(A) Test set (cases) & Model & Lesion images correct & Erased-lesion claims (\%) & Lesion-free claims (\%) & ESG & Net utility \\
\midrule
External datasets (169) & \ms & $0.727$ & $11.8$ & $4.0$ & $0.597$ & $0.530$ \\
 & \gem & $0.950$ & $58.8$ & $18.0$ & $0.370$ & $0.544$ \\
 & \opus & $0.966$ & $53.8$ & $30.0$ & $0.412$ & $0.544$ \\
 & \gpt & $0.945$ & $66.4$ & $24.0$ & $0.277$ & $0.424$ \\
\addlinespace
Fresh internal (76) & \ms & $0.681$ & $19.4$ & $12.5$ & $0.472$ & $0.402$ \\
 & \gem & $0.639$ & $61.1$ & $12.5$ & $0.056$ & $0.048$ \\
 & \opus & $0.653$ & $63.9$ & $25.0$ & $0.000$ & $-0.023$ \\
 & \gpt & $0.847$ & $75.0$ & $17.5$ & $0.083$ & $0.219$ \\
\addlinespace
Fresh external (316) & \ms & $0.476$ & $8.3$ & $4.7$ & $0.393$ & $-0.053$ \\
 & \gem & $0.741$ & $31.5$ & $9.5$ & $0.435$ & $0.310$ \\
 & \opus & $0.696$ & $42.3$ & $18.9$ & $0.292$ & $0.056$ \\
 & \gpt & $0.634$ & $35.7$ & $34.5$ & $0.268$ & $-0.106$ \\
\bottomrule
\end{tabular}
\end{adjustbox}

\vspace{4pt}
\begin{adjustbox}{max width=\linewidth}
\begin{tabular}{lcccccccc}
\toprule
 & & \multicolumn{5}{c}{\ms} & \multicolumn{2}{c}{\gem} \\
\cmidrule(lr){3-7}\cmidrule(lr){8-9}
(B) Fresh external set, by source & Cases & \Chd{1.6cm}{Lesion images correct} & \Chd{1.7cm}{Erased-lesion claims (\%)} & \Chd{1.5cm}{Lesion-free claims (\%)} & ESG & \Chd{1.1cm}{Net utility} & \Chd{1.6cm}{Lesion images correct} & \Chd{1.1cm}{Net utility} \\
\midrule
BUS-UCLM (breast ultrasound) & 99 & $0.73$ & $2$ & $6$ & $0.71$ & $0.67$ & $0.71$ & $0.55$ \\
ETIS-Larib (colonoscopy) & 39 & $0.69$ & $21$ & none$^{\ast}$ & $0.46$ & $0.38$ & $0.97$ & $0.75$ \\
BraTS 2023 meningioma (brain MRI) & 82 & $0.53$ & $16$ & $8$ & $0.41$ & $-0.03$ & $0.62$ & $0.09$ \\
SIIM-ACR pneumothorax (chest X-ray) & 96 & $0.00$ & $0$ & $0$ & $0.00$ & $-1.10$ & $0.66$ & $-0.06$ \\
\bottomrule
\end{tabular}
\end{adjustbox}
\\[2pt]\parbox{\linewidth}{\footnotesize External datasets: ESG of \ms{} above each reference model ($p \le 0.002$); net utility not shown to differ from that of any of them. Fresh internal: ESG above each reference model ($p \le 0.001$). Fresh external: net utility of \ms{} minus \gem{} $-0.363$ $\ci{-0.496}{-0.233}$, $p < 0.001$; ESG above \opus{} with $p = 0.037$ when cases are resampled and $p = 0.060$ when patients are resampled. In (B), claims are percentages; $^{\ast}$\,this source has no lesion-free images. External datasets, ESG of \ms{} by source: CVC-ClinicDB (colonoscopy, 40 cases) $0.85$; BraTS-Africa 2023 (brain MRI, 64) $0.71$; BUS-BRA (breast ultrasound, 45) $0.58$; QaTa-COV19 (chest X-ray, 20) $0.05$. The QaTa-COV19 cases show diffuse opacities, which may not be removable by erasing a masked region. One fresh external chest film is the same image as a lesion-free validation image of another dataset, which labels it differently; without it, the fresh external ESG of \ms{} is $0.395$, lesion images correct $0.479$ and net utility $-0.046$ (315 cases).}
\end{table}

\begin{table}[!htbp]
\caption{Pneumothorax: post hoc analysis of the miss by \ms{} and the remedy. (A) Share of 42 pneumothorax films of the fresh external set called present, on the 256-pixel copies that the set holds and at original resolution (42 of the 48 films could be recovered at original resolution). (B) Films of 100 patients with and 100 without pneumothorax from the same source (SIIM-ACR), in no other set. Models in (B) differ only in where pneumothorax films of 150 patients with and 150 without pneumothorax entered training.}
\label{tab:s_pneumothorax}
\centering
\small
\begin{adjustbox}{max width=\linewidth}
\begin{tabular}{L{7.2cm}cc}
\toprule
(A) Model (42 films) & 256-pixel copy called present (\%) & Original resolution called present (\%) \\
\midrule
\base{} (not fine-tuned) & $0$ & $12$ \\
First-stage model, no counterfactual views & $0$ & $5$ \\
First-stage model, with counterfactual views & $19$ & $40$ \\
Model after the second training stage & not run & $40$ \\
\ms{} (after the third stage) & $0$ & $0$ \\
\gem & $69$ & $69$ \\
\bottomrule
\end{tabular}
\end{adjustbox}

\vspace{4pt}
\begin{adjustbox}{max width=\linewidth}
\begin{tabular}{L{7.2cm}ccc}
\toprule
(B) Model (100 + 100 films of new patients) & Pneumothorax films called present & Films without pneumothorax called present & ESG \\
\midrule
\ms & 0 of 100 & 0 of 100 & $0.000$ \\
Third stage repeated with the added films (validation rule not met; not used further) & 26 of 100 & 1 of 100 & $0.080$ \\
\msp{} (added films in every stage; validation rule met) & 53 of 100 & 5 of 100 & $0.360$ \\
\bottomrule
\end{tabular}
\end{adjustbox}
\\[2pt]\parbox{\linewidth}{\footnotesize Training coverage: questions naming pneumothorax made up 54 of about 7{,}000 chest X-ray training views in the first stage (35 in the second, 15 in the third), with the answer ``absent'' in $80\%$, $100\%$ and $87\%$ of them; the first stage held 4 pneumothorax images, all large (median lesion box $37\%$ of the image against $6.9\%$ for the test films). The validation set, the main test and the fresh internal set hold no pneumothorax image asked about pneumothorax. On the 48 pneumothorax films of the fresh external set, \ms{} found 0, \opus{} 12, \gpt{} 14 and \gem{} 30. Asked ``Is there an abnormality?'', \ms{} answered ``yes'' on 35 of the 42 pneumothorax films and on 28 of 48 films without pneumothorax from the same source (Fisher's exact $p = 0.012$); this does not show that it identifies the pneumothorax. The model with the third stage repeated did not meet its validation rule (lesion-free claims on validation $10.0\%$ against $7.7\%$; limit 2 percentage points above) and was tested as planned; in its first run it found $27\%$ of the 100 pneumothorax films, and 26 of 100 when run again together with \msp. \msp{} met its rule (validation net utility $0.632$ against $0.613$ for \ms, limit at least $0.593$; lesion-free claims $9.1\%$ against $7.7\%$, limit at most $9.7\%$; pneumothorax found on validation films $65\%$) and is one training seed. On the fresh external pneumothorax films, held as 256-pixel copies, it was correct on $0.22$ of lesion images (\ms: $0.00$).}
\end{table}

\section{Additional results: unedited pairs, \texorpdfstring{\msn}{MedScope-4B-N} and alternative designs}
\label{sec:s_results_b}

This section gives the complete tables behind Sections~\ref{subsec:res_pairs} to~\ref{subsec:res_designs}. Two models appear here that the main text does not name: each of the two kinds of unedited training images in \msn{} was first added to \msp{} on its own, giving ``\msp{} with colonoscopy pairs'' and ``\msp{} with CheXthought films'' (Supplementary Table~\ref{tab:s_models}). Unless a caption says otherwise, brackets are 95\% confidence intervals (CI) from $2{,}000$ bootstrap resamples, differences are paired over the same cases, videos or patients, $p$ values are two-sided, and a bootstrap $p$ value of $0.000$ is printed as $<0.001$. Claims are printed as percentages and their differences as differences in proportions. The three training seeds of \msn{} are called first, second and third; the first is the model of the main text. n.r.: not recorded; for a difference marked (n.r.), only the fact that its interval includes zero was recorded. --: not applicable or not computed.

\subsection{Unedited colonoscopy pairs}
\label{subsec:s_pairs}

Supplementary Tables~\ref{tab:s_realcolon} to~\ref{tab:s_framegap} give every model on the REAL-Colon pairs, on the context-matched frames, with one centre withheld from training, and on LDPolypVideo, also with pairs restricted by frame distance.

\paragraph{Where \ms{} missed polyps (descriptive)} On the 103 polyp-free REAL-Colon frames \ms{} reported a polyp on 2 (\base{} 15, \opus{} 20) and asked for a better image on 12, of which 11 came from one centre; this is why $86.4\%$ (89 of 103) of its polyp-free frames count as answered ``absent'' or referred. On the polyp frames it called the polyp present on 36 and asked for a better image on 3. By thirds of polyp box area (small, middle, large) it called 5 of 34, 7 of 35 and 24 of 34 present; \base{} 5, 8 and 27; \opus{} 19, 23 and 31. The median polyp box covered $2.9\%$ of the frame, and 35 of 103 covered less than $2\%$. \ms{} made no tool call on 61 polyp frames (5 called present, 53 ``absent'', 3 requests for a better image) and one zoom on 42 (31 called present, 11 ``absent'').

\begin{table}[!htbp]
\caption{REAL-Colon, 103 unedited frame pairs from 46 videos of four centres: all models. Found: the polyp frame was not answered ``absent'' and was not referred. Called present and correct box: the polyp frame was answered ``present'', and answered ``present'' with a box overlapping the annotated polyp (intersection over union at least 0.3). Absent or referred: the answer to the polyp-free frame. A request for a better image counts as neither. Cross-validated: each pair is answered by the copy of the model trained without that pair's video (three folds). Brackets: 95\% intervals from $2{,}000$ bootstrap resamples of whole videos. The comparison of \ms{} with the other models was the first pre-specified test on these pairs; the comparison of cross-validated \msn{} was pre-specified later, on the same pairs. Fold values are natural ESG within the pairs of each fold. --: not applicable.}
\label{tab:s_realcolon}
\centering
\footnotesize
\begin{adjustbox}{max width=\linewidth}
\begin{tabular}{l cc c cc cc}
\toprule
Model & \shortstack{Polyp frames\\found} & \shortstack{Polyp-free frames\\absent or referred} & \shortstack{Natural ESG\\{}[95\% CI]} & \shortstack{\msn{} (cross-validated)\\minus model [95\% CI]} & $p$ & \shortstack{\ms{}\\minus model [95\% CI]} & $p$ \\
\midrule
\msn{}, cross-validated & $64.1\%$ & $89.3\%$ & $0.534$ $\ci{0.394}{0.661}$ & -- & -- & -- & -- \\
\msp{} with colonoscopy pairs, cross-validated & $63.1\%$ & $92.2\%$ & $0.553$ $\ci{0.430}{0.670}$ & $-0.019$ $\ci{-0.086}{+0.048}$ & $0.604$ & -- & -- \\
\ms{} & $37.9\%$ & $86.4\%$ & $0.243$ $\ci{0.149}{0.340}$ & $+0.291$ $\ci{+0.176}{+0.402}$ & $<0.001$ & -- & -- \\
\msp{} & $39.8\%$ & $86.4\%$ & $0.262$ $\ci{0.168}{0.357}$ & $+0.272$ $\ci{+0.162}{+0.378}$ & $<0.001$ & $-0.019$ $\ci{-0.094}{+0.047}$ & $0.678$ \\
\base{} & $38.8\%$ & $85.4\%$ & $0.243$ $\ci{0.143}{0.350}$ & $+0.291$ $\ci{+0.167}{+0.407}$ & $<0.001$ & $+0.000$ $\ci{-0.103}{+0.095}$ & $1.000$ \\
\gem{} & $68.0\%$ & $78.6\%$ & $0.466$ $\ci{0.336}{0.581}$ & $+0.068$ $\ci{-0.069}{+0.202}$ & $0.347$ & $-0.223$ $\ci{-0.346}{-0.102}$ & $0.001$ \\
\opus{} & $70.9\%$ & $73.8\%$ & $0.447$ $\ci{0.317}{0.579}$ & $+0.087$ $\ci{-0.049}{+0.230}$ & $0.259$ & $-0.204$ $\ci{-0.333}{-0.071}$ & $0.001$ \\
\gpt{} & $64.1\%$ & $68.0\%$ & $0.320$ $\ci{0.212}{0.437}$ & $+0.214$ $\ci{+0.071}{+0.343}$ & $0.003$ & $-0.078$ $\ci{-0.184}{+0.028}$ & $0.172$ \\
\bottomrule
\end{tabular}
\end{adjustbox}

\vspace{0.6em}
\begin{adjustbox}{max width=\linewidth}
\begin{tabular}{l cc ccc}
\toprule
 & \multicolumn{2}{c}{Polyp frames, $n$ of $103$ (\%)} & \multicolumn{3}{c}{Natural ESG by fold} \\
\cmidrule(lr){2-3}\cmidrule(lr){4-6}
Model & Called present & With a correct box & Fold 0 ($n = 38$) & Fold 1 ($n = 35$) & Fold 2 ($n = 30$) \\
\midrule
\msn{}, cross-validated & $66$ ($64.1\%$) & $54$ ($52.4\%$) & $0.58$ & $0.51$ & $0.50$ \\
\msp{} with colonoscopy pairs, cross-validated & $65$ ($63.1\%$) & $53$ ($51.5\%$) & $0.63$ & $0.49$ & $0.53$ \\
\ms{} & $36$ ($35.0\%$) & $32$ ($31.1\%$) & $0.32$ & $0.23$ & $0.17$ \\
\msp{} & $37$ ($35.9\%$) & $34$ ($33.0\%$) & $0.34$ & $0.31$ & $0.10$ \\
\base{} & $40$ ($38.8\%$) & $13$ ($12.6\%$) & $0.26$ & $0.23$ & $0.23$ \\
\gem{} & $69$ ($67.0\%$) & $52$ ($50.5\%$) & $0.58$ & $0.43$ & $0.37$ \\
\opus{} & $73$ ($70.9\%$) & $50$ ($48.5\%$) & $0.55$ & $0.43$ & $0.33$ \\
\gpt{} & $66$ ($64.1\%$) & $41$ ($39.8\%$) & $0.29$ & $0.46$ & $0.20$ \\
\bottomrule
\end{tabular}
\end{adjustbox}
\end{table}

\begin{table}[!htbp]
\caption{REAL-Colon, context-matched frames (secondary; descriptive): 96 of the 103 pairs, with the polyp-free frame taken 3 to 10~s from the nearest frame that shows the polyp, in place of the frame 10 to 30~s away used in Supplementary Table~\ref{tab:s_realcolon}; 7 pairs had no usable frame. Columns and intervals as in Supplementary Table~\ref{tab:s_realcolon}. The lower panel gives natural ESG within the pairs of each of the four centres.}
\label{tab:s_context}
\centering
\footnotesize
\begin{adjustbox}{max width=\linewidth}
\begin{tabular}{l cc c cc}
\toprule
Model & \shortstack{Polyp frames\\found} & \shortstack{Context frames\\absent or referred} & \shortstack{Natural ESG\\{}[95\% CI]} & \shortstack{\msn{} (cross-validated)\\minus model [95\% CI]} & $p$ \\
\midrule
\msn{}, cross-validated & $62.5\%$ & $85.4\%$ & $0.479$ $\ci{0.322}{0.615}$ & -- & -- \\
\msp{} with colonoscopy pairs, cross-validated & $61.5\%$ & $88.5\%$ & $0.500$ $\ci{0.367}{0.629}$ & $-0.021$ $\ci{-0.088}{+0.049}$ & $0.604$ \\
\ms{} & $36.5\%$ & $84.4\%$ & $0.208$ $\ci{0.089}{0.333}$ & $+0.271$ $\ci{+0.148}{+0.383}$ & $<0.001$ \\
\msp{} & $37.5\%$ & $86.5\%$ & $0.240$ $\ci{0.132}{0.352}$ & $+0.240$ $\ci{+0.101}{+0.363}$ & $0.001$ \\
\base{} & $37.5\%$ & $81.2\%$ & $0.188$ $\ci{0.032}{0.346}$ & $+0.292$ $\ci{+0.118}{+0.468}$ & $<0.001$ \\
\gem{} & $66.7\%$ & $66.7\%$ & $0.333$ $\ci{0.211}{0.458}$ & $+0.146$ $\ci{+0.000}{+0.284}$ & $0.067$ \\
\opus{} & $69.8\%$ & $62.5\%$ & $0.323$ $\ci{0.177}{0.466}$ & $+0.156$ $\ci{+0.032}{+0.277}$ & $0.019$ \\
\gpt{} & $63.5\%$ & $55.2\%$ & $0.188$ $\ci{0.033}{0.353}$ & $+0.292$ $\ci{+0.137}{+0.429}$ & $<0.001$ \\
\bottomrule
\end{tabular}
\end{adjustbox}

\vspace{0.6em}
\begin{adjustbox}{max width=\linewidth}
\begin{tabular}{l cccc}
\toprule
 & \multicolumn{4}{c}{Natural ESG by centre} \\
\cmidrule(lr){2-5}
Model & 1 ($n = 21$) & 2 ($n = 27$) & 3 ($n = 26$) & 4 ($n = 22$) \\
\midrule
\msn{}, cross-validated & $0.71$ & $0.41$ & $0.15$ & $0.73$ \\
\msp{} with colonoscopy pairs, cross-validated & $0.62$ & $0.44$ & $0.27$ & $0.73$ \\
\ms{} & $0.48$ & $0.11$ & $-0.04$ & $0.36$ \\
\msp{} & $0.43$ & $0.19$ & $0.08$ & $0.32$ \\
\base{} & $0.43$ & $-0.04$ & $-0.19$ & $0.68$ \\
\gem{} & $0.48$ & $0.33$ & $0.00$ & $0.59$ \\
\opus{} & $0.52$ & $0.15$ & $0.12$ & $0.59$ \\
\gpt{} & $0.29$ & $0.15$ & $-0.15$ & $0.55$ \\
\bottomrule
\end{tabular}
\end{adjustbox}
\end{table}

\begin{table}[!htbp]
\caption{REAL-Colon, leave-one-centre-out (pre-specified): \msp{} with colonoscopy pairs was trained four times, each time without every video of one of the four centres (45 training videos each), and each of the 103 pairs was answered by the model that had used no video of its centre. Shown against the same model cross-validated over videos (centres shared) and against the answers of the other models. Columns and intervals as in Supplementary Table~\ref{tab:s_realcolon}; the lower panel gives natural ESG within the pairs of each centre (for the first row, the centre withheld from training). The pre-specified reading compares the leave-one-centre-out answers with each reference model: above each with the interval above zero, above none, or mixed. The result is mixed (above \gpt{} only). The leave-one-centre-out model called 60 polyps present, 50 with a correct box. On the 96 context-matched frames its natural ESG was $0.427$ $\ci{0.286}{0.562}$, with differences of $+0.094$ $\ci{-0.042}{+0.220}$, $+0.104$ $\ci{-0.030}{+0.239}$ and $+0.240$ $\ci{+0.101}{+0.375}$ from \gem{}, \opus{} and \gpt{}. This analysis was not repeated for \msn{}.}
\label{tab:s_loco}
\centering
\footnotesize
\begin{adjustbox}{max width=\linewidth}
\begin{tabular}{l cc c cc}
\toprule
Model & \shortstack{Polyp frames\\found} & \shortstack{Polyp-free frames\\absent or referred} & \shortstack{Natural ESG\\{}[95\% CI]} & \shortstack{Leave-one-centre-out\\minus model [95\% CI]} & $p$ \\
\midrule
\msp{} with colonoscopy pairs, leave-one-centre-out & $58.3\%$ & $91.3\%$ & $0.495$ $\ci{0.370}{0.612}$ & -- & -- \\
\msp{} with colonoscopy pairs, cross-validated & $63.1\%$ & $92.2\%$ & $0.553$ $\ci{0.430}{0.670}$ & $-0.058$ $\ci{-0.132}{+0.010}$ & $0.119$ \\
\ms{} & $37.9\%$ & $86.4\%$ & $0.243$ $\ci{0.149}{0.340}$ & $+0.252$ $\ci{+0.147}{+0.360}$ & $<0.001$ \\
\msp{} & $39.8\%$ & $86.4\%$ & $0.262$ $\ci{0.168}{0.357}$ & $+0.233$ $\ci{+0.120}{+0.343}$ & $<0.001$ \\
\base{} & $38.8\%$ & $85.4\%$ & $0.243$ $\ci{0.143}{0.350}$ & $+0.252$ $\ci{+0.135}{+0.367}$ & $0.001$ \\
\gem{} & $68.0\%$ & $78.6\%$ & $0.466$ $\ci{0.336}{0.581}$ & $+0.029$ $\ci{-0.091}{+0.152}$ & $0.696$ \\
\opus{} & $70.9\%$ & $73.8\%$ & $0.447$ $\ci{0.317}{0.579}$ & $+0.049$ $\ci{-0.072}{+0.174}$ & $0.494$ \\
\gpt{} & $64.1\%$ & $68.0\%$ & $0.320$ $\ci{0.212}{0.437}$ & $+0.175$ $\ci{+0.039}{+0.311}$ & $0.013$ \\
\bottomrule
\end{tabular}
\end{adjustbox}

\vspace{0.6em}
\begin{adjustbox}{max width=\linewidth}
\begin{tabular}{l cccc}
\toprule
 & \multicolumn{4}{c}{Natural ESG by centre} \\
\cmidrule(lr){2-5}
Model & 1 ($n = 23$) & 2 ($n = 28$) & 3 ($n = 28$) & 4 ($n = 24$) \\
\midrule
\msp{} with colonoscopy pairs, leave-one-centre-out & $0.61$ & $0.39$ & $0.29$ & $0.75$ \\
\msp{} with colonoscopy pairs, cross-validated & $0.61$ & $0.50$ & $0.39$ & $0.75$ \\
\ms{} & $0.39$ & $0.11$ & $0.21$ & $0.29$ \\
\msp{} & $0.52$ & $0.11$ & $0.21$ & $0.25$ \\
\base{} & $0.35$ & $0.04$ & $0.14$ & $0.50$ \\
\gem{} & $0.57$ & $0.43$ & $0.36$ & $0.54$ \\
\opus{} & $0.43$ & $0.43$ & $0.29$ & $0.67$ \\
\gpt{} & $0.39$ & $0.21$ & $0.25$ & $0.46$ \\
\bottomrule
\end{tabular}
\end{adjustbox}
\end{table}

\begin{table}[!htbp]
\caption{LDPolypVideo, 116 unedited frame pairs from 160 clips of one hospital, not used in training or model selection (pre-specified confirmatory test): all models. The dataset holds $40{,}186$ labelled frames ($33{,}875$ with a polyp, $6{,}311$ polyp-free); one pair was built per clip, and 44 clips gave none (39 had no polyp-free frame at least 5 frames from a polyp frame; 5 had no polyp frame with a box of at least $1\%$ of the frame). Columns as in Supplementary Table~\ref{tab:s_realcolon}; brackets are 95\% intervals from $2{,}000$ bootstrap resamples of whole clips. Differences are \msn{} (first seed, the model of the main text) minus the row model. The pre-specified reading compared \msn{} with each reference model; the result was mixed: above two of three, with the difference from \gem{} not shown. Test and TrainValid are the two parts in which the dataset is distributed; no model was trained on either.}
\label{tab:s_ldpolyp}
\centering
\footnotesize
\begin{adjustbox}{max width=\linewidth}
\begin{tabular}{l cc c cc}
\toprule
Model & \shortstack{Polyp frames\\found} & \shortstack{Polyp-free frames\\absent or referred} & \shortstack{Natural ESG\\{}[95\% CI]} & \shortstack{\msn{} minus model\\{}[95\% CI]} & $p$ \\
\midrule
\msn{}, first seed & $62.1\%$ & $99.1\%$ & $0.612$ $\ci{0.526}{0.698}$ & -- & -- \\
\msn{}, second seed & $61.2\%$ & $96.6\%$ & $0.578$ $\ci{0.483}{0.672}$ & $+0.034$ $\ci{-0.017}{+0.095}$ & $0.312$ \\
\msn{}, third seed & $62.1\%$ & $95.7\%$ & $0.578$ $\ci{0.491}{0.664}$ & $+0.034$ $\ci{-0.017}{+0.095}$ & $0.300$ \\
\msn{}, three-seed vote & $61.2\%$ & $97.4\%$ & $0.586$ $\ci{0.491}{0.673}$ & $+0.026$ $\ci{-0.009}{+0.069}$ & $0.238$ \\
\ms{} & $45.7\%$ & $96.6\%$ & $0.422$ $\ci{0.328}{0.517}$ & $+0.190$ $\ci{+0.112}{+0.267}$ & $<0.001$ \\
\msp{} with colonoscopy pairs & $59.5\%$ & $99.1\%$ & $0.586$ $\ci{0.491}{0.672}$ & $+0.026$ $\ci{-0.017}{+0.069}$ & $0.351$ \\
\base{} & $40.5\%$ & $95.7\%$ & $0.362$ $\ci{0.276}{0.448}$ & $+0.250$ $\ci{+0.155}{+0.353}$ & $<0.001$ \\
\gem{} & $71.6\%$ & $81.0\%$ & $0.526$ $\ci{0.422}{0.621}$ & $+0.086$ $\ci{-0.017}{+0.190}$ & $0.117$ \\
\opus{} & $62.9\%$ & $83.6\%$ & $0.466$ $\ci{0.362}{0.569}$ & $+0.147$ $\ci{+0.043}{+0.250}$ & $0.010$ \\
\gpt{} & $55.2\%$ & $75.0\%$ & $0.302$ $\ci{0.198}{0.405}$ & $+0.310$ $\ci{+0.181}{+0.431}$ & $<0.001$ \\
\bottomrule
\end{tabular}
\end{adjustbox}

\vspace{0.6em}
\begin{adjustbox}{max width=\linewidth}
\begin{tabular}{l cc cc}
\toprule
 & \multicolumn{2}{c}{Polyp frames, $n$ of $116$ (\%)} & \multicolumn{2}{c}{Natural ESG by part of the dataset} \\
\cmidrule(lr){2-3}\cmidrule(lr){4-5}
Model & Called present & With a correct box & Test ($n = 51$) & TrainValid ($n = 65$) \\
\midrule
\msn{}, first seed & $72$ ($62.1\%$) & $66$ ($56.9\%$) & $0.59$ & $0.63$ \\
\msn{}, second seed & $71$ ($61.2\%$) & $65$ ($56.0\%$) & $0.59$ & $0.57$ \\
\msn{}, third seed & $71$ ($61.2\%$) & $66$ ($56.9\%$) & $0.59$ & $0.57$ \\
\msn{}, three-seed vote & $71$ ($61.2\%$) & $65$ ($56.0\%$) & $0.57$ & $0.60$ \\
\ms{} & $53$ ($45.7\%$) & $50$ ($43.1\%$) & $0.39$ & $0.45$ \\
\msp{} with colonoscopy pairs & $66$ ($56.9\%$) & $60$ ($51.7\%$) & $0.61$ & $0.57$ \\
\base{} & $46$ ($39.7\%$) & $13$ ($11.2\%$) & $0.29$ & $0.42$ \\
\gem{} & $83$ ($71.6\%$) & $76$ ($65.5\%$) & $0.51$ & $0.54$ \\
\opus{} & $73$ ($62.9\%$) & $66$ ($56.9\%$) & $0.47$ & $0.46$ \\
\gpt{} & $64$ ($55.2\%$) & $55$ ($47.4\%$) & $0.29$ & $0.31$ \\
\bottomrule
\end{tabular}
\end{adjustbox}
\end{table}

\begin{table}[!htbp]
\caption{LDPolypVideo pairs restricted by the distance between the polyp frame and the polyp-free frame of a pair (descriptive). A polyp-free frame close to a polyp frame may still show part of the polyp; the restriction tests whether such pairs drive the result. Brackets are 95\% intervals from 2,000 bootstrap resamples of whole clips.}
\label{tab:s_framegap}
\centering
\small
\begin{adjustbox}{max width=\linewidth}
\begin{tabular}{lcccc}
\toprule
 & \multicolumn{2}{c}{At least 25 frames apart (115 pairs)} & \multicolumn{2}{c}{At least 50 frames apart (98 pairs)} \\
\cmidrule(lr){2-3}\cmidrule(lr){4-5}
Model & Natural ESG & \msn{} minus model, $p$ & Natural ESG & \msn{} minus model, $p$ \\
\midrule
\msn & $0.609$ $\ci{0.522}{0.696}$ & -- & $0.622$ $\ci{0.520}{0.714}$ & -- \\
\msn, three-seed vote & $0.583$ & $+0.026$, $0.238$ & $0.592$ & $+0.031$, $0.231$ \\
\ms & $0.417$ & $+0.191$ $\ci{+0.113}{+0.278}$, $<0.001$ & $0.408$ & $+0.214$ $\ci{+0.132}{+0.306}$, $<0.001$ \\
\base & $0.357$ & $+0.252$ $\ci{+0.148}{+0.357}$, $<0.001$ & $0.357$ & $+0.265$ $\ci{+0.153}{+0.378}$, $<0.001$ \\
\gem & $0.522$ & $+0.087$ $\ci{-0.017}{+0.191}$, $0.126$ & $0.531$ & $+0.092$ $\ci{-0.031}{+0.204}$, $0.161$ \\
\opus & $0.461$ & $+0.148$ $\ci{+0.043}{+0.252}$, $0.004$ & $0.480$ & $+0.143$ $\ci{+0.031}{+0.255}$, $0.020$ \\
\gpt & $0.304$ & $+0.304$ $\ci{+0.174}{+0.435}$, $<0.001$ & $0.316$ & $+0.306$ $\ci{+0.173}{+0.439}$, $<0.001$ \\
\bottomrule
\end{tabular}
\end{adjustbox}
\\[2pt]\parbox{\linewidth}{\footnotesize \msn{} answered ``absent'' or referred on $99.1\%$ and $100.0\%$ of the polyp-free frames of the two subsets. Every difference kept the sign it had on all 116 pairs (Supplementary Table~\ref{tab:s_ldpolyp}).}
\end{table}

\clearpage
\subsection{Chest films}
\label{subsec:s_chest}

Supplementary Table~\ref{tab:s_chest} gives every model on the two sets of unedited pneumothorax films, and Supplementary Table~\ref{tab:s_cxrpairs} the chest X-ray follow-up pairs.

\begin{table}[!htbp]
\caption{Pneumothorax on unedited chest films: all models on both sets. CheXthought test patients: 300 pneumothorax films and 300 frontal films without pneumothorax, from patients not used in training (descriptive; no decision rule); pneumothorax films are those labelled positive in the image source and stated positive, with click points, by at least one CheXthought radiologist. New SIIM-ACR patients: 100 pneumothorax films and 100 films without pneumothorax; the pre-specified comparison is \msn{} (first seed) against \msp{} on pneumothorax films called present, with ``raised'' defined as an interval above zero and films without pneumothorax called positive not higher by more than 5 percentage points; the criterion was not met. The reference models were not run on the SIIM-ACR films. Top: each model's values and the difference in Youden index. Bottom: the differences in the two shares that make up the index, as differences in proportions. Correct box: number of pneumothorax films called present with a box overlapping the reference box (intersection over union at least 0.3; on CheXthought the box around the radiologists' click points), of 300 CheXthought or 100 SIIM-ACR films. Youden index: share of pneumothorax films called present plus share of films without pneumothorax answered ``absent'' or referred, minus one. CI: 95\% confidence interval from $2{,}000$ bootstrap resamples of whole patients. Differences are \msn{} (first seed) minus the row model.}
\label{tab:s_chest}
\centering
\footnotesize
\begin{adjustbox}{max width=\linewidth}
\begin{tabular}{l ccc c cc}
\toprule
Model & \shortstack{Pneumothorax films\\called present} & \shortstack{Correct\\box, $n$} & \shortstack{Films without pneumothorax\\absent or referred} & \shortstack{Youden index\\{}[95\% CI]} & \shortstack{\msn{} minus model,\\Youden index [95\% CI]} & $p$ \\
\midrule
\multicolumn{7}{l}{\emph{CheXthought test patients ($300 + 300$ films)}} \\
\msn{}, first seed & $75.0\%$ & $65$ & $91.0\%$ & $0.660$ $\ci{0.591}{0.726}$ & -- & -- \\
\msn{}, second seed & $79.7\%$ & $78$ & $86.7\%$ & $0.663$ $\ci{0.592}{0.731}$ & $-0.003$ $\ci{-0.049}{+0.045}$ & $0.927$ \\
\msn{}, third seed & $75.0\%$ & $70$ & $92.3\%$ & $0.673$ $\ci{0.606}{0.735}$ & $-0.013$ $\ci{-0.054}{+0.030}$ & $0.602$ \\
\msn{}, three-seed vote & $78.7\%$ & $65$ & $91.3\%$ & $0.700$ $\ci{0.633}{0.763}$ & $-0.040$ $\ci{-0.071}{-0.010}$ & $0.008$ \\
\msp{} with CheXthought films & $80.7\%$ & $61$ & $84.3\%$ & $0.650$ $\ci{0.582}{0.716}$ & $+0.010$ $\ci{-0.035}{+0.057}$ & $0.721$ \\
\msp{} & $35.7\%$ & $38$ & $93.0\%$ & $0.287$ $\ci{0.195}{0.378}$ & $+0.373$ $\ci{+0.280}{+0.461}$ & $<0.001$ \\
\ms{} & $0.0\%$ & $0$ & $100.0\%$ & $0.000$ $\ci{-0.135}{0.127}$ & $+0.660$ $\ci{+0.518}{+0.804}$ & $<0.001$ \\
\base{} & $6.3\%$ & $7$ & $85.7\%$ & $-0.080$ $\ci{-0.193}{0.034}$ & $+0.740$ $\ci{+0.618}{+0.865}$ & $<0.001$ \\
\gem{} & $62.0\%$ & $82$ & $84.0\%$ & $0.460$ $\ci{0.374}{0.535}$ & $+0.200$ $\ci{+0.122}{+0.284}$ & $<0.001$ \\
\opus{} & $37.0\%$ & $21$ & $83.7\%$ & $0.207$ $\ci{0.111}{0.306}$ & $+0.453$ $\ci{+0.344}{+0.560}$ & $<0.001$ \\
\gpt{} & $41.0\%$ & $35$ & $96.7\%$ & $0.377$ $\ci{0.286}{0.459}$ & $+0.283$ $\ci{+0.196}{+0.375}$ & $<0.001$ \\
\midrule
\multicolumn{7}{l}{\emph{New SIIM-ACR patients ($100 + 100$ films)}} \\
\msn{}, first seed & $54.0\%$ & $41$ & $96.0\%$ & $0.500$ $\ci{0.380}{0.610}$ & -- & -- \\
\msn{}, second seed & $75.0\%$ & $49$ & $93.0\%$ & $0.680$ $\ci{0.570}{0.780}$ & $-0.180$ $\ci{-0.280}{-0.090}$ & $<0.001$ \\
\msn{}, third seed & $70.0\%$ & $38$ & $94.0\%$ & $0.640$ $\ci{0.530}{0.750}$ & $-0.140$ $\ci{-0.230}{-0.060}$ & $<0.001$ \\
\msn{}, three-seed vote & $68.0\%$ & $49$ & $94.0\%$ & $0.620$ $\ci{0.510}{0.730}$ & $-0.120$ $\ci{-0.200}{-0.040}$ & $0.005$ \\
\msp{} with CheXthought films & $69.0\%$ & $49$ & $93.0\%$ & $0.620$ $\ci{0.510}{0.730}$ & $-0.120$ $\ci{-0.210}{-0.030}$ & $0.013$ \\
\msp{} with colonoscopy pairs & $57.0\%$ & $31$ & $93.0\%$ & $0.500$ $\ci{0.380}{0.620}$ & $+0.000$ $\ci{-0.090}{+0.090}$ & $1.000$ \\
\msp{} & $53.0\%$ & $33$ & $95.0\%$ & $0.480$ $\ci{0.360}{0.600}$ & $+0.020$ $\ci{-0.070}{+0.110}$ & $0.753$ \\
\ms{} & $0.0\%$ & $0$ & $100.0\%$ & $0.000$ $\ci{-0.140}{0.140}$ & $+0.500$ $\ci{+0.360}{+0.630}$ & $<0.001$ \\
\bottomrule
\end{tabular}
\end{adjustbox}

\vspace{0.6em}
\begin{adjustbox}{max width=\linewidth}
\begin{tabular}{l cc c}
\toprule
 & \multicolumn{3}{c}{\msn{} (first seed) minus model, difference in proportion [95\% CI]} \\
\cmidrule(lr){2-4}
Model & Pneumothorax films called present & $p$ & Films without pneumothorax called positive \\
\midrule
\multicolumn{4}{l}{\emph{CheXthought test patients ($300 + 300$ films)}} \\
\msn{}, second seed & $-0.047$ $\ci{-0.083}{-0.012}$ & $0.007$ & $-0.043$ $\ci{-0.072}{-0.017}$ \\
\msn{}, third seed & $+0.000$ $\ci{-0.036}{+0.036}$ & $1.000$ & $+0.013$ $\ci{-0.007}{+0.035}$ \\
\msn{}, three-seed vote & $-0.037$ $\ci{-0.063}{-0.013}$ & $0.003$ & $+0.003$ $\ci{-0.010}{+0.017}$ \\
\msp{} with CheXthought films & $-0.057$ $\ci{-0.087}{-0.029}$ & $<0.001$ & $-0.067$ $\ci{-0.099}{-0.037}$ \\
\msp{} & $+0.393$ $\ci{+0.328}{+0.455}$ & $<0.001$ & $+0.020$ $\ci{-0.014}{+0.055}$ \\
\ms{} & $+0.750$ $\ci{+0.688}{+0.805}$ & $<0.001$ & $+0.090$ $\ci{+0.058}{+0.129}$ \\
\base{} & $+0.687$ $\ci{+0.621}{+0.746}$ & $<0.001$ & $+0.087$ $\ci{+0.055}{+0.123}$ \\
\gem{} & $+0.130$ $\ci{+0.066}{+0.199}$ & $<0.001$ & $-0.060$ $\ci{-0.104}{-0.016}$ \\
\opus{} & $+0.380$ $\ci{+0.297}{+0.451}$ & $<0.001$ & $-0.073$ $\ci{-0.120}{-0.026}$ \\
\gpt{} & $+0.340$ $\ci{+0.278}{+0.410}$ & $<0.001$ & $+0.057$ $\ci{+0.024}{+0.092}$ \\
\midrule
\multicolumn{4}{l}{\emph{New SIIM-ACR patients ($100 + 100$ films)}} \\
\msn{}, second seed & $-0.210$ $\ci{-0.290}{-0.130}$ & $<0.001$ & $-0.030$ $\ci{-0.070}{+0.000}$ \\
\msn{}, third seed & $-0.160$ $\ci{-0.240}{-0.090}$ & $<0.001$ & $-0.020$ $\ci{-0.050}{+0.000}$ \\
\msn{}, three-seed vote & $-0.140$ $\ci{-0.210}{-0.070}$ & $<0.001$ & $-0.020$ $\ci{-0.050}{+0.000}$ \\
\msp{} with CheXthought films & $-0.150$ $\ci{-0.230}{-0.070}$ & $0.001$ & $-0.030$ $\ci{-0.070}{+0.000}$ \\
\msp{} with colonoscopy pairs & $-0.030$ $\ci{-0.110}{+0.050}$ & $0.510$ & $-0.030$ $\ci{-0.080}{+0.010}$ \\
\msp{} & $+0.010$ $\ci{-0.080}{+0.100}$ & $0.920$ & $-0.010$ $\ci{-0.050}{+0.020}$ \\
\ms{} & $+0.540$ $\ci{+0.440}{+0.630}$ & $<0.001$ & $+0.040$ $\ci{+0.010}{+0.080}$ \\
\bottomrule
\end{tabular}
\end{adjustbox}
\end{table}

\begin{table}[!htbp]
\caption{Chest X-ray follow-up pairs (secondary; descriptive): a film whose only label is pneumothorax, effusion, nodule or mass, and the nearest film of the same patient and view labelled as showing no finding (NIH ChestX-ray14; labels mined from reports). Two versions are shown. In the planned version, 200 pairs were built (50 per finding) and a near-copy check that compared coarse image hashes removed 156, leaving 44; the result is reported as run. In the corrected version (post hoc) the check also required pixel correlation, and 183 of the same 200 pairs were scored. Brackets: 95\% bootstrap intervals over patients where recorded. Differences are \ms{} minus the row model on the corrected version. By finding, natural ESG of \ms{} was $0.00$ for effusion ($n = 11$), $0.50$ for mass ($n = 10$), $-0.14$ for nodule ($n = 7$) and $-0.06$ for pneumothorax ($n = 16$) in the planned version, and $0.31$, $0.24$, $0.13$ and $0.00$ in the corrected version ($n = 45$ to $46$ each). \msn{} was not run on these pairs.}
\label{tab:s_cxrpairs}
\centering
\footnotesize
\begin{adjustbox}{max width=\linewidth}
\begin{tabular}{l c c cc}
\toprule
 & Planned ($44$ pairs) & \multicolumn{3}{c}{Corrected, post hoc ($183$ pairs)} \\
\cmidrule(lr){2-2}\cmidrule(lr){3-5}
Model & Natural ESG [95\% CI] & Natural ESG [95\% CI] & \ms{} minus model [95\% CI] & $p$ \\
\midrule
\ms{} & $0.068$ $\ci{-0.068}{0.205}$ & $0.169$ $\ci{0.098}{0.240}$ & -- & -- \\
\msp{} & $0.068$ & $0.126$ & n.r. & n.r. \\
\base{} & $0.068$ & $0.071$ $\ci{0.005}{0.137}$ & $+0.098$ $\ci{0.022}{0.175}$ & $0.012$ \\
\gem{} & $0.136$ $\ci{0.000}{0.295}$ & $0.109$ & $+0.060$ (n.r.) & n.r. \\
\opus{} & $0.045$ & $0.115$ & $+0.055$ (n.r.) & n.r. \\
\gpt{} & $0.159$ $\ci{0.000}{0.318}$ & $0.148$ & $+0.022$ (n.r.) & n.r. \\
\bottomrule
\end{tabular}
\end{adjustbox}
\end{table}

\clearpage
\subsection{\texorpdfstring{\msn}{MedScope-4B-N}: training seeds, training mix and the earlier tests}
\label{subsec:s_nmodel}

Supplementary Tables~\ref{tab:s_nseeds} and~\ref{tab:s_nseeds_ref} give the three training seeds of \msn{} and their vote, Supplementary Table~\ref{tab:s_mix} the two candidate training mixes, Supplementary Table~\ref{tab:s_additions} the two single-addition models, Supplementary Tables~\ref{tab:s_nmodel}, \ref{tab:s_next_add} and~\ref{tab:s_source} \msn{} against the other trained models, and Supplementary Table~\ref{tab:s_vqa} standard question answering.

\begin{table}[!htbp]
\caption{\msn{}: three training seeds and their majority vote on every test set. Each seed repeats all three training stages on the same training data. The vote takes, for each view, the answer class chosen by two or three seeds, and needs three model runs per image. Mean (SD) is over the three seeds and reflects variation between training runs; it was computed for the five measures of the counterfactual test sets only. Lesion images correct: share of original and control-edited lesion views answered correctly, which is why it differs from the count of original films called present on the SIIM-ACR set. The pre-specified conditions were met: every seed's main-test ESG exceeded that of each reference model with the paired interval above zero (Supplementary Table~\ref{tab:s_nseeds_ref}), and the vote had main-test lesion-free claims not above the seeds' mean and ESG not more than 0.02 below it. Under the pre-specified rule the vote therefore qualified as the reported operating point; the main text reports the first seed as the primary model, because the confirmatory comparison on LDPolypVideo was specified for the first seed, and gives the vote beside it. --: not computed.}
\label{tab:s_nseeds}
\centering
\footnotesize
\begin{adjustbox}{max width=\linewidth}
\begin{tabular}{ll ccc c c}
\toprule
Test set & Measure & First seed & Second seed & Third seed & Mean (SD) & Three-seed vote \\
\midrule
Main test ($507$ cases) & ESG & $0.744$ & $0.740$ & $0.726$ & $0.737$ ($0.010$) & $0.753$ \\
 & Erased-lesion claims & $9.1\%$ & $13.7\%$ & $14.6\%$ & $12.5\%$ ($2.9\%$) & $10.5\%$ \\
 & Lesion-free claims & $8.7\%$ & $15.3\%$ & $10.8\%$ & $11.6\%$ ($3.4\%$) & $9.0\%$ \\
 & Lesion images correct & $0.847$ & $0.872$ & $0.874$ & $0.865$ ($0.015$) & $0.863$ \\
 & Net utility & $0.709$ & $0.675$ & $0.721$ & $0.702$ ($0.024$) & $0.726$ \\
\midrule
External datasets ($169$ cases) & ESG & $0.672$ & $0.647$ & $0.672$ & $0.664$ ($0.015$) & $0.689$ \\
 & Erased-lesion claims & $10.9\%$ & $10.1\%$ & $17.6\%$ & $12.9\%$ ($4.1\%$) & $11.8\%$ \\
 & Lesion-free claims & $2.0\%$ & $8.0\%$ & $16.0\%$ & $8.7\%$ ($7.0\%$) & $2.0\%$ \\
 & Lesion images correct & $0.777$ & $0.761$ & $0.836$ & $0.791$ ($0.040$) & $0.803$ \\
 & Net utility & $0.625$ & $0.593$ & $0.619$ & $0.612$ ($0.017$) & $0.659$ \\
\midrule
Fresh internal ($76$ cases) & ESG & $0.417$ & $0.556$ & $0.528$ & $0.500$ ($0.073$) & $0.472$ \\
 & Erased-lesion claims & $25.0\%$ & $27.8\%$ & $19.4\%$ & $24.1\%$ ($4.2\%$) & $25.0\%$ \\
 & Lesion-free claims & $15.0\%$ & $15.0\%$ & $15.0\%$ & $15.0\%$ ($0.0\%$) & $12.5\%$ \\
 & Lesion images correct & $0.681$ & $0.806$ & $0.722$ & $0.736$ ($0.064$) & $0.722$ \\
 & Net utility & $0.372$ & $0.514$ & $0.435$ & $0.440$ ($0.071$) & $0.438$ \\
\midrule
Fresh external ($316$ cases) & ESG & $0.381$ & $0.381$ & $0.411$ & $0.391$ ($0.017$) & $0.399$ \\
 & Erased-lesion claims & $22.0\%$ & $28.6\%$ & $20.2\%$ & $23.6\%$ ($4.4\%$) & $22.6\%$ \\
 & Lesion-free claims & $5.4\%$ & $9.5\%$ & $13.5\%$ & $9.5\%$ ($4.1\%$) & $6.1\%$ \\
 & Lesion images correct & $0.607$ & $0.676$ & $0.601$ & $0.628$ ($0.041$) & $0.619$ \\
 & Net utility & $0.114$ & $0.192$ & $0.094$ & $0.134$ ($0.052$) & $0.147$ \\
\midrule
New SIIM-ACR patients ($200$ cases) & ESG & $0.270$ & $0.140$ & $0.280$ & $0.230$ ($0.078$) & $0.270$ \\
 & Erased-lesion claims & $25.0\%$ & $59.0\%$ & $43.0\%$ & $42.3\%$ ($17.0\%$) & $42.0\%$ \\
 & Lesion-free claims & $4.0\%$ & $7.0\%$ & $6.0\%$ & $5.7\%$ ($1.5\%$) & $6.0\%$ \\
 & Lesion images correct & $0.530$ & $0.740$ & $0.705$ & $0.658$ ($0.113$) & $0.685$ \\
 & Net utility & $-0.134$ & $0.083$ & $0.089$ & $0.013$ ($0.127$) & $0.080$ \\
 & Pneumothorax films called present ($n$ of $100$) & $54$ & $75$ & $70$ & -- & $68$ \\
 & Youden index & $0.500$ & $0.680$ & $0.640$ & -- & $0.620$ \\
\midrule
CheXthought test patients ($300 + 300$ films) & Pneumothorax films called present & $75.0\%$ & $79.7\%$ & $75.0\%$ & -- & $78.7\%$ \\
 & Youden index & $0.660$ & $0.663$ & $0.673$ & -- & $0.700$ \\
\midrule
LDPolypVideo ($116$ pairs) & Natural ESG & $0.612$ & $0.578$ & $0.578$ & -- & $0.586$ \\
\bottomrule
\end{tabular}
\end{adjustbox}
\end{table}

\begin{table}[!htbp]
\caption{\msn{}: each training seed and the three-seed vote against the three reference models on the main test (507 cases). Entries are the row model minus the reference model, paired over cases, with 95\% bootstrap intervals; differences in claims are differences in proportions. All differences in ESG and in erased-lesion claims have $p < 0.001$. Lesion-free claims of the second seed were higher than those of \gem{} ($p = 0.011$).}
\label{tab:s_nseeds_ref}
\centering
\footnotesize
\begin{adjustbox}{max width=\linewidth}
\begin{tabular}{ll ccccc}
\toprule
\msn{} & Reference model & ESG & Erased-lesion claims & Lesion-free claims & Lesion images correct & Net utility \\
\midrule
First seed & \gem{} & $+0.274$ $\ci{+0.189}{+0.359}$ & $-0.320$ $\ci{-0.390}{-0.249}$ & $-0.003$ $\ci{-0.048}{+0.040}$ & $-0.027$ $\ci{-0.070}{+0.013}$ & $+0.145$ $\ci{+0.057}{+0.231}$ \\
 & \opus{} & $+0.416$ $\ci{+0.337}{+0.495}$ & $-0.447$ $\ci{-0.522}{-0.373}$ & $-0.149$ $\ci{-0.207}{-0.097}$ & $-0.016$ $\ci{-0.060}{+0.029}$ & $+0.352$ $\ci{+0.261}{+0.440}$ \\
 & \gpt{} & $+0.438$ $\ci{+0.350}{+0.525}$ & $-0.498$ $\ci{-0.568}{-0.425}$ & $-0.191$ $\ci{-0.251}{-0.138}$ & $-0.039$ $\ci{-0.084}{+0.007}$ & $+0.401$ $\ci{+0.309}{+0.503}$ \\
\midrule
Second seed & \gem{} & $+0.269$ $\ci{+0.185}{+0.351}$ & $-0.274$ $\ci{-0.348}{-0.199}$ & $+0.063$ $\ci{+0.017}{+0.114}$ & $-0.002$ $\ci{-0.039}{+0.032}$ & $+0.111$ $\ci{+0.024}{+0.192}$ \\
 & \opus{} & $+0.411$ $\ci{+0.329}{+0.489}$ & $-0.402$ $\ci{-0.478}{-0.324}$ & $-0.083$ $\ci{-0.142}{-0.023}$ & $+0.009$ $\ci{-0.038}{+0.057}$ & $+0.318$ $\ci{+0.224}{+0.411}$ \\
 & \gpt{} & $+0.434$ $\ci{+0.347}{+0.519}$ & $-0.452$ $\ci{-0.522}{-0.380}$ & $-0.125$ $\ci{-0.188}{-0.064}$ & $-0.014$ $\ci{-0.056}{+0.033}$ & $+0.367$ $\ci{+0.274}{+0.459}$ \\
\midrule
Third seed & \gem{} & $+0.256$ $\ci{+0.171}{+0.338}$ & $-0.265$ $\ci{-0.340}{-0.190}$ & $+0.017$ $\ci{-0.028}{+0.064}$ & $+0.000$ $\ci{-0.039}{+0.039}$ & $+0.156$ $\ci{+0.074}{+0.234}$ \\
 & \opus{} & $+0.397$ $\ci{+0.322}{+0.470}$ & $-0.393$ $\ci{-0.470}{-0.318}$ & $-0.128$ $\ci{-0.184}{-0.071}$ & $+0.011$ $\ci{-0.032}{+0.055}$ & $+0.363$ $\ci{+0.277}{+0.445}$ \\
 & \gpt{} & $+0.420$ $\ci{+0.336}{+0.505}$ & $-0.443$ $\ci{-0.512}{-0.373}$ & $-0.170$ $\ci{-0.233}{-0.114}$ & $-0.011$ $\ci{-0.055}{+0.031}$ & $+0.412$ $\ci{+0.327}{+0.497}$ \\
\midrule
Three-seed vote & \gem{} & $+0.283$ $\ci{+0.203}{+0.364}$ & $-0.306$ $\ci{-0.376}{-0.236}$ & $+0.000$ $\ci{-0.045}{+0.042}$ & $-0.011$ $\ci{-0.050}{+0.027}$ & $+0.162$ $\ci{+0.081}{+0.242}$ \\
 & \opus{} & $+0.425$ $\ci{+0.350}{+0.500}$ & $-0.434$ $\ci{-0.507}{-0.360}$ & $-0.146$ $\ci{-0.201}{-0.092}$ & $+0.000$ $\ci{-0.046}{+0.047}$ & $+0.369$ $\ci{+0.279}{+0.457}$ \\
 & \gpt{} & $+0.447$ $\ci{+0.368}{+0.528}$ & $-0.484$ $\ci{-0.550}{-0.416}$ & $-0.188$ $\ci{-0.247}{-0.133}$ & $-0.023$ $\ci{-0.066}{+0.020}$ & $+0.418$ $\ci{+0.333}{+0.509}$ \\
\bottomrule
\end{tabular}
\end{adjustbox}
\end{table}

\begin{table}[!htbp]
\caption{The two candidate training mixes of \msn{}. Top: composition of the last training stage. Adding the unedited images lowers the share of erased-lesion views; the second candidate repeats erased-lesion views until their share equals that in \msp{}. Middle: validation images only (main validation set, 540 cases and $1{,}473$ views; SIIM-ACR validation patients, 98 cases and 242 views). The rule, fixed before either candidate was trained, selected the second candidate unless the first had a higher net utility over both validation sets together by more than 0.005; the values were $0.547$ and $0.452$, so the first candidate (no repetition) is \msn{}. It also met the pre-specified limits on the main validation set (net utility $0.642$, limit at least $0.593$; lesion-free claims $9.1\%$, limit at most $9.7\%$). Bottom: both candidates on three test sets, run as planned for description only, with differences (selected minus not selected) paired over cases and 95\% bootstrap intervals.}
\label{tab:s_mix}
\centering
\footnotesize
\begin{adjustbox}{max width=\linewidth}
\begin{tabular}{l ccc}
\toprule
Last training stage & Training views & Erased-lesion views (share) & Lesion-free images (share) \\
\midrule
\msp{} & $6{,}217$ & $1{,}756$ ($28.2\%$) & $420$ ($6.8\%$) \\
\msn{} (selected: no repetition) & $12{,}292$ & $1{,}223$ ($9.9\%$) & $3{,}801$ ($30.9\%$) \\
Not selected: erased-lesion views repeated & $15{,}428$ & $4{,}359$ ($28.3\%$) & $3{,}801$ ($24.6\%$) \\
\bottomrule
\end{tabular}
\end{adjustbox}

\vspace{0.6em}
\begin{adjustbox}{max width=\linewidth}
\begin{tabular}{ll cccccc}
\toprule
Validation set & Candidate & Net utility [95\% CI] & ESG & Lesion images correct & Erased-lesion claims & Control edit denied & Lesion-free claims \\
\midrule
Main ($540$ cases) & Selected & $0.642$ $\ci{0.556}{0.721}$ & $0.706$ & $0.819$ & $10.6\%$ & $18.8\%$ & $9.1\%$ \\
 & Not selected & $0.621$ $\ci{0.533}{0.704}$ & $0.716$ & $0.792$ & $7.5\%$ & $20.9\%$ & $10.0\%$ \\
SIIM-ACR patients ($98$ cases) & Selected & $-0.031$ $\ci{-0.313}{0.272}$ & $0.292$ & $0.604$ & $31.3\%$ & $39.6\%$ & $8.0\%$ \\
 & Not selected & $-0.576$ $\ci{-0.851}{-0.267}$ & $0.167$ & $0.302$ & $14.6\%$ & $68.8\%$ & $8.0\%$ \\
\bottomrule
\end{tabular}
\end{adjustbox}

\vspace{0.6em}
\begin{adjustbox}{max width=\linewidth}
\begin{tabular}{ll cc cc}
\toprule
Test set & Measure & Selected (\msn{}) & Not selected & Difference [95\% CI] & $p$ \\
\midrule
Main test ($507$ cases) & ESG & $0.744$ & $0.772$ & $-0.027$ $\ci{-0.077}{+0.020}$ & $0.270$ \\
 & Erased-lesion claims & $9.1\%$ & $8.2\%$ & $+0.009$ $\ci{-0.026}{+0.047}$ & $0.693$ \\
 & Lesion-free claims & $8.7\%$ & $8.0\%$ & $+0.007$ $\ci{-0.020}{+0.034}$ & $0.676$ \\
 & Lesion images correct & $0.847$ & $0.872$ & $-0.025$ $\ci{-0.061}{+0.011}$ & $0.193$ \\
 & Net utility & $0.709$ & $0.765$ & $-0.056$ $\ci{-0.115}{+0.003}$ & $0.062$ \\
\midrule
Fresh external ($316$ cases) & ESG & $0.381$ & $0.405$ & $-0.024$ $\ci{-0.077}{+0.031}$ & $0.413$ \\
 & Erased-lesion claims & $22.0\%$ & $17.3\%$ & $+0.048$ $\ci{+0.006}{+0.092}$ & $0.035$ \\
 & Lesion-free claims & $5.4\%$ & $7.4\%$ & $-0.020$ $\ci{-0.058}{+0.019}$ & $0.423$ \\
 & Lesion images correct & $0.607$ & $0.589$ & $+0.018$ $\ci{-0.020}{+0.056}$ & $0.395$ \\
 & Net utility & $0.114$ & $0.107$ & $+0.007$ $\ci{-0.068}{+0.078}$ & $0.878$ \\
\midrule
New SIIM-ACR patients ($200$ cases) & ESG & $0.270$ & $0.220$ & $+0.050$ $\ci{-0.050}{+0.151}$ & $0.349$ \\
 & Erased-lesion claims & $25.0\%$ & $10.0\%$ & $+0.150$ $\ci{+0.054}{+0.250}$ & $0.002$ \\
 & Lesion-free claims & $4.0\%$ & $5.0\%$ & $-0.010$ $\ci{-0.046}{+0.022}$ & $0.798$ \\
 & Lesion images correct & $0.530$ & $0.320$ & $+0.210$ $\ci{+0.111}{+0.307}$ & $<0.001$ \\
 & Net utility & $-0.134$ & $-0.512$ & $+0.378$ $\ci{+0.195}{+0.567}$ & $<0.001$ \\
\bottomrule
\end{tabular}
\end{adjustbox}
\end{table}

\begin{table}[!htbp]
\caption{The two single-addition models on the earlier test sets. Top: \msp{} with colonoscopy pairs (all 60 REAL-Colon videos in training; one seed; post hoc) against \ms{} and \msp{}; differences are this model minus the other, paired over cases, with 95\% bootstrap intervals; all comparisons with \ms{} on the SIIM-ACR set have $p \le 0.001$. Bottom: \msp{} with CheXthought films ($1{,}500$ training films with and $1{,}500$ without pneumothorax; one seed; post hoc) against \msp{}. Its pre-specified primary comparison was pneumothorax films called present on the new SIIM-ACR patients, with ``raised'' defined as an interval above zero and films without pneumothorax called positive not higher by more than 5 percentage points; the criterion was met. Both models met their pre-specified limits on validation images before any test was run (net utility $0.654$, limit at least $0.593$, and lesion-free claims $9.1\%$, limit at most $9.7\%$, for the first; $0.628$, limit at least $0.611$, and $9.5\%$, limit at most $10.6\%$, for the second). Their results on unedited pairs and chest films are in Supplementary Tables~\ref{tab:s_realcolon}, \ref{tab:s_ldpolyp} and~\ref{tab:s_chest}. --: not computed.}
\label{tab:s_additions}
\centering
\footnotesize
\begin{adjustbox}{max width=\linewidth}
\begin{tabular}{ll c c cc c cc}
\toprule
Test set & Measure & \shortstack{\msp{} with\\colonoscopy pairs} & \ms{} & Difference [95\% CI] & $p$ & \msp{} & Difference [95\% CI] & $p$ \\
\midrule
Main test ($507$ cases) & ESG & $0.744$ & $0.758$ & $-0.014$ $\ci{-0.062}{+0.036}$ & $0.652$ & $0.740$ & $+0.005$ $\ci{-0.044}{+0.051}$ & $0.956$ \\
 & Erased-lesion claims & $11.0\%$ & $9.6\%$ & $+0.014$ $\ci{-0.027}{+0.055}$ & $0.608$ & $11.0\%$ & $+0.000$ $\ci{-0.036}{+0.035}$ & $1.000$ \\
 & Lesion-free claims & $8.7\%$ & $7.6\%$ & $+0.010$ $\ci{-0.021}{+0.042}$ & $0.572$ & $9.0\%$ & $-0.003$ $\ci{-0.034}{+0.026}$ & $0.907$ \\
 & Lesion images correct & $0.854$ & $0.849$ & $+0.005$ $\ci{-0.026}{+0.036}$ & $0.840$ & $0.856$ & $-0.002$ $\ci{-0.027}{+0.022}$ & $0.907$ \\
 & Net utility & $0.707$ & $0.732$ & $-0.025$ $\ci{-0.082}{+0.029}$ & $0.420$ & $0.714$ & $-0.007$ $\ci{-0.054}{+0.038}$ & $0.757$ \\
\midrule
External datasets ($169$ cases) & ESG & $0.681$ & $0.597$ & $+0.084$ $\ci{-0.000}{+0.172}$ & $0.073$ & $0.739$ & $-0.059$ $\ci{-0.132}{+0.017}$ & $0.156$ \\
 & Erased-lesion claims & $11.8\%$ & $11.8\%$ & $+0.000$ $\ci{-0.050}{+0.052}$ & $1.000$ & $6.7\%$ & $+0.050$ $\ci{+0.017}{+0.095}$ & $0.005$ \\
 & Lesion-free claims & $2.0\%$ & $4.0\%$ & $-0.020$ $\ci{-0.093}{+0.042}$ & $0.802$ & $6.0\%$ & $-0.040$ $\ci{-0.104}{+0.000}$ & $0.276$ \\
 & Lesion images correct & $0.798$ & $0.727$ & $+0.071$ $\ci{+0.014}{+0.131}$ & $0.019$ & $0.815$ & $-0.017$ $\ci{-0.068}{+0.036}$ & $0.618$ \\
 & Net utility & $0.645$ & $0.530$ & $+0.115$ $\ci{+0.018}{+0.217}$ & $0.022$ & $0.696$ & $-0.050$ $\ci{-0.131}{+0.032}$ & $0.270$ \\
\midrule
Fresh internal ($76$ cases) & ESG & $0.500$ & $0.472$ & $+0.028$ $\ci{-0.189}{+0.270}$ & $0.858$ & $0.528$ & $-0.028$ $\ci{-0.154}{+0.103}$ & $0.760$ \\
 & Erased-lesion claims & $33.3\%$ & $19.4\%$ & $+0.139$ $\ci{-0.026}{+0.296}$ & $0.132$ & $25.0\%$ & $+0.083$ $\ci{+0.000}{+0.184}$ & $0.106$ \\
 & Lesion-free claims & $20.0\%$ & $12.5\%$ & $+0.075$ $\ci{-0.095}{+0.244}$ & $0.470$ & $17.5\%$ & $+0.025$ $\ci{-0.105}{+0.158}$ & $0.861$ \\
 & Lesion images correct & $0.819$ & $0.681$ & $+0.139$ $\ci{-0.029}{+0.324}$ & $0.127$ & $0.778$ & $+0.042$ $\ci{-0.088}{+0.167}$ & $0.617$ \\
 & Net utility & $0.484$ & $0.402$ & $+0.082$ $\ci{-0.176}{+0.349}$ & $0.533$ & $0.500$ & $-0.016$ $\ci{-0.192}{+0.143}$ & $0.879$ \\
\midrule
Fresh external ($316$ cases) & ESG & $0.387$ & $0.393$ & $-0.006$ $\ci{-0.074}{+0.058}$ & $0.880$ & $0.417$ & $-0.030$ $\ci{-0.088}{+0.032}$ & $0.349$ \\
 & Erased-lesion claims & $19.6\%$ & $8.3\%$ & $+0.113$ $\ci{+0.052}{+0.175}$ & $0.001$ & $11.3\%$ & $+0.083$ $\ci{+0.031}{+0.138}$ & $0.003$ \\
 & Lesion-free claims & $4.1\%$ & $4.7\%$ & $-0.007$ $\ci{-0.042}{+0.028}$ & $0.880$ & $6.8\%$ & $-0.027$ $\ci{-0.066}{+0.008}$ & $0.216$ \\
 & Lesion images correct & $0.577$ & $0.476$ & $+0.101$ $\ci{+0.043}{+0.162}$ & $0.003$ & $0.536$ & $+0.042$ $\ci{-0.006}{+0.092}$ & $0.106$ \\
 & Net utility & $0.088$ & $-0.053$ & $+0.141$ $\ci{+0.044}{+0.241}$ & $0.006$ & $0.048$ & $+0.040$ $\ci{-0.048}{+0.128}$ & $0.355$ \\
\midrule
New SIIM-ACR patients ($200$ cases) & ESG & $0.400$ & $0.000$ & $+0.400$ $\ci{+0.299}{+0.500}$ & $\le 0.001$ & $0.360$ & $+0.040$ $\ci{-0.060}{+0.140}$ & $0.482$ \\
 & Erased-lesion claims & $20.0\%$ & $0.0\%$ & $+0.200$ $\ci{+0.125}{+0.280}$ & $\le 0.001$ & $19.0\%$ & $+0.010$ $\ci{-0.063}{+0.083}$ & $0.933$ \\
 & Lesion-free claims & $7.0\%$ & $0.0\%$ & $+0.070$ $\ci{+0.021}{+0.125}$ & $\le 0.001$ & $5.0\%$ & $+0.020$ $\ci{+0.000}{+0.052}$ & $0.284$ \\
 & Lesion images correct & $0.585$ & $0.000$ & $+0.585$ $\ci{+0.490}{+0.670}$ & $\le 0.001$ & $0.540$ & $+0.045$ $\ci{-0.043}{+0.131}$ & $0.342$ \\
 & Net utility & $-0.028$ & $-1.100$ & $+1.072$ $\ci{+0.873}{+1.263}$ & $\le 0.001$ & $-0.096$ & $+0.068$ $\ci{-0.116}{+0.253}$ & $0.470$ \\
\bottomrule
\end{tabular}
\end{adjustbox}

\vspace{0.6em}
\begin{adjustbox}{max width=\linewidth}
\begin{tabular}{ll cc cc}
\toprule
Test set & Measure & \shortstack{\msp{} with\\CheXthought films} & \msp{} & Difference [95\% CI] & $p$ \\
\midrule
New SIIM-ACR patients ($100 + 100$ films) & Pneumothorax films called present & $69$ of $100$ & $53$ of $100$ & $+0.160$ $\ci{+0.070}{+0.260}$ & $0.002$ \\
 & Called present with a correct box & $49$ of $100$ & $33$ of $100$ & -- & -- \\
 & Films without pneumothorax called positive & $7$ of $100$ & $5$ of $100$ & $+0.020$ $\ci{-0.020}{+0.060}$ & -- \\
 & Youden index & $0.620$ $\ci{0.510}{0.730}$ & $0.480$ & $+0.140$ $\ci{+0.040}{+0.250}$ & $0.010$ \\
 & ESG & $0.380$ $\ci{0.278}{0.476}$ & $0.360$ & -- & -- \\
 & Erased-lesion claims & $31.0\%$ & $19.0\%$ & -- & -- \\
 & Lesion images correct & $0.690$ & $0.540$ & -- & -- \\
 & Net utility & $0.145$ & $-0.096$ & -- & -- \\
\midrule
Main test ($507$ cases) & ESG & $0.753$ $\ci{0.692}{0.811}$ & $0.740$ & $+0.014$ $\ci{-0.037}{+0.064}$ & -- \\
 & Erased-lesion claims & $7.3\%$ & $11.0\%$ & $-0.037$ $\ci{-0.071}{-0.004}$ & $0.039$ \\
 & Lesion-free claims & $7.6\%$ & $9.0\%$ & n.r. & n.r. \\
 & Lesion images correct & $0.840$ & $0.856$ & -- & -- \\
 & Net utility & $0.725$ & $0.714$ & n.r. & n.r. \\
\midrule
Fresh external ($316$ cases) & ESG & $0.381$ $\ci{0.298}{0.461}$ & $0.417$ & n.r. & n.r. \\
 & Erased-lesion claims & $20.2\%$ & $11.3\%$ & $+0.089$ $\ci{+0.037}{+0.143}$ & $<0.001$ \\
 & Lesion-free claims & $5.4\%$ & $6.8\%$ & -- & -- \\
 & Lesion images correct & $0.583$ & $0.536$ & -- & -- \\
 & Net utility & $0.127$ & $0.048$ & n.r. & n.r. \\
\bottomrule
\end{tabular}
\end{adjustbox}
\end{table}

\begin{table}[!htbp]
\caption{\msn{} (first seed) and its three-seed vote against \ms{} and \msp{} on every test set of the main text except the unedited colonoscopy pairs: the full version of Table~\ref{tab:nmodel}. Lesion images correct: share of original and control-edited lesion images answered correctly. Youden index: share of pneumothorax films called present plus share of films without pneumothorax answered ``absent'' or referred, minus one. The vote needs three model runs per image. Paired differences are in Supplementary Table~\ref{tab:s_next} for the four counterfactual test sets and in Supplementary Table~\ref{tab:s_chest} for the chest films; the three seeds are in Supplementary Table~\ref{tab:s_nseeds}.}
\label{tab:s_nmodel}
\centering
\footnotesize
\begin{adjustbox}{max width=\linewidth}
\begin{tabular}{l cccc}
\toprule
Measure & \ms{} & \msp{} & \msn{} & Three-seed vote \\
\midrule
\multicolumn{5}{l}{\emph{Main test ($507$ cases)}} \\
ESG & $0.758$ & $0.740$ & $0.744$ & $0.753$ \\
Erased-lesion claims & $9.6\%$ & $11.0\%$ & $9.1\%$ & $10.5\%$ \\
Lesion-free claims & $7.6\%$ & $9.0\%$ & $8.7\%$ & $9.0\%$ \\
Lesion images correct & $0.849$ & $0.856$ & $0.847$ & $0.863$ \\
Net utility & $0.732$ & $0.714$ & $0.709$ & $0.726$ \\
\midrule
\multicolumn{5}{l}{\emph{External datasets ($169$ cases)}} \\
ESG & $0.597$ & $0.739$ & $0.672$ & $0.689$ \\
Erased-lesion claims & $11.8\%$ & $6.7\%$ & $10.9\%$ & $11.8\%$ \\
Lesion-free claims & $4.0\%$ & $6.0\%$ & $2.0\%$ & $2.0\%$ \\
Lesion images correct & $0.727$ & $0.815$ & $0.777$ & $0.803$ \\
Net utility & $0.530$ & $0.696$ & $0.625$ & $0.659$ \\
\midrule
\multicolumn{5}{l}{\emph{Fresh internal ($76$ cases)}} \\
ESG & $0.472$ & $0.528$ & $0.417$ & $0.472$ \\
Erased-lesion claims & $19.4\%$ & $25.0\%$ & $25.0\%$ & $25.0\%$ \\
Lesion-free claims & $12.5\%$ & $17.5\%$ & $15.0\%$ & $12.5\%$ \\
Lesion images correct & $0.681$ & $0.778$ & $0.681$ & $0.722$ \\
Net utility & $0.402$ & $0.500$ & $0.372$ & $0.438$ \\
\midrule
\multicolumn{5}{l}{\emph{Fresh external ($316$ cases)}} \\
ESG & $0.393$ & $0.417$ & $0.381$ & $0.399$ \\
Erased-lesion claims & $8.3\%$ & $11.3\%$ & $22.0\%$ & $22.6\%$ \\
Lesion-free claims & $4.7\%$ & $6.8\%$ & $5.4\%$ & $6.1\%$ \\
Lesion images correct & $0.476$ & $0.536$ & $0.607$ & $0.619$ \\
Net utility & $-0.053$ & $0.048$ & $0.114$ & $0.147$ \\
\midrule
\multicolumn{5}{l}{\emph{New SIIM-ACR patients ($200$ cases; $100 + 100$ films)}} \\
ESG & $0.000$ & $0.360$ & $0.270$ & $0.270$ \\
Erased-lesion claims & $0.0\%$ & $19.0\%$ & $25.0\%$ & $42.0\%$ \\
Lesion-free claims & $0.0\%$ & $5.0\%$ & $4.0\%$ & $6.0\%$ \\
Lesion images correct & $0.000$ & $0.540$ & $0.530$ & $0.685$ \\
Net utility & $-1.100$ & $-0.096$ & $-0.134$ & $0.080$ \\
Pneumothorax films called present ($n$ of $100$) & $0$ & $53$ & $54$ & $68$ \\
Films without pneumothorax absent or referred & $100.0\%$ & $95.0\%$ & $96.0\%$ & $94.0\%$ \\
Youden index & $0.000$ & $0.480$ & $0.500$ & $0.620$ \\
\midrule
\multicolumn{5}{l}{\emph{CheXthought test patients ($300 + 300$ films)}} \\
Pneumothorax films called present & $0.0\%$ & $35.7\%$ & $75.0\%$ & $78.7\%$ \\
Films without pneumothorax absent or referred & $100.0\%$ & $93.0\%$ & $91.0\%$ & $91.3\%$ \\
Youden index & $0.000$ & $0.287$ & $0.660$ & $0.700$ \\
\bottomrule
\end{tabular}
\end{adjustbox}
\end{table}

\begin{table}[!htbp]
\caption{Paired differences for Supplementary Table~\ref{tab:s_nmodel}: \msn{} (first seed) minus \ms{} and minus \msp{} on the four counterfactual test sets, paired over cases, with 95\% CI and two-sided $p$ values; differences in claims are differences in proportions. The pre-specified readings were: on the main test, an ESG more than 0.05 below that of \ms{} would count as a loss (not reached); on the fresh external set, an interval above zero for erased-lesion claims against \msp{} would mean that the added training images carry a cost (interval above zero). The other comparisons are descriptive.}
\label{tab:s_next}
\centering
\footnotesize
\begin{adjustbox}{max width=\linewidth}
\begin{tabular}{l cc cc}
\toprule
Measure & \msn{} minus \ms{} [95\% CI] & $p$ & \msn{} minus \msp{} [95\% CI] & $p$ \\
\midrule
\multicolumn{5}{l}{\emph{Main test ($507$ cases)}} \\
ESG & $-0.014$ $\ci{-0.068}{+0.040}$ & $0.680$ & $+0.005$ $\ci{-0.039}{+0.047}$ & $0.928$ \\
Erased-lesion claims & $-0.005$ $\ci{-0.044}{+0.034}$ & $0.906$ & $-0.018$ $\ci{-0.054}{+0.018}$ & $0.414$ \\
Lesion-free claims & $+0.010$ $\ci{-0.022}{+0.042}$ & $0.583$ & $-0.003$ $\ci{-0.030}{+0.025}$ & $0.867$ \\
Lesion images correct & $-0.002$ $\ci{-0.040}{+0.034}$ & $0.963$ & $-0.009$ $\ci{-0.036}{+0.019}$ & $0.584$ \\
Net utility & $-0.023$ $\ci{-0.089}{+0.040}$ & $0.487$ & $-0.005$ $\ci{-0.057}{+0.044}$ & $0.851$ \\
\midrule
\multicolumn{5}{l}{\emph{External datasets ($169$ cases)}} \\
ESG & $+0.076$ $\ci{+0.009}{+0.143}$ & $0.033$ & $-0.067$ $\ci{-0.126}{-0.009}$ & $0.025$ \\
Erased-lesion claims & $-0.008$ $\ci{-0.061}{+0.046}$ & $0.902$ & $+0.042$ $\ci{+0.000}{+0.088}$ & $0.067$ \\
Lesion-free claims & $-0.020$ $\ci{-0.093}{+0.045}$ & $0.758$ & $-0.040$ $\ci{-0.103}{+0.000}$ & $0.281$ \\
Lesion images correct & $+0.050$ $\ci{+0.000}{+0.103}$ & $0.062$ & $-0.038$ $\ci{-0.089}{+0.017}$ & $0.196$ \\
Net utility & $+0.096$ $\ci{+0.011}{+0.182}$ & $0.020$ & $-0.070$ $\ci{-0.149}{+0.014}$ & $0.104$ \\
\midrule
\multicolumn{5}{l}{\emph{Fresh internal ($76$ cases)}} \\
ESG & $-0.056$ $\ci{-0.270}{+0.162}$ & $0.706$ & $-0.111$ $\ci{-0.276}{+0.056}$ & $0.251$ \\
Erased-lesion claims & $+0.056$ $\ci{-0.125}{+0.222}$ & $0.651$ & $+0.000$ $\ci{-0.139}{+0.135}$ & $1.000$ \\
Lesion-free claims & $+0.025$ $\ci{-0.125}{+0.179}$ & $0.862$ & $-0.025$ $\ci{-0.156}{+0.105}$ & $0.876$ \\
Lesion images correct & $+0.000$ $\ci{-0.163}{+0.171}$ & $1.000$ & $-0.097$ $\ci{-0.221}{+0.027}$ & $0.122$ \\
Net utility & $-0.030$ $\ci{-0.277}{+0.218}$ & $0.840$ & $-0.128$ $\ci{-0.314}{+0.042}$ & $0.165$ \\
\midrule
\multicolumn{5}{l}{\emph{Fresh external ($316$ cases)}} \\
ESG & $-0.012$ $\ci{-0.095}{+0.061}$ & $0.752$ & $-0.036$ $\ci{-0.113}{+0.038}$ & $0.335$ \\
Erased-lesion claims & $+0.137$ $\ci{+0.077}{+0.202}$ & $<0.001$ & $+0.107$ $\ci{+0.054}{+0.162}$ & $<0.001$ \\
Lesion-free claims & $+0.007$ $\ci{-0.033}{+0.045}$ & $0.850$ & $-0.014$ $\ci{-0.058}{+0.032}$ & $0.665$ \\
Lesion images correct & $+0.131$ $\ci{+0.069}{+0.197}$ & $<0.001$ & $+0.071$ $\ci{+0.017}{+0.124}$ & $0.011$ \\
Net utility & $+0.168$ $\ci{+0.056}{+0.279}$ & $0.002$ & $+0.067$ $\ci{-0.039}{+0.162}$ & $0.212$ \\
\bottomrule
\end{tabular}
\end{adjustbox}
\end{table}

\begin{table}[!htbp]
\caption{\msn{} (first seed) against the two single-addition models on the four counterfactual test sets: values, paired differences (\msn{} minus the other model) with 95\% CI, and two-sided $p$ values. \msp{} with CheXthought films was not run on the external datasets or the fresh internal set (--). On the fresh external set no difference in erased-lesion claims between \msn{} and either single-addition model was shown; each single-addition model reported more erased lesions there than \msp{} (Supplementary Table~\ref{tab:s_additions}).}
\label{tab:s_next_add}
\centering
\footnotesize
\begin{adjustbox}{max width=\linewidth}
\begin{tabular}{ll c c cc c cc}
\toprule
Test set & Measure & \msn{} & \shortstack{\msp{} with\\colonoscopy pairs} & Difference [95\% CI] & $p$ & \shortstack{\msp{} with\\CheXthought films} & Difference [95\% CI] & $p$ \\
\midrule
Main test ($507$ cases) & ESG & $0.744$ & $0.744$ & $+0.000$ $\ci{-0.038}{+0.038}$ & $1.000$ & $0.753$ & $-0.009$ $\ci{-0.054}{+0.035}$ & $0.731$ \\
 & Erased-lesion claims & $9.1\%$ & $11.0\%$ & $-0.018$ $\ci{-0.052}{+0.014}$ & $0.356$ & $7.3\%$ & $+0.018$ $\ci{-0.009}{+0.049}$ & $0.255$ \\
 & Lesion-free claims & $8.7\%$ & $8.7\%$ & $+0.000$ $\ci{-0.029}{+0.028}$ & $1.000$ & $7.6\%$ & $+0.010$ $\ci{-0.016}{+0.038}$ & $0.537$ \\
 & Lesion images correct & $0.847$ & $0.854$ & $-0.007$ $\ci{-0.032}{+0.017}$ & $0.645$ & $0.840$ & $+0.007$ $\ci{-0.020}{+0.035}$ & $0.683$ \\
 & Net utility & $0.709$ & $0.707$ & $+0.002$ $\ci{-0.040}{+0.046}$ & $0.927$ & $0.725$ & $-0.016$ $\ci{-0.066}{+0.033}$ & $0.525$ \\
\midrule
External datasets ($169$ cases) & ESG & $0.672$ & $0.681$ & $-0.008$ $\ci{-0.073}{+0.052}$ & $0.828$ & -- & -- & -- \\
 & Erased-lesion claims & $10.9\%$ & $11.8\%$ & $-0.008$ $\ci{-0.036}{+0.018}$ & $0.798$ & -- & -- & -- \\
 & Lesion-free claims & $2.0\%$ & $2.0\%$ & $+0.000$ $\ci{-0.057}{+0.057}$ & $1.000$ & -- & -- & -- \\
 & Lesion images correct & $0.777$ & $0.798$ & $-0.021$ $\ci{-0.069}{+0.027}$ & $0.426$ & -- & -- & -- \\
 & Net utility & $0.625$ & $0.645$ & $-0.020$ $\ci{-0.094}{+0.052}$ & $0.588$ & -- & -- & -- \\
\midrule
Fresh internal ($76$ cases) & ESG & $0.417$ & $0.500$ & $-0.083$ $\ci{-0.282}{+0.105}$ & $0.478$ & -- & -- & -- \\
 & Erased-lesion claims & $25.0\%$ & $33.3\%$ & $-0.083$ $\ci{-0.229}{+0.059}$ & $0.342$ & -- & -- & -- \\
 & Lesion-free claims & $15.0\%$ & $20.0\%$ & $-0.050$ $\ci{-0.176}{+0.073}$ & $0.553$ & -- & -- & -- \\
 & Lesion images correct & $0.681$ & $0.819$ & $-0.139$ $\ci{-0.250}{-0.038}$ & $0.007$ & -- & -- & -- \\
 & Net utility & $0.372$ & $0.484$ & $-0.111$ $\ci{-0.289}{+0.072}$ & $0.240$ & -- & -- & -- \\
\midrule
Fresh external ($316$ cases) & ESG & $0.381$ & $0.387$ & $-0.006$ $\ci{-0.084}{+0.067}$ & $0.886$ & $0.381$ & $-0.000$ $\ci{-0.077}{+0.071}$ & $1.000$ \\
 & Erased-lesion claims & $22.0\%$ & $19.6\%$ & $+0.024$ $\ci{-0.031}{+0.078}$ & $0.484$ & $20.2\%$ & $+0.018$ $\ci{-0.037}{+0.075}$ & $0.571$ \\
 & Lesion-free claims & $5.4\%$ & $4.1\%$ & $+0.014$ $\ci{-0.013}{+0.042}$ & $0.458$ & $5.4\%$ & $+0.000$ $\ci{-0.037}{+0.036}$ & $1.000$ \\
 & Lesion images correct & $0.607$ & $0.577$ & $+0.030$ $\ci{-0.022}{+0.078}$ & $0.294$ & $0.583$ & $+0.024$ $\ci{-0.022}{+0.069}$ & $0.345$ \\
 & Net utility & $0.114$ & $0.088$ & $+0.026$ $\ci{-0.068}{+0.118}$ & $0.609$ & $0.127$ & $-0.013$ $\ci{-0.099}{+0.065}$ & $0.754$ \\
\bottomrule
\end{tabular}
\end{adjustbox}
\end{table}

\begin{table}[!htbp]
\caption{Results by image source. Top: main test (507 cases) by imaging type for \msn{} and \ms{}. Bottom: fresh external set (316 cases) by source for the five trained models. Point estimates to two decimals; erased-lesion and lesion-free claims are proportions. ETIS-Larib has no lesion-free images (--).}
\label{tab:s_source}
\centering
\footnotesize
\begin{adjustbox}{max width=\linewidth}
\begin{tabular}{ll c ccccc}
\toprule
Source (cases) & Model & & Lesion images correct & ESG & Erased-lesion claims & Lesion-free claims & Net utility \\
\midrule
\multicolumn{8}{l}{Main test} \\
Brain MRI, TCGA-LGG ($78$) & \msn{} & & $0.82$ & $0.69$ & $0.11$ & $0.10$ & $0.57$ \\
 & \ms{} & & $0.88$ & $0.75$ & $0.14$ & $0.00$ & $0.75$ \\
Breast ultrasound, BUSI ($57$) & \msn{} & & $0.92$ & $0.76$ & $0.14$ & $0.13$ & $0.84$ \\
 & \ms{} & & $0.88$ & $0.74$ & $0.14$ & $0.07$ & $0.78$ \\
Chest X-ray, ChestX-Det ($91$) & \msn{} & & $0.70$ & $0.51$ & $0.15$ & $0.16$ & $0.42$ \\
 & \ms{} & & $0.70$ & $0.51$ & $0.17$ & $0.20$ & $0.40$ \\
Colonoscopy, Kvasir-SEG and HyperKvasir ($174$) & \msn{} & & $0.98$ & $0.94$ & $0.04$ & $0.01$ & $1.05$ \\
 & \ms{} & & $0.96$ & $0.95$ & $0.01$ & $0.00$ & $1.06$ \\
SLAKE CT and MRI ($107$) & \msn{} & & $0.47$ & $0.41$ & $0.06$ & $0.11$ & $0.20$ \\
 & \ms{} & & $0.53$ & $0.47$ & $0.12$ & $0.12$ & $0.22$ \\
\midrule
\multicolumn{8}{l}{Fresh external set} \\
Brain MRI, BraTS 2023 meningioma ($82$) & \msn{} & & $0.44$ & $0.31$ & $0.12$ & $0.06$ & $-0.19$ \\
 & \msp{} & & $0.48$ & $0.31$ & $0.12$ & $0.18$ & $-0.17$ \\
 & \msp{} with colonoscopy pairs & & $0.42$ & $0.22$ & $0.16$ & $0.06$ & $-0.20$ \\
 & \msp{} with CheXthought films & & $0.48$ & $0.31$ & $0.16$ & $0.06$ & $-0.09$ \\
 & \ms{} & & $0.53$ & $0.41$ & $0.16$ & $0.08$ & $-0.03$ \\
Breast ultrasound, BUS-UCLM ($99$) & \msn{} & & $0.72$ & $0.59$ & $0.14$ & $0.10$ & $0.56$ \\
 & \msp{} & & $0.71$ & $0.63$ & $0.08$ & $0.02$ & $0.63$ \\
 & \msp{} with colonoscopy pairs & & $0.73$ & $0.67$ & $0.08$ & $0.06$ & $0.64$ \\
 & \msp{} with CheXthought films & & $0.64$ & $0.59$ & $0.06$ & $0.06$ & $0.52$ \\
 & \ms{} & & $0.73$ & $0.71$ & $0.02$ & $0.06$ & $0.67$ \\
Chest X-ray, SIIM-ACR pneumothorax ($96$) & \msn{} & & $0.42$ & $0.06$ & $0.33$ & $0.00$ & $-0.42$ \\
 & \msp{} & & $0.22$ & $0.12$ & $0.10$ & $0.00$ & $-0.70$ \\
 & \msp{} with colonoscopy pairs & & $0.36$ & $0.04$ & $0.35$ & $0.00$ & $-0.56$ \\
 & \msp{} with CheXthought films & & $0.51$ & $0.04$ & $0.46$ & $0.04$ & $-0.33$ \\
 & \ms{} & & $0.00$ & $0.00$ & $0.00$ & $0.00$ & $-1.10$ \\
Colonoscopy, ETIS-Larib ($39$) & \msn{} & & $0.83$ & $0.56$ & $0.26$ & -- & $0.58$ \\
 & \msp{} & & $0.74$ & $0.59$ & $0.15$ & -- & $0.53$ \\
 & \msp{} with colonoscopy pairs & & $0.77$ & $0.59$ & $0.18$ & -- & $0.55$ \\
 & \msp{} with CheXthought films & & $0.68$ & $0.59$ & $0.10$ & -- & $0.46$ \\
 & \ms{} & & $0.69$ & $0.46$ & $0.21$ & -- & $0.38$ \\
\bottomrule
\end{tabular}
\end{adjustbox}
\end{table}

\begin{table}[!htbp]
\caption{Accuracy on three standard visual question answering test sets, by question type (closed: yes or no; open: free text). Point estimates; no interval was computed. The training splits of the three benchmarks were used in the last training stage of each of the four trained models; \base{} is the base model without fine-tuning. For \gem{} only the accuracy over all questions was recorded. The share of requests for a better image by type of image degradation, planned for the four counterfactual test sets, is not reported, nor is standard question answering by \opus{} and \gpt{} or by the open-weight models with their own prompt, which was planned on 2 November 2026 and has no recorded result.}
\label{tab:s_vqa}
\centering
\footnotesize
\begin{adjustbox}{max width=\linewidth}
\begin{tabular}{l ccc ccc ccc}
\toprule
 & \multicolumn{3}{c}{PathVQA ($300$ questions)} & \multicolumn{3}{c}{SLAKE ($300$ questions)} & \multicolumn{3}{c}{VQA-RAD ($451$ questions)} \\
\cmidrule(lr){2-4}\cmidrule(lr){5-7}\cmidrule(lr){8-10}
Model & Closed & Open & All & Closed & Open & All & Closed & Open & All \\
\midrule
\msn{} & $0.859$ & $0.125$ & $0.507$ & $0.881$ & $0.665$ & $0.743$ & $0.757$ & $0.250$ & $0.532$ \\
\msp{} with colonoscopy pairs & $0.840$ & $0.139$ & $0.503$ & $0.881$ & $0.670$ & $0.747$ & $0.725$ & $0.255$ & $0.517$ \\
\msp{} & $0.833$ & $0.111$ & $0.487$ & $0.872$ & $0.681$ & $0.750$ & $0.773$ & $0.245$ & $0.539$ \\
\ms{} & $0.801$ & $0.125$ & $0.477$ & $0.881$ & $0.644$ & $0.730$ & $0.741$ & $0.230$ & $0.514$ \\
\base{} & $0.545$ & $0.021$ & $0.293$ & $0.826$ & $0.424$ & $0.570$ & $0.657$ & $0.260$ & $0.481$ \\
\gem{} & -- & -- & $0.417$ & -- & -- & $0.780$ & -- & -- & $0.674$ \\
\bottomrule
\end{tabular}
\end{adjustbox}
\end{table}

\clearpage
\subsection{Counterfactual views, unedited pairs and alternative designs}
\label{subsec:s_designs}

\begin{figure}[!htbp]
\centering
\includegraphics[width=\textwidth]{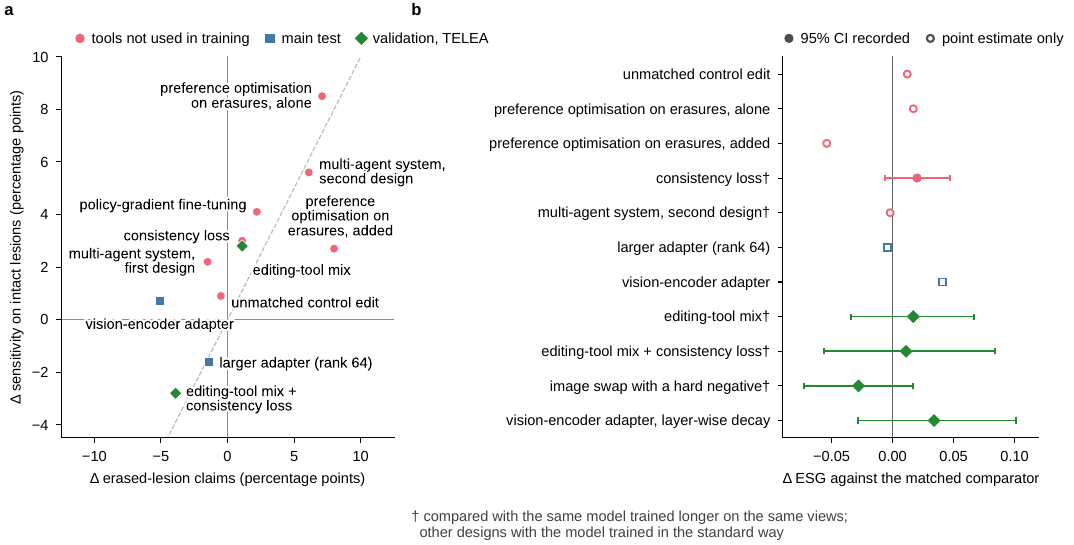}
\caption{Alternative designs against their comparators (values in Supplementary Tables~\ref{tab:s_designs} to~\ref{tab:s_designs_later}). (a) Each trained design or agent design against its comparator: change in erased-lesion claims (horizontal axis) and in the share of intact lesions answered correctly (vertical axis), in percentage points; the plotted changes are differences between values in these tables. Tools not used in training are TELEA and Stable Diffusion, pooled (219 cases with TELEA, 191 with Stable Diffusion); the two designs with a mix of editing tools are on validation TELEA views; the adapter designs are on the main test (507 cases). (b) Change in ESG for every design with a recorded ESG difference (filled symbols: 95\% bootstrap interval over cases recorded; open symbols: point estimate only). The point estimates lie between $-0.054$ and $+0.041$, and every recorded interval includes zero. The design at $-0.054$, preference optimisation added after counterfactual training, has no recorded interval; its decrease had $p = 0.027$. Reinforcement learning and the first multi-agent system have no recorded ESG difference on these sets and appear in panel (a) only. A dagger marks designs compared with the same model trained longer on the same views. The last two rows of panel (b) are the image swap with a hard negative and the vision adapter with layer-wise decay of Supplementary Table~\ref{tab:s_designs_later}. In the figure, ``sensitivity on intact lesions'' is the share of intact lesions answered correctly.}
\label{fig:s_designs}
\end{figure}

Supplementary Table~\ref{tab:s_twobytwo} gives the two-by-two comparison of training images. The seventeen alternative designs of Section~\ref{subsec:res_designs} are the twelve of a first series and five tested later. Supplementary Table~\ref{tab:s_designs} lists the twelve and the 270 decision rules, Supplementary Tables~\ref{tab:s_designs_obj}, \ref{tab:s_designs_cons} and~\ref{tab:s_designs_tools} give their complete results, Supplementary Table~\ref{tab:s_designs_later} three designs tested later, Supplementary Table~\ref{tab:s_inspect} the other two, which rewarded inspection, and Supplementary Fig.~\ref{fig:s_designs} the changes against each comparator.

\paragraph{The designs in words} Preference optimisation is direct preference optimisation \citep{rafailov2023dpo} on pairs of a preferred and a rejected answer built from lesion removal. The consistency loss asks for the same decision score on an original view and its control edit and for a margin between the control edit and the erasure. Reinforcement learning is group relative policy optimisation \citep{shao2024deepseekmath} on the cost-matrix reward. The multi-agent system combines the trained model, a copy that does not receive the question, a lesion segmenter, a verifier that erases the reported region, makes an equal edit elsewhere and asks again, and a deciding agent that reads their reports; in the first design the deciding agent was trained on reports for training images, in the second on reports for validation images. Of the 270 decision rules, 128 combine the outputs of these agents by thresholds, regularised logistic regression, boosted trees and referral bands. The 142 rules with test-time evidence add three altered copies of the image (mirror, centre crop, contrast), a close-up of the reported region with an equal-area control close-up, a repeated question with a cautious prompt, the probability of the first answer token, a second model (Qwen2.5-VL-7B) and the segmenter score. Among the 128 rules that combine agent outputs, the largest gain in net utility on the second half of the validation set came with erased-lesion claims of $3.9\%$ against $11.2\%$ and lesion images correct of $0.727$ against $0.773$.

\paragraph{Continued training as a comparator (post hoc)} The comparator of the consistency loss, continued training on the same views with no new objective, itself reported fewer erased lesions with tools not used in training than the first-stage model in two of three seeds ($-0.080$ and $-0.085$, $p < 0.001$; $-0.012$, interval includes zero) and had a higher ESG with these tools in one ($+0.076$, $p = 0.001$; $+0.022$ and $+0.027$, intervals include zero), with fewer unedited lesion images correct at one seed ($-0.049$, $p < 0.001$). This comparison was not planned in advance.

\begin{table}[!htbp]
\caption{Counterfactual views and unedited pairs in training: the two-by-two comparison (pre-specified). Four models received the first training stage only (one seed; no pneumothorax films, no replay of general question answering): with or without counterfactual views, then with or without a continuation on unedited REAL-Colon pairs. The two models with unedited pairs were trained three times, each without one third of the videos, and each of the 103 pairs was answered by the copy that had not used its video. Top: the 103 REAL-Colon pairs; columns as in Supplementary Table~\ref{tab:s_realcolon}; brackets are 95\% intervals from $2{,}000$ bootstrap resamples of whole videos. The pre-specified primary comparison is both kinds minus unedited pairs only: $-0.058$ $\ci{-0.168}{+0.042}$, $p = 0.311$ (not shown; no claim of equivalence). Descriptive: unedited pairs only minus neither $+0.350$ $\ci{+0.227}{+0.474}$ and both kinds minus counterfactual views only $+0.282$ $\ci{+0.156}{+0.405}$ (both $p < 0.001$); both kinds minus \gem{} $+0.039$ ($p = 0.635$), minus \opus{} $+0.058$ ($p = 0.409$) and minus \gpt{} $+0.184$ $\ci{+0.047}{+0.321}$ ($p = 0.014$). Bottom: main test (507 cases) for the copies trained without the first third of the videos, paired over cases.}
\label{tab:s_twobytwo}
\centering
\footnotesize
\begin{adjustbox}{max width=\linewidth}
\begin{tabular}{l cc c cc ccc}
\toprule
 & & & & & & \multicolumn{3}{c}{Natural ESG by fold} \\
\cmidrule(lr){7-9}
Training images & \shortstack{Polyp frames\\found} & \shortstack{Polyp-free frames\\absent or referred} & \shortstack{Natural ESG\\{}[95\% CI]} & \shortstack{Correct box,\\$n$ of $103$} & \shortstack{Polyp reported on\\polyp-free frames, $n$ of $103$} & 0 & 1 & 2 \\
\midrule
Counterfactual views and unedited pairs & $62.1\%$ & $88.3\%$ & $0.505$ $\ci{0.369}{0.629}$ & $48$ & $12$ & $0.63$ & $0.40$ & $0.47$ \\
Unedited pairs only & $68.9\%$ & $87.4\%$ & $0.563$ $\ci{0.444}{0.676}$ & $57$ & $13$ & $0.58$ & $0.51$ & $0.60$ \\
Counterfactual views only & $35.9\%$ & $86.4\%$ & $0.223$ $\ci{0.133}{0.309}$ & $33$ & $3$ & -- & -- & -- \\
Neither & $52.4\%$ & $68.9\%$ & $0.214$ $\ci{0.079}{0.343}$ & $43$ & $27$ & -- & -- & -- \\
\midrule
\gem{} & $68.0\%$ & $78.6\%$ & $0.466$ & $52$ & n.r. & -- & -- & -- \\
\opus{} & $70.9\%$ & $73.8\%$ & $0.447$ & $50$ & $20$ & -- & -- & -- \\
\gpt{} & $64.1\%$ & $68.0\%$ & $0.320$ & $41$ & n.r. & -- & -- & -- \\
\bottomrule
\end{tabular}
\end{adjustbox}

\vspace{0.6em}
\begin{adjustbox}{max width=\linewidth}
\begin{tabular}{l cccc}
\toprule
Main test, training images & ESG & Erased-lesion claims & Lesion-free claims & Net utility \\
\midrule
Counterfactual views and unedited pairs & $0.744$ & $13.7\%$ & $16.3\%$ & $0.684$ \\
Unedited pairs only & $0.260$ & $68.0\%$ & n.r. & $0.343$ \\
Counterfactual views only & $0.735$ & $11.9\%$ & $9.4\%$ & $0.692$ \\
\midrule
Both kinds minus unedited pairs only & $+0.484$ $\ci{+0.408}{+0.558}$ & n.r. & n.r. & n.r. \\
Both kinds minus counterfactual views only & $+0.009$ ($p = 0.777$) & n.r. & $+0.069$ $\ci{+0.030}{+0.111}$ ($p = 0.001$) & n.r. \\
\bottomrule
\end{tabular}
\end{adjustbox}
\end{table}

\begin{table}[!htbp]
\caption{Twelve alternative designs and 270 decision rules, each tested against a criterion fixed in advance (the criteria and dates are in Supplementary Table~\ref{tab:s_plan}). Each row gives what the design changes, its comparator, the result and whether the criterion was met. Tools not used in training: erasures and control edits remade with TELEA and Stable Diffusion, pooled (219 cases with TELEA, 191 with Stable Diffusion). Comparator ``first-stage model'': counterfactual training, first stage, one seed. Comparator ``continued training'': the first-stage model trained further on the same views for the same number of steps, without the change. Decision rules were screened on two halves of the validation set; a rule passed if, against the model alone, net utility rose by at least 0.03, ESG rose by at least 0.02, lesion images correct fell by no more than 0.03 and lesion-free claims rose by no more than 0.03 on the first half, and the same held on the second half. $^{\ast}$Comparisons with tools not used in training against continued training; lesion images correct against the first-stage model.}
\label{tab:s_designs}
\centering
\footnotesize
\begin{adjustbox}{max width=\linewidth}
\begin{tabular}{L{4.6cm}L{2.6cm}L{8.2cm}L{1.9cm}}
\toprule
What the design changes & Comparator & Result & Criterion \\
\midrule
\multicolumn{4}{l}{\emph{Control edit and training objective}} \\
Control edit matched to the erasure in shape and editing tool & Unmatched control edit & Main-test ESG $0.744$ against $0.753$; ESG with tools not used in training $0.749$ against $0.761$ (within the variation between seeds; the unmatched design shows a response bias on SIIM-ACR films) & Inconclusive \\
Counterfactual training itself & No counterfactual views, then preference optimisation (three seeds) & Fresh external ESG $0.381 \pm 0.018$ against $0.355 \pm 0.015$ (preference optimisation has lower lesion images correct, $0.485$ against $0.587$, and fewer erased-lesion claims) & No difference shown \\
Preference optimisation added after counterfactual training & First-stage model & ESG with tools not used in training $0.649 \to 0.595$; erased-lesion claims $16.3\% \to 24.4\%$ ($+0.080$ $\ci{0.046}{0.118}$) & Not met \\
Consistency loss between an original view and its control edit & Continued training (three seeds) & Pooled ESG with tools not used in training $+0.020$ ($p = 0.143$); all endpoints move towards ``present'' (fresh external erased-lesion claims $+0.073$, $p < 0.001$) & Not met \\
\midrule
\multicolumn{4}{l}{\emph{Adapter, reinforcement learning and agent design}} \\
Adapter rank 64 instead of 16 & First-stage model & Main-test ESG $0.731$ against $0.735$; calls 25 of 48 SIIM-ACR pneumothorax films present, and also 21 of 48 erased films and 12 of 48 films without pneumothorax (SIIM-ACR ESG $0.042$) & Not met \\
Adapter on the vision encoder, learning rate $2 \times 10^{-4}$ & First-stage model & ESG $0.000$ on every set; the answer depends on the question only & Not met \\
Adapter on the vision encoder, learning rate $2 \times 10^{-5}$ & First-stage model & Main-test ESG $0.776$, but fresh external lesion images correct $0.560 \to 0.488$ ($-0.071$, $p = 0.03$); erased-lesion claims $6.8\%$ & Not met \\
Reinforcement learning on the cost-matrix reward (150 steps) & Continued training & Net utility with tools not used in training $+0.056$, erased-lesion claims $+0.022$ (the condition on erased-lesion claims was missed); fresh external net utility $+0.124$, ESG $+0.083$, lesion-free claims $+0.054$ & Not met \\
Multi-agent system, first design & Continued training$^{\ast}$ & Net utility with tools not used in training $+0.090$ $\ci{+0.004}{+0.180}$; lesion images correct $0.769 \to 0.513$ on the external datasets and $0.560 \to 0.378$ on the fresh external set & Not met \\
Multi-agent system, second design (deciding agent trained on validation views) & Continued training$^{\ast}$ & Main-test net utility $0.809$ (its verifier uses the editing tool of the main test), but ESG with tools not used in training $-0.002$ and erased-lesion claims $+0.061$ $\ci{+0.027}{+0.096}$ & Not met \\
\midrule
\multicolumn{4}{l}{\emph{Mix of editing tools in first-stage training}} \\
Four editing tools & Continued training & Validation TELEA ESG $0.665$ against $0.648$ ($+0.017$ $\ci{-0.034}{0.067}$); LaMa ESG $-0.022$; TELEA control edit denied $-0.021$ & Not met \\
Four editing tools with the consistency loss & Continued training & Validation TELEA ESG $0.659$ against $0.648$ ($+0.011$); lesion-free claims $+0.075$ $\ci{0.013}{0.150}$; edit-site control denied $+0.061$ & Not met \\
\midrule
\multicolumn{4}{l}{\emph{Decision rules on the outputs of the first-stage model}} \\
128 rules that recombine the agents' outputs & Two halves of the validation set & 19 of 128 pass on the first half, 0 of 19 on the second (net utility $0.539 \to 0.706$ through referral; change in ESG at most $+0.026$) & None passed \\
142 rules that add test-time evidence & Two halves of the validation set & 0 of 142 pass on the first half (TELEA net utility $0.444 \to 0.686$ through abstention; ESG mostly $0.55$ to $0.61$, at most $0.642$, against $0.600$ for the model alone) & None passed \\
\bottomrule
\end{tabular}
\end{adjustbox}
\end{table}

\begin{table}[!htbp]
\caption{Control edit and training objective: complete results for the first four rows of Supplementary Table~\ref{tab:s_designs} (first training stage, first seed). Control edit denied: share of intact lesions answered ``absent'' or referred after the control edit. Tools not used in training: TELEA and Stable Diffusion pooled (219 cases with TELEA, 191 with Stable Diffusion). Point estimates.}
\label{tab:s_designs_obj}
\centering
\footnotesize
\begin{adjustbox}{max width=\linewidth}
\begin{tabular}{llccccc}
\toprule
Test set (cases) & Training & Lesion images correct & ESG & Erased-lesion claims (\%) & Control edit denied (\%) & Net utility \\
\midrule
Main test ($507$) & Erasures with matched control edits & $0.897$ & $0.744$ & $15.1$ & $10.5$ & $0.727$ \\
 & Erasures with unmatched control edits & $0.904$ & $0.753$ & $15.1$ & $9.1$ & $0.739$ \\
 & Counterfactual training, first stage & $0.847$ & $0.735$ & $11.9$ & $14.6$ & $0.692$ \\
 & No counterfactual views, then preference optimisation & $0.900$ & $0.658$ & $23.7$ & $10.0$ & $0.602$ \\
 & Counterfactual training, then preference optimisation & $0.872$ & $0.753$ & $11.4$ & $13.2$ & $0.700$ \\
\midrule
Tools not used in training ($219$) & Erasures with matched control edits & $0.859$ & $0.749$ & $11.0$ & $13.7$ & $0.733$ \\
 & Erasures with unmatched control edits & $0.868$ & $0.761$ & $10.5$ & $12.9$ & $0.772$ \\
 & Counterfactual training, first stage & $0.810$ & $0.649$ & $16.3$ & $18.8$ & $0.566$ \\
 & No counterfactual views, then preference optimisation & $0.895$ & $0.666$ & $23.4$ & $10.0$ & $0.644$ \\
 & Counterfactual training, then preference optimisation & $0.837$ & $0.595$ & $24.4$ & $16.1$ & $0.502$ \\
\midrule
Fresh external ($316$) & Erasures with matched control edits & $0.488$ & $0.405$ & $8.3$ & $51.2$ & $-0.083$ \\
 & Erasures with unmatched control edits & $0.670$ & $0.411$ & $27.4$ & $31.5$ & $0.165$ \\
 & Counterfactual training, first stage & $0.560$ & $0.381$ & $16.7$ & $45.2$ & $0.061$ \\
 & No counterfactual views, then preference optimisation & $0.461$ & $0.369$ & $8.9$ & $54.2$ & $-0.087$ \\
 & Counterfactual training, then preference optimisation & $0.542$ & $0.381$ & $15.5$ & $46.4$ & $0.057$ \\
\bottomrule
\end{tabular}
\end{adjustbox}
\end{table}

\begin{table}[!htbp]
\caption{Consistency loss between an original view and its control edit, against continued training without the loss: three pairs of seeds with tools not used in training (TELEA and Stable Diffusion pooled; 219 cases with TELEA, 191 with Stable Diffusion), with the pre-specified check on the main test and one endpoint on the fresh external set. The pre-specified primary endpoint was erased-lesion claims with tools not used in training (not met). The pre-specified check on the fresh external set concerned ESG and net utility, which did not fall ($+0.058$ and $+0.146$); the last row shows erased-lesion claims there, which rose. Differences are paired by case with 95\% bootstrap intervals.}
\label{tab:s_designs_cons}
\centering
\footnotesize
\begin{adjustbox}{max width=\linewidth}
\begin{tabular}{llcccc}
\toprule
Seed and test set & Endpoint & Consistency loss & Continued training & Difference [95\% CI] & $p$ \\
\midrule
First seed, tools not used in training & Erased-lesion claims & $0.122$ & $0.151$ & $-0.029$ $\ci{-0.062}{+0.002}$ & $0.085$ \\
Second seed, tools not used in training & Erased-lesion claims & $0.137$ & $0.078$ & $+0.059$ $\ci{+0.031}{+0.089}$ & $<0.001$ \\
Third seed, tools not used in training & Erased-lesion claims & $0.120$ & $0.115$ & $+0.005$ $\ci{-0.029}{+0.040}$ & $0.818$ \\
\midrule
Pooled, tools not used in training & Erased-lesion claims & $0.126$ & $0.115$ & $+0.011$ $\ci{-0.009}{+0.030}$ & $0.265$ \\
Pooled, tools not used in training & ESG & $0.729$ & $0.709$ & $+0.020$ $\ci{-0.006}{+0.047}$ & $0.143$ \\
Pooled, tools not used in training & Control edit denied & $0.143$ & $0.176$ & $-0.033$ $\ci{-0.054}{-0.012}$ & $0.001$ \\
Pooled, tools not used in training & Lesion images correct & $0.854$ & $0.824$ & $+0.030$ $\ci{+0.010}{+0.051}$ & $0.004$ \\
Pooled, tools not used in training & Net utility & $0.712$ & $0.661$ & $+0.051$ $\ci{+0.002}{+0.100}$ & $0.042$ \\
\midrule
Main test (pre-specified check) & Lesion-free claims & $0.131$ & $0.079$ & $+0.052$ $\ci{+0.032}{+0.075}$ & $<0.001$ \\
Fresh external & Erased-lesion claims & $0.141$ & $0.067$ & $+0.073$ $\ci{+0.045}{+0.106}$ & $<0.001$ \\
\bottomrule
\end{tabular}
\end{adjustbox}
\end{table}

\begin{table}[!htbp]
\caption{Four editing tools in first-stage training, on validation images (179 lesion cases and 80 lesion-free images; one seed). Each training case's erasure and control edit were made with one tool drawn from LaMa, blur fill, mean fill with noise and shuffled pixels; TELEA was not used in training. The designs continue the first-stage model on the $2{,}739$ training cases of the comparator (continued training on the same views without the change). Entries are values with, in parentheses, the difference from the comparator paired by case and its 95\% bootstrap interval where recorded. The criterion required a rise in TELEA ESG of at least 0.03 and a fall in TELEA control edit denied of at least 0.03, with LaMa ESG not lower by more than 0.02, unedited lesions found not lower by more than 0.03 and lesion-free claims not higher by more than 0.03. The comparator's value in the last row differs between the two designs ($0.191$ and $0.153$) because each comparison is restricted to cases that both models of the pair found on the unedited image.}
\label{tab:s_designs_tools}
\centering
\footnotesize
\begin{adjustbox}{max width=\linewidth}
\begin{tabular}{lccc}
\toprule
Validation quantity & Four editing tools & Four editing tools with the consistency loss & Continued training \\
\midrule
Unedited lesions found & $0.821$ ($+0.028$ $\ci{0.000}{0.056}$) & $0.765$ ($-0.028$ $\ci{-0.073}{0.017}$) & $0.793$ \\
Lesion-free claims & $0.025$ ($+0.013$) & $0.087$ ($+0.075$ $\ci{0.013}{0.150}$) & $0.013$ \\
ESG, LaMa (used in training) & $0.637$ ($-0.022$ $\ci{-0.073}{0.028}$) & $0.654$ ($-0.006$) & $0.659$ \\
ESG, TELEA (not used in training) & $0.665$ ($+0.017$ $\ci{-0.034}{0.067}$) & $0.659$ ($+0.011$ $\ci{-0.056}{0.084}$) & $0.648$ \\
Erasure answered ``absent'' or referred, LaMa & $0.821$ ($-0.050$ $\ci{-0.095}{-0.006}$) & $0.899$ ($+0.028$) & $0.872$ \\
Erasure answered ``absent'' or referred, TELEA & $0.899$ ($-0.011$) & $0.950$ ($+0.039$ $\ci{0.000}{0.078}$) & $0.911$ \\
Control edit denied (lesion found unedited), TELEA & $0.078$ ($-0.021$ $\ci{-0.057}{0.007}$) & $0.107$ ($+0.008$) & $0.099$ \\
Edit-site control denied (lesion found unedited), TELEA & $0.163$ ($-0.028$ $\ci{-0.064}{0.000}$) & $0.214$ ($+0.061$ $\ci{0.023}{0.107}$) & $0.191$ / $0.153$ \\
\bottomrule
\end{tabular}
\end{adjustbox}
\end{table}

\begin{table}[!htbp]
\caption{Three of the five designs tested later, each against a criterion fixed in advance (not among the twelve of Supplementary Table~\ref{tab:s_designs}; the other two are in Supplementary Table~\ref{tab:s_inspect}). Top: two training designs on validation images (179 lesion cases, 80 lesion-free images; one seed), as differences from the comparator paired by case with 95\% bootstrap intervals. The image swap adds a training term that asks the answer to change when the image is replaced by the most similar lesion-free training image of another patient \citep{zafar2026coral}; its criterion was a rise in TELEA ESG of 0.03. The vision adapter is a low-rank adapter on the vision encoder with a learning rate that decays from the top layer down; its criterion was a rise of 0.03 in unedited lesions found. Bottom: a decision rule that subtracts from the model's decision score the score it gives to a uniform grey image with the same question \citep{huang2026medcap}, with a fixed or an adaptive weight; weights and threshold were chosen on half of the validation cases with LaMa views, and results are on the other half (118 cases). Its criterion was a rise of 0.03, with the interval above zero, in the area under the curve for separating TELEA erasures from edit-site controls (a ring-shaped edit next to the intact lesion).}
\label{tab:s_designs_later}
\centering
\footnotesize
\begin{adjustbox}{max width=\linewidth}
\begin{tabular}{lcc}
\toprule
Quantity & \shortstack{Image swap with a hard negative\\against continued training} & \shortstack{Vision adapter with layer-wise decay\\against first-stage model} \\
\midrule
Unedited lesions found & $+0.017$ $\ci{-0.011}{0.045}$ & $+0.017$ $\ci{-0.034}{0.067}$ \\
Lesion-free claims & $+0.013$ $\ci{0.000}{0.037}$ & $-0.075$ $\ci{-0.150}{0.000}$ \\
ESG, LaMa & $+0.011$ $\ci{-0.022}{0.050}$ & $+0.061$ $\ci{-0.011}{0.134}$ \\
ESG, TELEA (not used in training) & $-0.028$ $\ci{-0.073}{0.017}$ & $+0.034$ $\ci{-0.028}{0.101}$ \\
Control edit denied (lesion found unedited), TELEA & $+0.029$ $\ci{0.000}{0.064}$ & $0.000$ $\ci{-0.053}{0.053}$ \\
Edit-site control denied (lesion found unedited), TELEA & $-0.014$ $\ci{-0.036}{0.000}$ & $+0.075$ $\ci{0.030}{0.128}$ \\
Criterion & Not met & Not met \\
\bottomrule
\end{tabular}
\end{adjustbox}

\vspace{0.6em}
\begin{adjustbox}{max width=\linewidth}
\begin{tabular}{lcc}
\toprule
Subtraction of the score obtained without the image ($118$ cases) & Fixed weight & Adaptive weight \\
\midrule
Change in area under the curve, TELEA erasure against edit-site control (unchanged model: $0.731$) & $+0.003$ $\ci{-0.018}{0.024}$ & $-0.003$ $\ci{-0.022}{0.019}$ \\
Criterion & Not met & Not met \\
\bottomrule
\end{tabular}
\end{adjustbox}
\end{table}

\begin{table}[!htbp]
\caption{Rewarding inspection before the answer: the other two designs tested later, with one seed each (criterion fixed in advance). 38 REAL-Colon pairs of the videos in the first fold, which no model in the table had used. Inspection-first training: the last training stage repeated with training examples in which the model zooms on the lesion site, or on a region of a lesion-free image, before it answers. Reward for inspection: 150 steps of reinforcement learning on the cost-matrix reward plus a reward for a zoom that covers the lesion or the erased site. Its control is the same reinforcement learning without the added reward. The comparison model is \msp{} with colonoscopy pairs trained without the first fold. The last two columns are on validation images. The criterion required, against the comparison model, a natural ESG higher by at least 0.05, at least three more polyps called present with a correct box, validation net utility not lower by more than 0.02 and validation lesion-free claims not higher by more than 0.02; the reward design also had to exceed its control on natural ESG. Neither design met it. Point estimates; no interval was computed for these 38 pairs.}
\label{tab:s_inspect}
\centering
\footnotesize
\begin{adjustbox}{max width=\linewidth}
\begin{tabular}{l c cc cc cc}
\toprule
Model & Natural ESG & \shortstack{Polyp frames\\found} & \shortstack{Polyp-free frames\\absent or referred} & \shortstack{Correct box,\\$n$ of $38$} & \shortstack{Polyp frames\\inspected, $n$ of $38$} & \shortstack{Validation\\net utility} & \shortstack{Validation\\lesion-free claims} \\
\midrule
Comparison model & $0.632$ & $71.1\%$ & $92.1\%$ & $21$ & $27$ & $0.626$ & $11.4\%$ \\
Inspection-first training & $0.500$ & $78.9\%$ & $71.1\%$ & $21$ & $38$ & $0.560$ & $25.5\%$ \\
Reinforcement learning with a reward for inspection & $0.263$ & $84.2\%$ & $42.1\%$ & $20$ & $38$ & $0.564$ & $10.0\%$ \\
Reinforcement learning, usual reward (control) & $0.579$ & $63.2\%$ & $94.7\%$ & $20$ & $24$ & $0.630$ & $7.3\%$ \\
\bottomrule
\end{tabular}
\end{adjustbox}
\end{table}

\clearpage
\section{Reporting checklist}\label{sec:s_claim}

Table~\ref{tab:s_claim} maps the manuscript to the Checklist for Artificial Intelligence in Medical Imaging (CLAIM) \citep{mongan2020claim}, 2024 update \citep{tejani2024claim}, grouped by the sections of the checklist.

\begin{table}[!htbp]
\caption{Where each part of the CLAIM 2024 checklist is addressed. ESG: evidence sensitivity gap. LoRA: low-rank adapter.}
\label{tab:s_claim}
\centering
\small
\renewcommand{\arraystretch}{1.12}
\begin{tabular}{@{}L{0.25\textwidth}L{0.71\textwidth}@{}}
\toprule
\textbf{Checklist section} & \textbf{Where and how it is addressed} \\
\midrule
Title, abstract & Title; abstract (design, data, main results with intervals, and the comparison that was not shown). \\
Introduction: background, objectives, hypotheses & Section~\ref{sec:intro}: the clinical setting, the logical limit of accuracy on unaltered images, four requirements for an intervention on the finding, and the aim. Predictions for a model that uses the image and for one that does not: Section~\ref{subsec:design}. \\
Study design and intended use & Retrospective study of public, de-identified datasets. A benchmark and research models, not a clinical device (Sections~\ref{subsec:design} and~\ref{subsec:data}; Table~\ref{tab:overview}). \\
Data sources, selection, preprocessing & Sources and counts (Section~\ref{subsec:data}; Table~\ref{tab:datasets}; Supplementary Table~\ref{tab:s_datasets}); lesions smaller than 20 pixels or larger than 20\% of the image excluded; closed questions; automated checks of the edited views (Section~\ref{subsec:views};~\ref{sec:s_views}); long side of every image at most 768 pixels (Section~\ref{subsec:models}). \\
De-identification & Public releases, de-identified by their providers; no new patient data were collected; the reader study used the same images. \\
Missing data & Cases that failed an automated check or an edit were dropped with both edited views; exclusions before splitting are counted in~\ref{sec:s_plan}. Analyses on fewer cases state the number kept and the reason: 125 cases with valid views for all four editing tools, 116 of 160 clips with an unedited pair, 183 of 200 chest X-ray follow-up pairs (Sections~\ref{subsec:data} and~\ref{subsec:pairs};~\ref{sec:s_views}). \\
Imaging parameters & As released by each source; not available per image and not analysed (Section~\ref{sec:discussion}, limitation iv). \\
Reference standard & Expert lesion masks and labels of the source datasets. For an erased image, absence by construction, assessed so far by one board-certified specialist per imaging type (a radiologist and a gastroenterologist) and by an earlier round with two trainee physicians; further radiologists are reading (Sections~\ref{subsec:measures}, \ref{subsec:readers} and~\ref{subsec:res_readers}; \ref{sec:s_readers}). \\
Data partitions and independence & Splits by patient or by group of near-identical images, made before any view was built; hash comparison of all images across sets; an analysis without the test cases most similar to a training image; cross-validated answers where a source serves in training and in testing (Sections~\ref{subsec:data}, \ref{subsec:models} and~\ref{subsec:res_trained}). \\
Test-set size & Set by the available cases and fixed in advance, not by a power calculation (Section~\ref{subsec:data}). \\
Model, software, initialisation & \base{} with a LoRA adapter of rank 16 and a frozen vision encoder, initialised from the public base model; software versions (Sections~\ref{subsec:models} and~\ref{subsec:stats}; \ref{sec:s_plan}). \\
Training and model selection & Three training stages and a selection rule fixed in advance for \ms, the primary model; \msp{} and \msn{} post hoc, each with a validation rule fixed before its test results; three training seeds; for \msn, a three-seed majority vote reported next to the single model (Section~\ref{subsec:models}; Table~\ref{tab:models}; Supplementary Tables~\ref{tab:s_models} and~\ref{tab:s_mix}). \\
Metrics and statistics & ESG, erased-lesion claims, lesion-free claims, lesion images correct, net utility, natural ESG and Youden index, each with the value a constant answer earns (Section~\ref{subsec:measures}); bootstrap over cases, videos or patients, paired comparisons, Bonferroni correction over 27 primary comparisons (Section~\ref{subsec:stats}). \\
Robustness and interpretability & Prompting conditions, question wordings, cost matrices, editing tools, lesion size and training seeds (Sections~\ref{subsec:res_main}, \ref{subsec:res_trained} and~\ref{subsec:res_alternatives}; Supplementary Tables~\ref{tab:s_prompts}, \ref{tab:s_sensitivity}, \ref{tab:s_cost}, \ref{tab:s_tools}, \ref{tab:s_size} and~\ref{tab:s_seeds}); probes of hidden states (Supplementary Table~\ref{tab:s_site}). \\
Internal and external testing & Main test; external datasets, fresh internal and fresh external sets (Section~\ref{subsec:res_external}, Supplementary Table~\ref{tab:s_external}); unedited same-patient pairs, including a hospital not used in training (Section~\ref{subsec:res_pairs}); chest films (Section~\ref{subsec:res_nmodel}). \\
Registration & The analysis plan is a set of dated files in the public project repository, the authors' own record and not a third-party registry (Section~\ref{subsec:stats}, Supplementary Table~\ref{tab:s_plan}). The study is not a clinical trial. \\
Data flow and demographics & Counts per source and per evaluation set (Table~\ref{tab:datasets}; Supplementary Table~\ref{tab:s_datasets}); demographics were not compiled (Section~\ref{sec:discussion}, limitation iv). \\
Performance and errors & Tables with intervals; results by imaging type, editing tool and lesion size (Supplementary Tables~\ref{tab:s_modality}, \ref{tab:s_tools} and~\ref{tab:s_size}); a counted analysis of the missed pneumothoraces (Section~\ref{subsec:res_external}); model answers re-scored on erasures that both trainee readers called clean (Section~\ref{subsec:res_readers}). \\
Discussion & What the results do and do not show, limitations and the evidence still needed (Section~\ref{sec:discussion}). \\
Protocols, code, data & Data and code availability statement: analysis plan, code, models and model answers. \\
Funding & No external funding. \\
\bottomrule
\end{tabular}
\end{table}

\end{document}